\documentclass[journal]{IEEEtran}
\ifCLASSINFOpdf
\else
\fi
\usepackage{amsmath}
\usepackage{mathtools}
\usepackage{epsfig}
\usepackage{graphicx}
\usepackage{amssymb}
\usepackage{multirow}
\usepackage{hhline}
\usepackage{capt-of}
\usepackage[dvipsnames]{xcolor}
\usepackage[caption=false,font=footnotesize]{subfig}
\usepackage{lipsum}
\usepackage{footnote}
\makesavenoteenv{tabular}
\usepackage{listings}
\usepackage{enumitem}

\newcommand{\conv}[1]{{\color{RoyalBlue}#1}}
\newcommand{\deconv}[1]{{\color{PineGreen}#1}}

\DeclarePairedDelimiter{\ceil}{\lceil}{\rceil}
\begin{document}
\setcounter{table}{0}
\setcounter{figure}{0}
%
\title{InversionNet3D: Efficient and Scalable Learning for 3D Full Waveform Inversion}
%
%
%

\author{Qili~Zeng$^{*}$, Shihang~Feng,  Brendt~Wohlberg,~\IEEEmembership{Senior~Member,~IEEE}, and Youzuo Lin$^{*}$,~\IEEEmembership{Member,~IEEE}
\thanks{The authors are with Los Alamos National Laboratory, Los Alamos, NM 87544, USA.
($^{*}$Corresponding authors: Qili Zeng (qzeng@lanl.gov) and Youzuo Lin (ylin@lanl.gov)).}
}

%
%

\markboth{IEEE Transactions on Geoscience and Remote Sensing}%
{Zeng \MakeLowercase{\textit{et al.}}: InversionNet3D: Efficient and Scalable Learning for 3D Full Waveform Inversion}
%



\maketitle

\begin{abstract}
Seismic full-waveform inversion~(FWI) techniques aim to find a high-resolution subsurface geophysical model provided with waveform data. Some recent effort in data-driven FWI has shown some encouraging results in obtaining 2D velocity maps. However, due to high computational complexity and large memory consumption, the  reconstruction of 3D high-resolution velocity maps via deep networks is still a great challenge. In this paper, we present InversionNet3D, an efficient and scalable encoder-decoder network for 3D FWI. The proposed method employs group convolution in the encoder to establish an effective hierarchy for learning information from multiple sources while cutting down unnecessary parameters and operations at the same time. The introduction of invertible layers further reduces the memory consumption of intermediate features during training and thus enables the development of deeper networks with more layers and higher capacity as required by different application scenarios.  Experiments on the 3D Kimberlina dataset demonstrate that InversionNet3D achieves state-of-the-art reconstruction performance with lower computational cost and lower memory footprint compared to the baseline. 
\end{abstract}


%
\IEEEpeerreviewmaketitle

\section{Introduction}
\label{sec:intro}

\begin{figure*}
    \centering
    \includegraphics[width=0.98\linewidth]{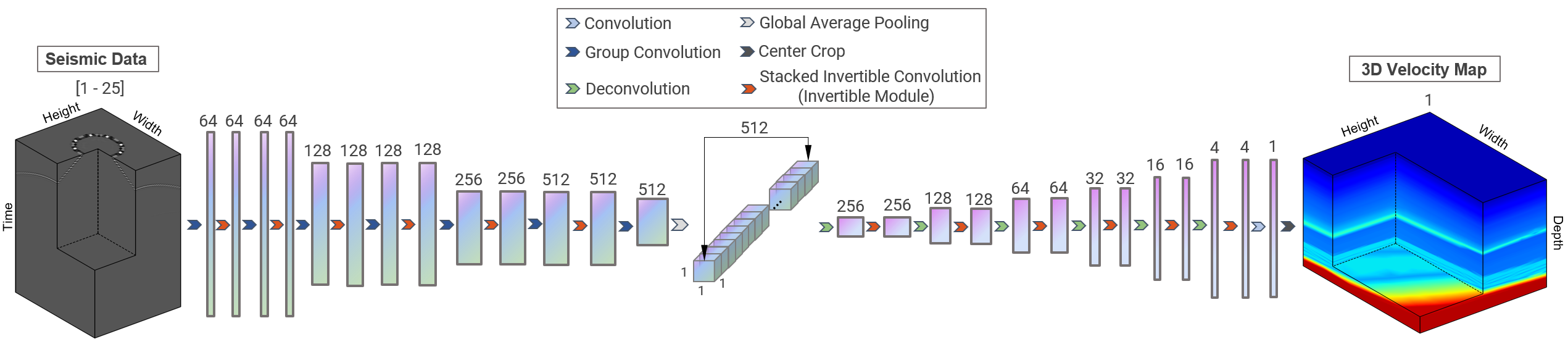}
    \begin{minipage}{0.96\linewidth}
    \centering
    \caption{\textbf{Network Architecture of InversionNet3D.} The shape of displayed tensors is adjusted for better perception and thus does not reflect the real size of data. The number above each tensor represents the number of channels. There are 25 available seismic data records per sample in our experiment, each resulted from one seismic source and they are considered as different channels in the input. We extract one of the channels for display. We crop out a sub-volume from seismic data and velocity map, respectively, for clearer visualization of their internal structure.}
    \label{fig:front}
    \end{minipage}
\vspace{-0.75em}
\end{figure*}
\IEEEPARstart{F}{ull}  waveform inversion (FWI) is a high-resolution seismic imaging technique that reconstructs the velocity maps by making use of seismic waveform information~\cite{tarantola2005inverse}. Given an initial velocity map, FWI aims to find an improved map by minimizing the residual between the predicted and observed seismic data~\cite{feng2019transmission+}. Depending on the optimization techniques used to solve  FWI, two categories of methods have been developed: those based on local gradient-related methods~\cite{virieux2009overview} and those based on global optimization methods~\cite{datta2016estimating,mazzotti2016two}. The gradient-based FWI methods utilize an efficient forward-modeling engine and a local differential approach to calculate the gradient term~(and the Hessian term if needed)~\cite{virieux2009overview}. On the other hand, FWI methods with global optimization~\cite{datta2016estimating,mazzotti2016two}, employ stochastic global optimization approaches to overcome the local minima issue. Solving FWI can be computationally expensive and this is particularly true for 3D scenarios. When using local gradient-based methods, the cost of forward-modeling and the gradient calculation in the 3D scenarios will be significantly larger than that in the 2D scenarios~\cite{zheng20183d,wang20193d}. Similarly, the computational cost will exponentially increase with the number of unknowns for the global optimization methods~\cite{sajeva2016estimation}. Thus solving 3D FWI is still a challenging problem due to the high computational cost.



FWI and many image generation tasks in computer vision such as style transfer \cite{gatys2015,gatys2016,luan2017,johnson2016}, image-to-image translation \cite{isola2017,zhu2017b,zhu2017c}, and cross-modal image synthesis \cite{reed2016,zhang2017b,zhang2019a,xu2018} share similar problem background and learning objectives. Due to the advances in image generation based on deep learning, some researchers have made attempts in using deep networks to reconstruct velocity maps from seismic data in an end-to-end manner. Deep models with recurrent network \cite{fabien-ouellet2019a}, encoder-decoder architecture \cite{li2020c,wang2018,yang2019,wu2020a} or generative adversarial networks (GAN) \cite{zhang2019c} achieve acceptable performance in FWI. However, there are also obvious limitations associated with these prior methods including the weak generalization ability, the lack of underlying physics, and availability to only 2D FWI problems. Although 2D FWI can be useful in some scenarios, real-world applications often require 3D velocity map reconstruction for a more detailed and more comprehensive analysis of subsurface structures.

Nevertheless, simple inflation of 2D convolution kernels into their 3D counterparts is not adequate for solving the problem. 3D networks typically involve a large number of parameters, which brings about extra obstacles in optimization. Considering that current 3D FWI datasets~\cite{aminzadeh1996three, vigh2014elastic, guo2017elastic} contain very limited numbers of samples, large-scale 3D networks are very likely to overfit to the dataset during training. Moreover, 3D FWI leads to much larger memory consumption due to input, output and intermediate features all being volumetric. This further hinders building deeper networks or establishing models with special components. It has previously been demonstrated that deeper networks with larger capacity can serve as better feature extractors \cite{simonyan2015,he2016}. Some complementary methods such as attention mechanism \cite{xu2018}, multi-task learning \cite{li2020b}, and multi-scale processing \cite{leong2019} have also been proved to help generate image of higher quality. These techniques may also lead to better reconstruction performance in 3D FWI. However, when a basic encoder-decoder network already occupies most of the memory, adding extra layers or components becomes nearly impossible. Therefore, it is essential to have a compact, efficient, and scalable backbone network in order to address complex challenges in 3D FWI.

In this paper, we propose InversionNet3D (abbreviated as InvNet3D), an efficient and scalable network to solve 3D FWI. InvNet3D employs group convolution in its encoder to reduce unnecessary global operations on channel dimension and establishes a hierarchical framework of learning and aggregating channel information. Compared to prior networks with conventional convolution with a global receptive field over channels, InvNet3D better utilizes information from multiple sources with fewer parameters and lower computational complexity. Furthermore, in order to control the memory footprint resulting from intermediate representations, InvNet3D adopts invertible layers in every convolutional block, which transforms the model into a partially reversible network. Input tensors to invertible layers no longer need to be stored during inference since they can be easily restored from the output in the backward pass, which saves memory during training at an affordable cost of training time, and enables the introduction of additional layers and structures to the model according to a specific environment and task requirements. 

Our paper is organized as follows. Section \ref{sec:related-works} briefly reviews related works. Section \ref{sec:methods} provides detailed description of the baseline model (Section \ref{sec:baseline}), the Channel-Separated Encoder built upon group convolution (Section \ref{sec:group-conv}), and Invertible Modules (Section \ref{sec:invertible-module}). Section \ref{sec:experiment} introduces InversionNet3D, compares it to the baseline models through experiments, and analyzes the improvement in detail. Section \ref{sec:as} conducts ablation studies on the influence of temporal subsampling and source selection strategy as well as the robustness and the generalization of our model in new environment. Concluding remarks are presented in Section \ref{sec:conclusion}. Additional diagrams, visualization, and explanations to the experiments are provided in the supplementary materials.

\section{Related Works}
\label{sec:related-works}
\subsection{Deep Learning for Full Waveform Inversion}
The success of deep networks in many image generation tasks indicates their great potential in handling the ill-posedness of seismic inversion. Compared to traditional FWI methods, deep models are much more efficient since the inference can be done within seconds after a few hours'  training. Many prior explorations focused on data-driven end-to-end reconstruction, which regarded FWI as a pure image-to-image translation problem. GeoDNN \cite{araya-polo2018} made an early attempt at using a fully connected network with 8 layers for FWI. InversionNet \cite{wu2020a} then introduced a encoder-decoder network with modern convolutional neural network (CNN) features. ModifiedFCN \cite{wang2018} employed a similar network design and SeisInvNet \cite{li2020c} enhanced each seismic trace with auxiliary knowledge from neighborhood traces for better spatial
correspondence. VelocityGAN \cite{zhang2019c} achieved improved performance by using generative adversarial networks (GANs). 
Long Short-Term Memory (LSTM)~\cite{hochreiter1997} was also used for decoding representations of seismic data into velocity vectors~\cite{fabien-ouellet2019a}. Some researchers have explored the combination of deep learning and traditional FWI pipelines, which takes the physical background into consideration. NNFWI \cite{zhu2020a} used deep models to generate a physical velocity model, which is then fed to a PDE solver to simulate seismic waveforms. A thorough review on deep learning for FWI can be found in~\cite{Adler2021Deep}. Although these effective neural networks are inspiring to the area and demonstrated the advantage of deep learning in performance and efficiency, they all focused on 2D FWI, which limits their application in real-world problems. To the best of our knowledge, InversionNet3D, proposed in this paper, is the first deep-learning-based solution in addressing 3D FWI. It is also worth mentioning that, another deep learning approach to solve 3D FWI \cite{xu2021} is proposed in the same time frame as our work, though targeted at a different application scenario.


\subsection{Invertible Networks}
Invertible networks are a special kind of neural networks that allow the input to be reconstructed from the output. The invertibility is typically obtained via coupling layers \cite{dinh2015a, dinh2017, kingma2018, gomez2017, jacobsen2018}, autoregressive models \cite{papamkarios2017, kingma2016}, or numerical inversion \cite{chen2018, behrmann2019, chen2019}. Invertible networks has been demonstrated to provide desirable performance in discriminative tasks \cite{dinh2015a, dinh2017, kingma2018}, generative tasks \cite{gomez2017, jacobsen2018}, and solving both problems with one model \cite{nalisnick2019}. Moreover, a full-invertible networks built upon a sequence of invertible operations alleviates the need to store intermediate activations in memory for gradient calculation during back-propagation~\cite{rumelhart1986}. Partial invertibility could also lead to substantial memory saving \cite{gomez2017, brugger2019} when non-invertible operators, such as downsampling and deconvolution, exist in certain stage of the network.

\begin{table*}[]
\centering
\renewcommand\arraystretch{1.25}
\begin{minipage}{0.9\linewidth}
\caption{\textbf{InversionNet3D-Simple (InvNet3DS) Architecture.} The properties of each convolutional/deconvolutional layer are listed as kernel size ($t\times w\times h$), number of kernels, and stride ($t\times w\times h$). Layers with kernel size colored in \conv{blue} are \conv{convolutional layers} while the others with kernel size colored in \deconv{green} are \deconv{deconvolutional layers}. If not specified, the kernel is with a default stride of $1\times1\times1$. All the layers are followed by a batch normalization \cite{ioffe} layer and a LeakyReLU activation \cite{maas2013}, except layer conv\_7 in the decoder, which is followed by a batch normalization layer and a hyperbolic tangent activation. }
\label{tab:invnet3d-simple-arch}
\end{minipage}
\begin{minipage}{0.45\linewidth}
\centering
\begin{tabular}{c|c|c}
\hline
Layer                     & Output Shape                           & InvNet3D-Encoder                        \\ \hline
\multirow{2}{*}{conv1\_x} & \multirow{2}{*}{$\ceil{T/3}\times40\times40$} & $\conv{7\times3^2}$, $64$, stride $3\times1^2$ \\ \cline{3-3} 
                          &                                        & $\conv{3\times3^2}$, $64$                                         \\ \hline
\multirow{2}{*}{conv2\_x} & \multirow{2}{*}{$\ceil{T/6}\times40\times40$}                      &$\conv{3\times3^2}$, $64$, stride $2\times1^2$                                         \\ \cline{3-3} 
                          &                                        & $\conv{3\times3^2}$, $64$                                         \\ \hline
\multirow{2}{*}{conv3\_x} & \multirow{2}{*}{$\ceil{T/12}\times20\times20$}                      &$\conv{3\times3^2}$, $128$, stride $2\times2^2$                                         \\ \cline{3-3} 
                          &                                        &$\conv{3\times3^2}$, $128$                                         \\ \hline
\multirow{2}{*}{conv4\_x} & \multirow{2}{*}{$\ceil{T/24}\times20\times20$}                      & $\conv{3\times3^2}$, $128$, stride $2\times1^2$                                        \\ \cline{3-3} 
                          &                                        &$\conv{3\times3^2}$, $128$                                         \\ \hline
\multirow{2}{*}{conv5\_x} & \multirow{2}{*}{$\ceil{T/48}\times10\times10$}                      & $\conv{3\times3^2}$, $256$, stride $2\times2^2$                                        \\ \cline{3-3} 
                          &                                        &$\conv{3\times3^2}$, $256$                                         \\ \hline
\multirow{2}{*}{conv6\_x} & \multirow{2}{*}{$\ceil{T/96}\times10\times10$}                      &$\conv{3\times3^2}$, $512$, stride $2\times1^2$                                         \\ \cline{3-3} 
                          &                                        &$\conv{3\times3^2}$, $512$                                         \\ \hline
conv7                     & $\ceil{T/192}\times5\times5$                                       &$\conv{3\times3^2}$, $512$, stride $2\times2^2$                                         \\ \hline
GAP                          & $1\times1\times1$                  & global average pooling \\ \hline
\end{tabular}
\end{minipage}
\begin{minipage}{0.45\linewidth}
\centering
\begin{tabular}{c|c|c}
\hline
Layer                     & Output Shape                           & InvNet3D-Decoder                        \\ \hline
\multirow{2}{*}{conv1\_x} & \multirow{2}{*}{$2\times2\times2$} & $\deconv{4\times4^2}$, $256$, stride $2\times2^2$ \\ \cline{3-3} 
                          &                                        & $\conv{3\times3^2}$, $256$                                         \\ \hline
\multirow{2}{*}{conv2\_x} & \multirow{2}{*}{$4\times4\times4$}     & $\deconv{4\times4^2}$, $128$, stride $2\times2^2$                                         \\ \cline{3-3} 
                          &                                        & $\conv{3\times3^2}$, $128$                                         \\ \hline
\multirow{2}{*}{conv3\_x} & \multirow{2}{*}{$8\times8\times8$}     &$\deconv{4\times4^2}$, $64$, stride $2\times2^2$                                         \\ \cline{3-3} 
                          &                                        &$\conv{3\times3^2}$, $64$                                         \\ \hline
\multirow{2}{*}{conv4\_x} & \multirow{2}{*}{$24\times16\times16$}  & $\deconv{5\times4^2}$, $32$, stride $3\times2^2$                                        \\ \cline{3-3} 
                          &                                        &$\conv{3\times3^2}$, $32$                                         \\ \hline
\multirow{2}{*}{conv5\_x} & \multirow{2}{*}{$72\times80\times80$}  & $\deconv{5\times7^2}$, $16$, stride $3\times5^2$                                        \\ \cline{3-3} 
                          &                                        &$\conv{3\times3^2}$, $16$                                         \\ \hline
\multirow{2}{*}{conv6\_x} & \multirow{2}{*}{$360\times400\times400$} &$\deconv{7\times7^2}$, $4$, stride $5\times5^2$                                         \\ \cline{3-3} 
                          &                                        &$\conv{3\times3^2}$, $4$                                         \\ \hline
conv7                     & $360\times400\times400$                &$\conv{3\times3^2}$, $1$                                         \\ \hline
Crop                      & $350\times400\times400$                & center crop \\ \hline
\end{tabular}
\end{minipage}
\vspace{-0.75em}
\end{table*}

\subsection{Efficient Deep Learning}
Deep models with an enormous number of parameters trained on large-scale dataset achieve superior performance on many tasks. However, restricted computing resources and limited labeled data in real-world applications often impede the deployment of deep models. In order to adapt these over-parameterized, data-hungry, and computationally intensive models to a real-world environment, recent attempts have been made to reduce the dependency of deep networks on computing hardware and large-scale datasets. 3D FWI is a typical case where the input and output data are very large and are very expensive to acquire. Therefore, it is desirable that deep models designed for 3D FWI are light-weight and memory-friendly and can be trained with as little data as possible. The proposed method in this paper mainly focuses on achieving the first two goals. Several approaches have been proposed to reduce the number of parameters and/or the number of operations through group convolution \cite{howard2017, sandler2018, zhang2018b, ma2018a}, or neural architecture search \cite{tan2019b, tan2019a}, or the replacement of normal operator with hardware-friendly equivalent ones \cite{wu2018a, he2019, lin2019}. 
The reduction of memory footprint in a memory-friendly model has been obtained via domain-specific network design~\cite{liu2019a} or storing fewer activations during training \cite{chen2016a, gomez2017}. Similar to Partially Reversible U-Net \cite{brugger2019} for medical image segmentation, our model based on invertible layers falls within the second category.

\section{Proposed Methods}
\label{sec:methods}

\subsection{Baseline}
\label{sec:baseline}

In order to demonstrate the effectiveness of the proposed methods, we first establish a baseline model with simple network structures. We inflate all the 2D convolutions in InversionNet \cite{wu2020a} (InvNet) into corresponding 3D counterparts and denote this baseline model as InversionNet3D-Simple (InvNet3DS). The detailed network architecture of InvNet3DS is provided in Table \ref{tab:invnet3d-simple-arch}. 

Although this baseline model is designed with reference to InvNet and thus has the same number of layers and similar topological structures, we made extra adjustments on necessary layer parameters due to the large memory consumption of 3D FWI. As the decoder network gradually magnifies the input vector via deconvolution, the intermediate representations also become larger in deeper layers and occupy more GPU memory. In order to alleviate the heavy burden of memory usage, InvNet3DS presents a much faster shrinking of the size of the channel dimension. Convolutional blocks that are close to the output layer have many fewer filters compared to InvNet and thus the number of large volumetric output features becomes smaller. We also place the layers with a larger amplification factor between the size of output and input feature at the deeper stage of the decoder, leading to lower overall memory consumption. We found empirically that our computational environment supports a maximum of 16 filters in conv6\_x. However, for the following experiments, the number of filters in this block is set to 4 due to the need to compare with deeper network architectures.

Considering the temporal length of input seismic data is typically much larger than its width and height, the encoder network of InvNet3DS perform more aggressive downsampling on temporal dimension than on spatial dimensions. In addition, unlike InvNet which uses a convolutional layer with a large kernel for final spatio-temporal downsampling, InvNet3DS utilizes global average pooling after conv\_7 in the encoder network to generate a 512-dimensional compact representation for seismic data. This enables the network to handle input data of different spatial sizes and temporal lengths without modifying the structure of preceding layers. 

\subsection{Channel-Separated Encoder}
\label{sec:cs-encoder}

Conventional convolution aggregates information on channel dimension in a fully-connected manner. Every convolutional filter receives the input of all the channels. However, these dense connections between input and output feature maps on channel dimension lead to high computational cost and larger model size. Consider a convolutional layer with the input of size $C_{i}\times T\times H\times W$ and $C_{o}$ filter of size $C_{i}\times t\times h\times w$, then the number of parameters and floating-point operations (FLOPs) can be calculated as
\begin{eqnarray}
\mathrm{\#Params} &=& C_{i}\times C_{o}\times (thw),  \\
\mathrm{FLOPs} &\approx& 2\times C_{i}\times C_{o}\times (thw) \times (THW) \;.
\end{eqnarray}

\begin{figure}
\centering
\subfloat[]{%
    \includegraphics[width=0.45\linewidth]{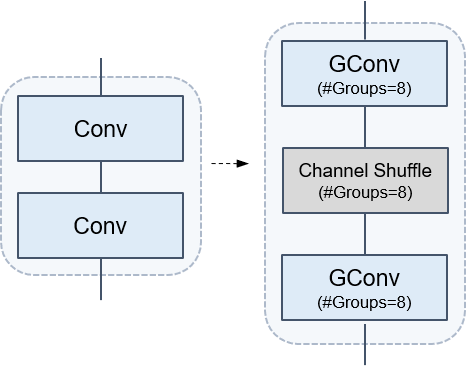}%
    \label{fig:group-conv-structure}
}
\hfill
\subfloat[]{%
    \includegraphics[width=0.45\linewidth]{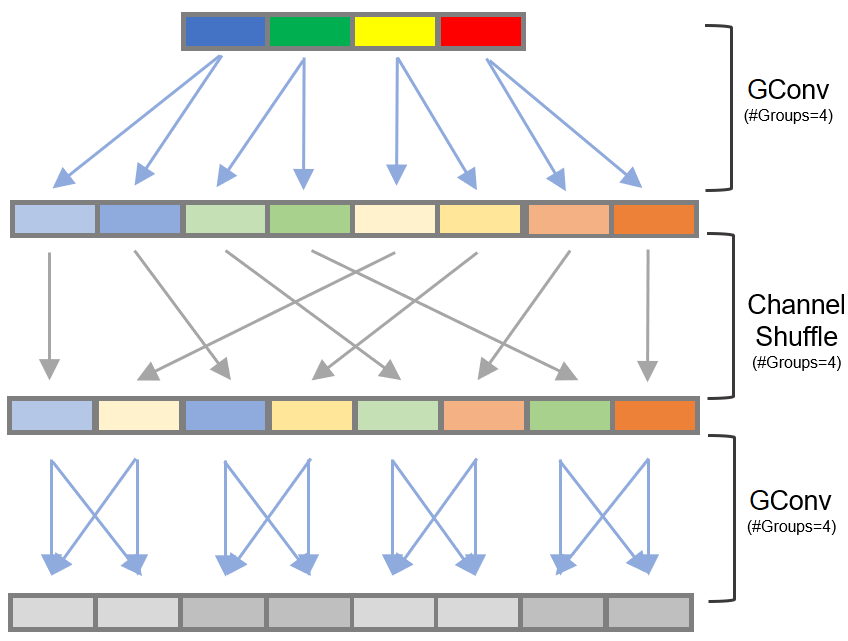}%
    \label{fig:group-conv-example}}
\caption{\textbf{Illustration of a two-layer group convolution block with channel shuffle.}  (a) The conversion of a convolution block into a group convolution block with channel shuffle. Three components in the new block share the same group size. (b) Dependencies and reorganization of information on channel dimension. The example describes the computational workflow with a 4-channel input, an 8-channel output and a group size of 4.}
\label{fig:group-conv}
\vspace{-1.0em}
\end{figure}

By dividing $C_{o}$ filters of a conventional convolutional layer into $G$ group with each group corresponding to $1/G$ channels in the input, a group convolutional layer also divides \#Params and FLOPs by $G$ since the channel size of every filter becomes $C_{i}/G$. Note that a group convolutional layer with the group size of $1$ is equivalent to a conventional convolutional layer. Although group convolution makes the network smaller and more efficient, stacked group convolutional layers with the same group size actually form several sub-networks without interactions in between. This strong regularization and restriction on information exchange may lead to inferior performance. To enforce the interaction between sub-networks, channel shuffle \cite{zhang2018b} is introduced to swap feature maps on channel dimension after one group convolutional layer so that every group in the next group convolutional layer perceives information from different proceeding sub-networks. An illustration of a \textit{GConv}-\textit{CS}-\textit{GConv} block is given in Figure \ref{fig:group-conv-structure}, where \textit{GConv} and \textit{CS} stand for group convolution and channel shuffle, respectively. We also provide an example of the computing workflow of such a group convolutional block in Figure \ref{fig:group-conv-example}. It can be observed that, with the group size equal to the channel size of the input, the first group convolutional layer in the block generate $C_{o}/C_{i}$ representations for each input channel and every channel in the final output feature maps represents certain information generated from $C_{o}/C_{i}$ channels in the input. 

By converting all the convolutional blocks in the encoder into group convolutional blocks, we obtain a more light-weight and more efficient encoder. We denote the result of such conversion as a Channel-Separated Encoder. Channel-Separated Encoder in seismic data processing, offers extra benefits in addition to higher efficiency. Typically, a complete set of seismic data consists of multiple records resulting from sources placed at different locations on the testing area. In prior research \cite{li2020c, wang2018, yang2019, wu2020a, zhang2019c}, these records are placed along the channel dimension of the input tensor. However, unlike RGB channels in natural images, which represent different color components, channels in seismic input correspond to different spatial information. Conventional convolutional layers, in the context of FWI, implement a sub-optimal strategy of fusing information from these channels since the semantic-level relationship and correspondence between records could hardly be found at shallow layers in the encoder. Feeding every filter with all the channels of raw input or low-level features may also lead to information loss due to the limited amount of filters at shallow layers. A progressive channel fusion scheme should serve as a better solution, which is what a Channel-Separated Encoder implements. Consider stacking the aforementioned \textit{GConv}-\textit{CS}-\textit{GConv} block illustrated in Figure~\ref{fig:group-conv-structure}. Different components generated from each input channel will be gradually aggregated, leading to a hierarchical learning scheme where lower-level filters are dedicated to extracting useful information from local channels and higher-level filters fuse local representations to form a global description for seismic records of multiple sources. In particular, layer conv\_7 in this encoder is with a group size of 512, which is referred to as depthwise convolution \cite{chollet2017}.

\label{sec:group-conv}
\subsection{Invertible Module}
\label{sec:invertible-module}

\begin{figure}
\centering
\subfloat[]{%
    \includegraphics[width=0.5\linewidth]{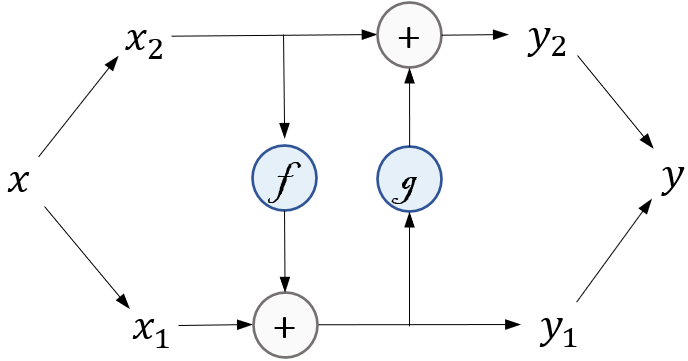}%
    \label{fig:invertible-forward}
}
\subfloat[]{%
    \includegraphics[width=0.5\linewidth]{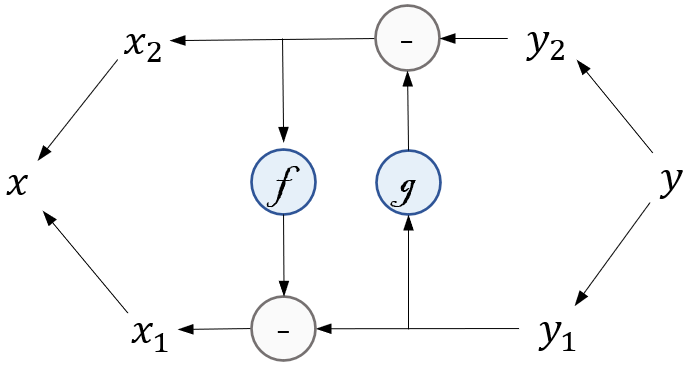}%
    \label{fig:invertible-backward}}
\caption{\textbf{Computational workflow for an invertible layer in forward pass (a) and backward pass (b).} $f(\cdot)$ and $g(\cdot)$ in our models are convolutional layers that preserve the shape of input tensor.}
\label{fig:invertible-compgraph}
\vspace{-1.0em}
\end{figure}

Since back-propagation \cite{rumelhart1986} is now the standard method of weight updating in modern neural networks, the inputs to all the layers have to be stored during inference for future gradient calculation in the backward pass. In the training stage, the memory consumption of intermediate activations can be orders of magnitude larger then that of model parameters. Therefore, minimizing the need to store activations can substantially reduce memory usage.
Gradient Checkpointing \cite{chen2016a} involves a strategy of only storing activations at certain layers and recomputing the missing ones through additional partial forward pass when needed. Invertible networks make a step further by enabling the reconstruction of input from the output at all the invertible layers.

Invertible layers can be realized in many ways, as reviewed in Section \ref{sec:related-works}. Similar to \cite{dinh2015a}, our implementation is based on additive coupling. This kind of invertible layer consists of a pair of operators $f$ and $g$ entangled with each other in a way that is also used in Lifting Scheme \cite{sweldens1998} for fast in-place calculation of the wavelet transform. In the forward pass, the input $\mathbf{x}$ will be divided into two parts along channel dimension, denoted as $\mathbf{x}_1$ and $\mathbf{x}_2$ and their corresponding output $\mathbf{y}_1$ and $\mathbf{y}_2$ can be calculated via 
\begin{equation}
\label{equ:invertible-forward}
\begin{aligned}
    \mathbf{y}_1 = \mathbf{x}_1 + f(\mathbf{x}_2), \\
    \mathbf{y}_2 = \mathbf{x}_2 + g(\mathbf{y}_1).
\end{aligned}
\end{equation}

The complete output $\mathbf{y}$ can be obtained by simple concatenation of $\mathbf{y}_1$ and $\mathbf{y}_2$ on channel dimension. In the backward pass, the input $\mathbf{x}$ can be figured out in the same way based on $\mathbf{x}_1$ and $\mathbf{x}_2$ as
\begin{equation}
\label{equ:invertible-backward}
\begin{aligned}
    \mathbf{x}_2 = \mathbf{y}_2 - g(\mathbf{y}_1), \\
    \mathbf{x}_1 = \mathbf{y}_1 - f(\mathbf{x}_2).
\end{aligned}
\end{equation}

\begin{figure}
\centering
\subfloat[]{%
    \includegraphics[width=0.5\linewidth]{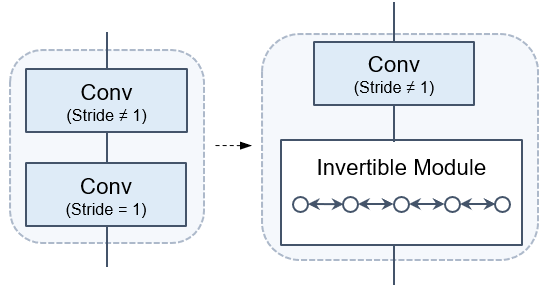}%
    \label{fig:conv-invertible}
}
\subfloat[]{%
    \includegraphics[width=0.5\linewidth]{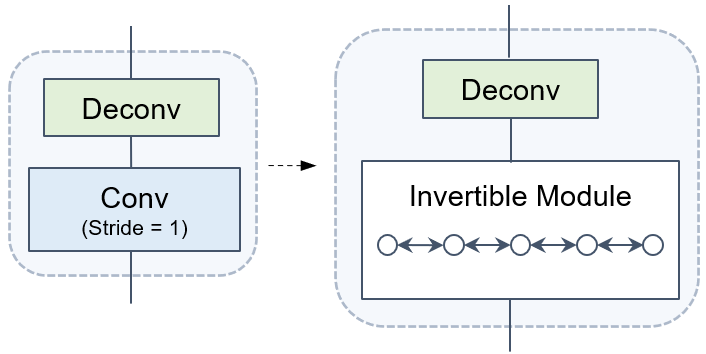}%
    \label{fig:deconv-invertible}}
\caption{\textbf{Illustration of a convolution/deconvolution block with an invertible module.}  (a) and (b) illustrate the replacement of the second convolutional layer in the corresponding block with an invertible module, which may consist of multiple invertible layers. The second convolutional layers are typically established with the stride of 1, leading to an unchanged shape between input and output features, which makes the direct replacement possible.}
\label{fig:invertible-insertion}
\vspace{-0.75em}
\end{figure}

Figure~\ref{fig:invertible-compgraph} provides an intuitive illustration of the computation above. Generally speaking, $f$ and $g$ can be arbitrary types of operators; in our context, we would expect them to be both convolutional layers followed by an activation function.  However, it can be observed from Equations~\eqref{equ:invertible-forward} and \eqref{equ:invertible-backward} that convolutional operators in an invertible layer must have a stride of 1 because otherwise the layer discards information and generates an output of different shape from the input. Considering the network architecture of InvNet3DS, the replacement of conventional layers with invertible layers can only take place at every second layer in each convolutional or deconvolutional block, e.g., conv1\_2, as illustrated in Figure \ref{fig:invertible-insertion}. This means an entire model with invertible layers is partially reversible since activations at non-invertible layers have to be preserved.

By stacking several invertible layers, we obtain an Invertible Module. Compared to a Non-invertible Module with $N$ stacked conventional layers, an Invertible Module of the same scale reduces the internal memory consumption from $O(N)$ to $O(1)$ since back-propagation could work as long as the activations and their derivatives for the top layer are given and thus all the memory taken by intermediate representation could be freed up. Invertible Modules introduce extra computational overhead during training due to another equivalent forward pass at each layer in gradient calculation. However, since the inference speed of a modern network is typically high, slightly longer training time is generally affordable. Note that Invertible Modules do not lead to slower inference and thus would not affect efficiency in a testing environment.

It is worth noticing that Invertible Modules implicitly introduce group convolution with a group size of 2. Therefore, Invertible Modules would also lead to a lighter-weighted network and potential inference acceleration. Furthermore, Invertible Modules could work with the proposed Channel-Separated Encoder as long as the latter is built with even group size. Suppose a Channel-Separated Encoder is with a group size of $2N$, in order to replace its every second convolutional layer with an Invertible Module, we simply need to convert the sub-layers, $f$ and $g$, in every invertible layer into group convolutional layers with a group size of $N$.

\section{Experiments}
\label{sec:experiment}
\subsection{Experimental Settings}

\subsubsection{Dataset}
\label{sec:dataset}

The 4D Kimberlina dataset~\cite{wagoner2009} is generated from hypothetical numerical models built on the geologic structure of a commercial-scale geologic carbon sequestration (GCS) reservoir at the Kimberlina site in the southern San Joaquin Basin, 30 km northwest of Bakersfield, CA, USA. Particularly, \cite{wagoner2009} describes the 3D geological model. The details of the overburden and reservoir models are provided in \cite{Wainwright2013Modeling} and \cite{Birkholzer2011Sensitivity}. The geophysical modeling for 3D P-wave velocity maps generation is given in \cite{Modeling-2018-Wang}. DOE-EDX platform~\cite{NETL-2018-Kimberlina} provides a portion of the Kimberlina Simulations. 

Our experimental dataset consists of 1,827 pairs of seismic data and 3D velocity maps  generated  with  20 different  non-uniformed  sampled  timestamps  ranging  from year 10 to year 200. We randomly split the dataset into a training subset and a validation subset with the ratio of 1664:163. Seismic data are simulated using the finite-difference method with 1,600 receivers uniformly distributed over the 2D earth surface with a receiver interval of 100 $m$ and each of them captures vibration signals as time-series data of length 5,001 with a time spacing of 0.001 $s$. In order to simplify the evaluation of our models, the seismic data in this dataset only involves acoustic-wave signals. 
 There are 25 sources serving as stimulus placed evenly on the 2D spatial grid over the surface with a shot interval of 800 $m$, each of which leads to one seismic record. Therefore, the shape of raw seismic data is $25\times5001\times40\times40$ ($\mathrm{\underline{C}hannel}\times \mathrm{\underline{T}ime}\times \mathrm{\underline{W}idth}\times \mathrm{\underline{H}eight}$). The corresponding velocity map as output, reflecting subsurface structures beneath the testing area, is with the size of $350\times400\times400$ ($\mathrm{\underline{D}epth}\times \mathrm{\underline{W}idth}\times \mathrm{\underline{H}eight}$), where the grid spacing is 10 $m$ in all dimensions.

\subsubsection{Training Configurations}
\label{sec:training-conf}
We found that it is not necessary to include all 25 channels in the input due to their high redundancy, so our input only contains randomly selected 8 channels (1, 2, 14, 15, 16, 20, 23, 24), which corresponds to a trivial scheme of source placement. See Figure \ref{fig:source-distribution} for an explanation of these serial numbers. We provide detailed discussion on channel selection in Section \ref{sec:as-source}. Moreover, since the temporal size of the raw seismic data is too large for a normal spatio-temporal encoder and the information on temporal dimension is visually redundant, we uniformly sample 896 frames from the raw sequence. The temporal downsampling strategy is further discussed in Section \ref{sec:as-temp-downsample}. Therefore, the default input seismic data has the shape $8\times896\times40\times40$. The ground truth velocity maps are rescaled into $[-1, 1]$ via min-max normalization in order to be compatible with the final output layer in the decoder with a $tanh(\cdot)$ activation.

We implement our models with PyTorch \cite{paszke2019} and train the models on 32 NVIDIA Tesla P100 GPUs with 16GB memory each. Batch Normalization \cite{ioffe} layers are synchronized across processes. We adopt the AdamW \cite{loshchilov2019} optimizer in all of our experiments with the initial learning rate of $1\times10^{-4}$ and a weight decay of $5\times10^{-4}$. We use the first 10 epochs for warm-up \cite{he2016} and divide the learning rate by 10 at epoch 40, 60, and 70. A typical training cycle consists of 80 epochs. 

\subsubsection{Evaluation Metrics}
\label{sec:metrics}
Since the output velocity map represents the transmission velocity of the seismic waves at each spatial position, we adopt Mean Absolute Error (MAE) as the main metric to measure the prediction error. We also employ Root Mean Squared Error (RMSE), which implicitly gives a higher weight to large errors, as a complementary metric. Since we found most large errors occur at the location where complex subsurface layers are present, RMSE should better reflect the prediction accuracy in important high-frequency areas. In addition, considering the velocity maps are often displayed as images for visual analysis, we also employ mean Structural Similarity (SSIM) \cite{zhouwang2004} for perceptual similarity evaluation. Note that MSE and RMSE are calculated with denormalized prediction and ground truth, i.e., velocity maps in their original data scale, while SSIM is computed in normalized $[-1, 1]$ scale as required by the algorithm. 

\subsection{Experimental Results}

In order to demonstrate the effectiveness of the proposed model, we first analyze the influence of Channel-Separated Encoder and Invertible Module on the reconstruction performance in Section \ref{sec:one-block}. Then we expand the Invertible Module to construct deeper networks and compare them with the models with stacked conventional convolution of the same scale in Section \ref{sec:n-blocks} for a demonstration of the model's scalability resulted from partial invertibility.

\begin{table}[]
\centering
\caption{\textbf{Naming convention of networks involved in experiments.} A checked box indicates the network contains corresponding component(s).}
\label{tab:network-name}
\begin{tabular}{l|cc}
\hline
\multicolumn{1}{c|}{Model} & Channel-Separated Encoder & Invertible Module \\\hline
InvNet3DS           &                        &                    \\\hline
InvNet3DI       &                        & $\surd$            \\\hline
InvNet3DG            & $\surd$                &                    \\\hline
InvNet3D                  & $\surd$                & $\surd$            \\\hline 
\end{tabular}
\vspace{-0.75em}
\end{table}

\subsubsection{Comparisons with the baseline}
\label{sec:one-block}

\begin{figure*}[]
\centering
\begin{minipage}{0.36\linewidth}
\centering
    \includegraphics[width=0.85\linewidth]{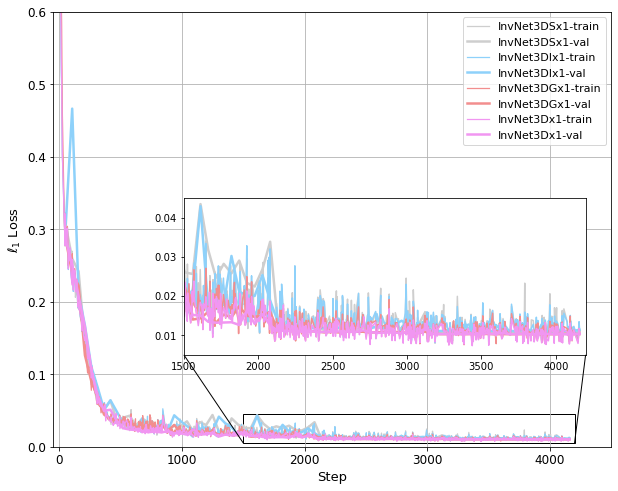}
    \begin{minipage}{0.95\linewidth}
    \centering
    \captionof{figure}{\textbf{Training and validation $\ell_1$ loss of models with $\mathrm{\#blocks}=1$.} Separated loss curves for each model can be found in the supplementary materials.}
    \label{fig:main-loss}
    \end{minipage}
\end{minipage}
\hspace{16pt}
\begin{minipage}{0.55\linewidth}
\centering
    \begin{minipage}{0.98\linewidth}
    \captionof{table}{\textbf{Performance Comparison between the baseline model (InvNet3DS) and the model with a channel-separated encoder (InvNet3DG)}. Computational complexity measured by floating point operations (FLOPs) is calculated during inference.}
    \label{tab:main-comparison}
    \end{minipage}
\footnotesize
    \renewcommand\arraystretch{1.2}
    \begin{tabular}{c|c|c|c|c|c|c}
    \hline
    Model     & \#Blocks          & \#Params & GFLOPs  & MAE $\downarrow$    & RMSE $\downarrow$   & SSIM $\uparrow$    \\ \hline
    InvNet3DS & \multirow{4}{*}{1}& 35.95M   & 3062.90  & 10.20  & 26.95  & 0.9820  \\
    InvNet3DI &                   & 30.97M   & 2953.02  & 10.38  & 27.36  & 0.9809  \\ 
    InvNet3DG &                   & 15.60M   & 2760.88 & 9.82   & 26.00  & 0.9831  \\ 
    InvNet3D  &                   & 14.42M   & 2734.54 & 9.83   & 26.11  & 0.9826   \\ \hline
    \end{tabular}
\end{minipage}
\vspace{-0.75em}
\end{figure*}

\begin{figure}
\centering
\subfloat[]{%
    \includegraphics[width=0.48\linewidth]{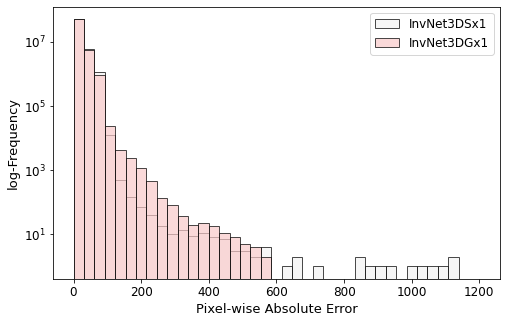}%
    \label{fig:error-hist-a}
}
\subfloat[]{%
    \includegraphics[width=0.48\linewidth]{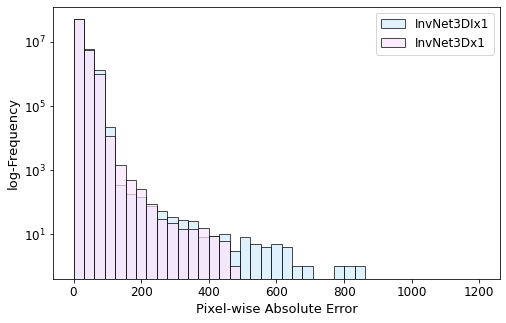}%
    \label{fig:error-hist-b}}
\caption{\textbf{Histograms of pixel-wise absolute error.}  (a) Comparison between InvNet3DS and InvNet3DG. (b) Comparison between InvNet3DI and InvNet3D. Statistical data obtained from evaluations on the whole validation subset.}
\label{fig:error-hist}
\vspace{-1.0em}
\end{figure}

\begin{figure*}[p]
\centering
    \subfloat[]{
        \begin{minipage}{0.18\linewidth}
        \centering
        \includegraphics[width=0.95\linewidth]{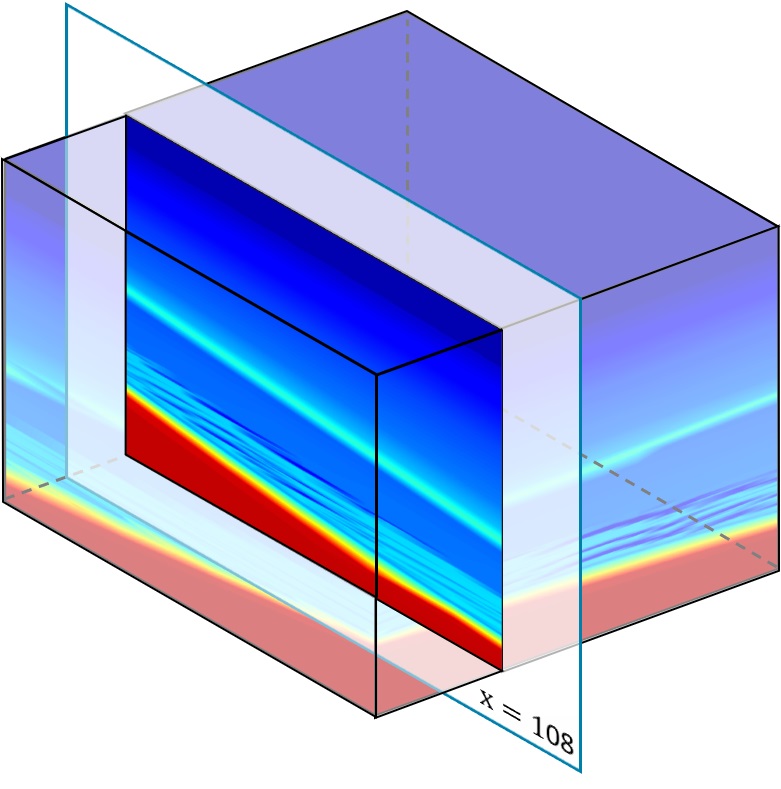} \\ \vspace{5pt}
        \includegraphics[width=0.98\linewidth]{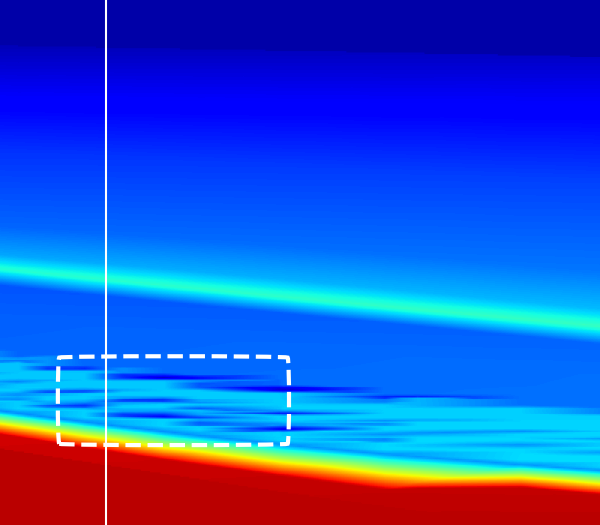} \\ \vspace{5pt}
        \includegraphics[width=0.98\linewidth]{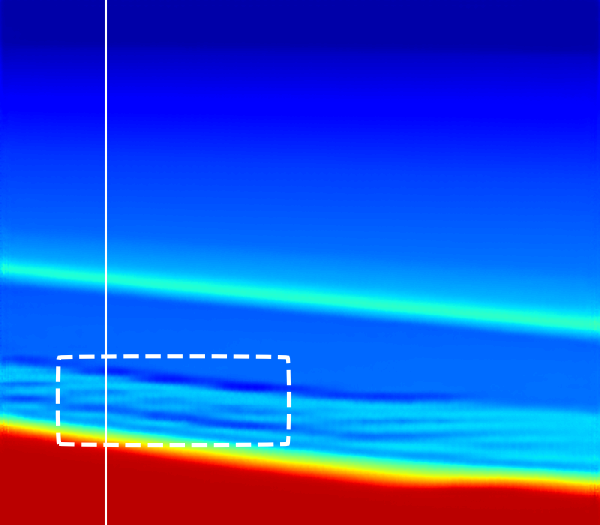} \\ \vspace{5pt}
        \includegraphics[width=0.98\linewidth]{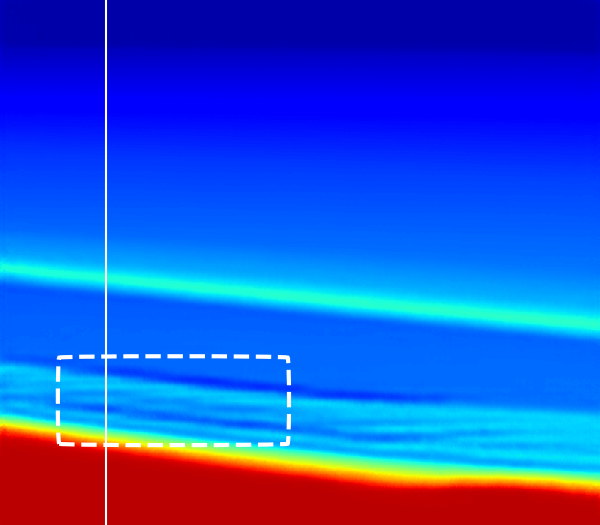} \\ \vspace{5pt}
        \includegraphics[width=0.98\linewidth]{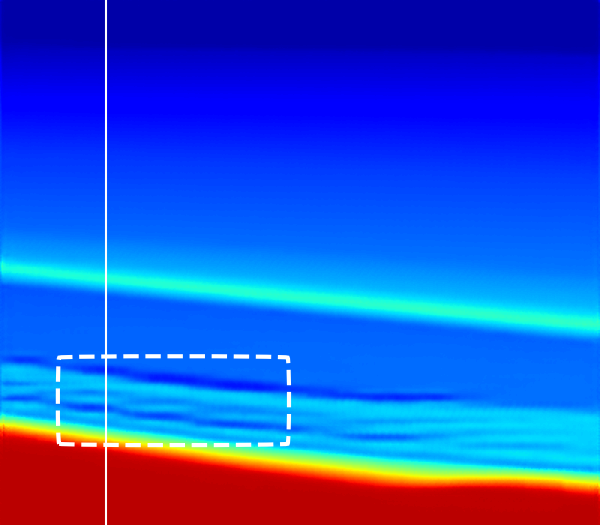} \\ \vspace{5pt}
        \includegraphics[width=0.98\linewidth]{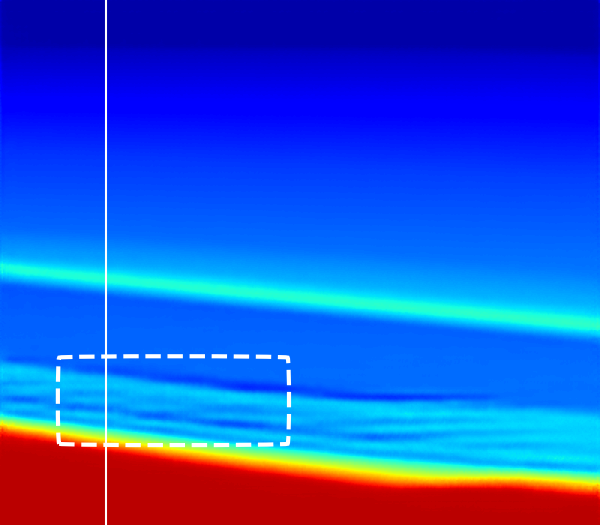}  \\\vspace{5pt}
        \includegraphics[width=0.98\linewidth]{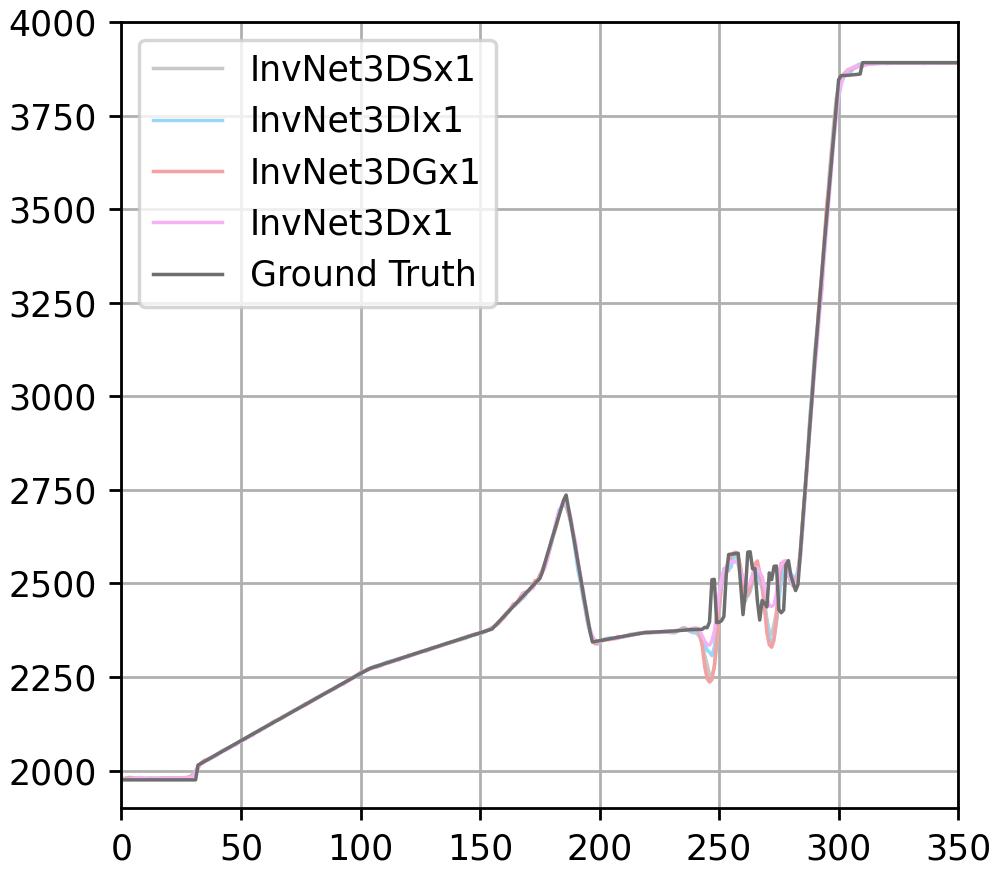} 
        \label{fig:vis-baseline-1}
        \end{minipage}
    }
    \subfloat[]{
        \begin{minipage}{0.18\linewidth}
        \centering
        \includegraphics[width=0.95\linewidth]{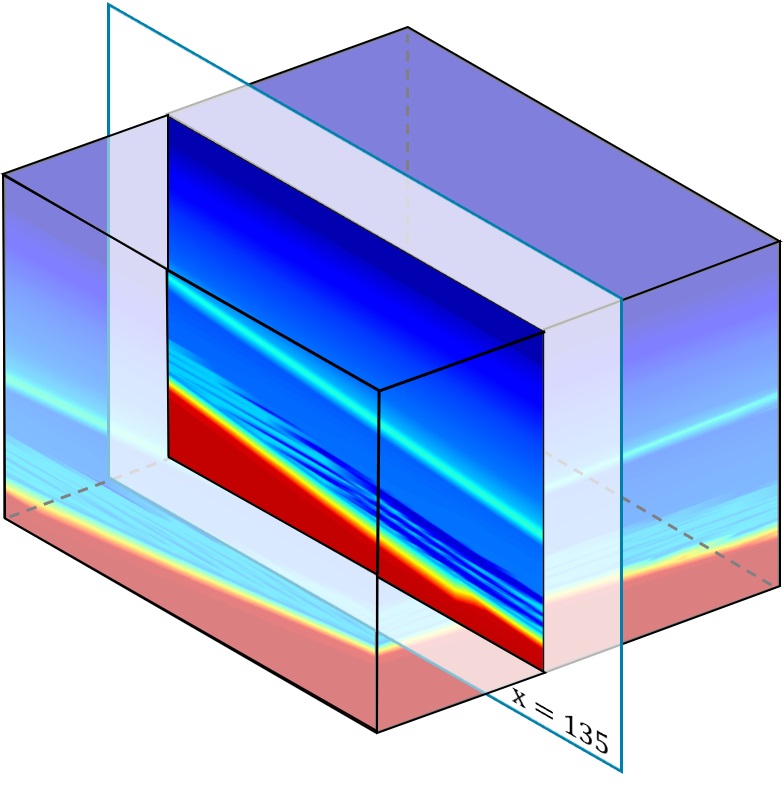}  \\\vspace{5pt}
        \includegraphics[width=0.98\linewidth]{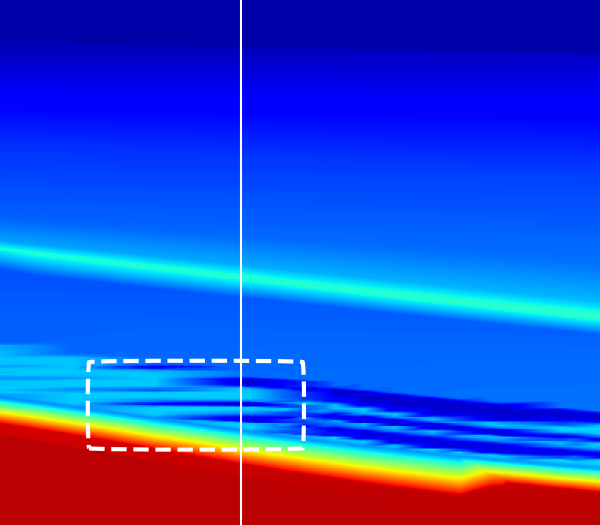}  \\\vspace{5pt}
        \includegraphics[width=0.98\linewidth]{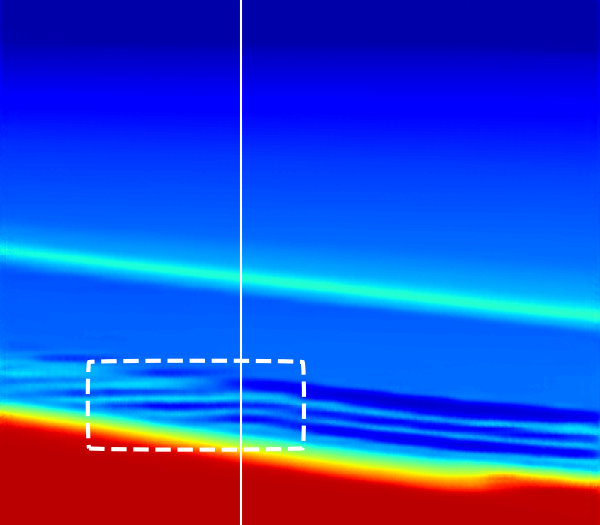}  \\\vspace{5pt}
        \includegraphics[width=0.98\linewidth]{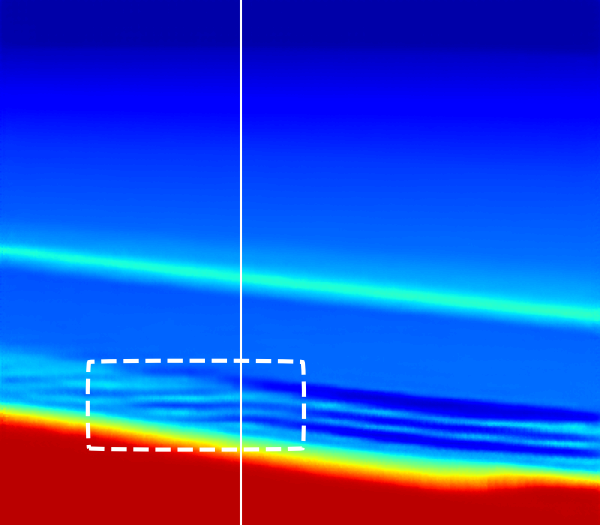}  \\\vspace{5pt}
        \includegraphics[width=0.98\linewidth]{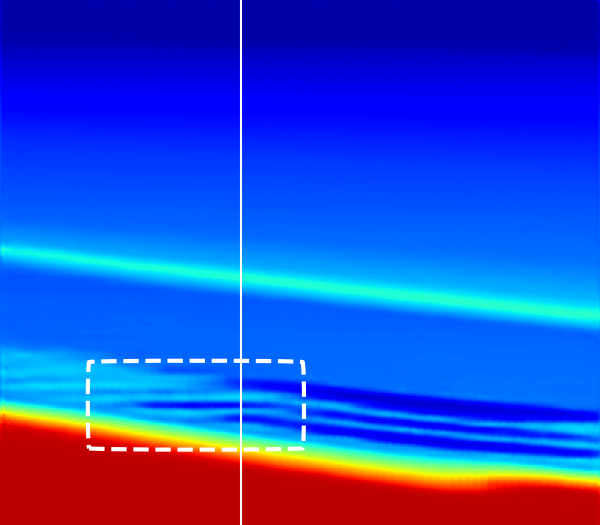}  \\\vspace{5pt}
        \includegraphics[width=0.98\linewidth]{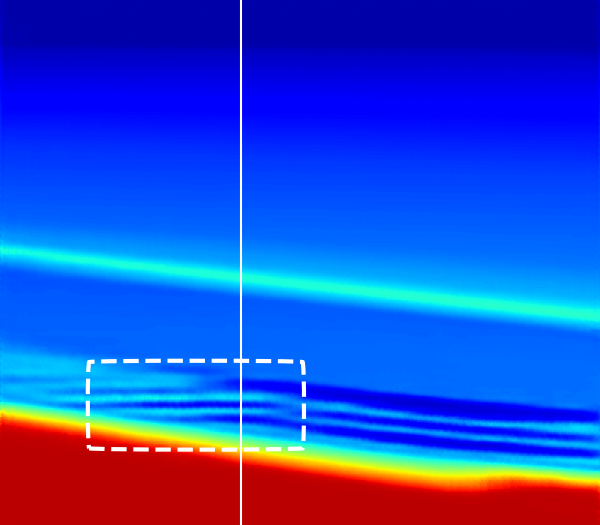} \\\vspace{5pt}
        \includegraphics[width=0.98\linewidth]{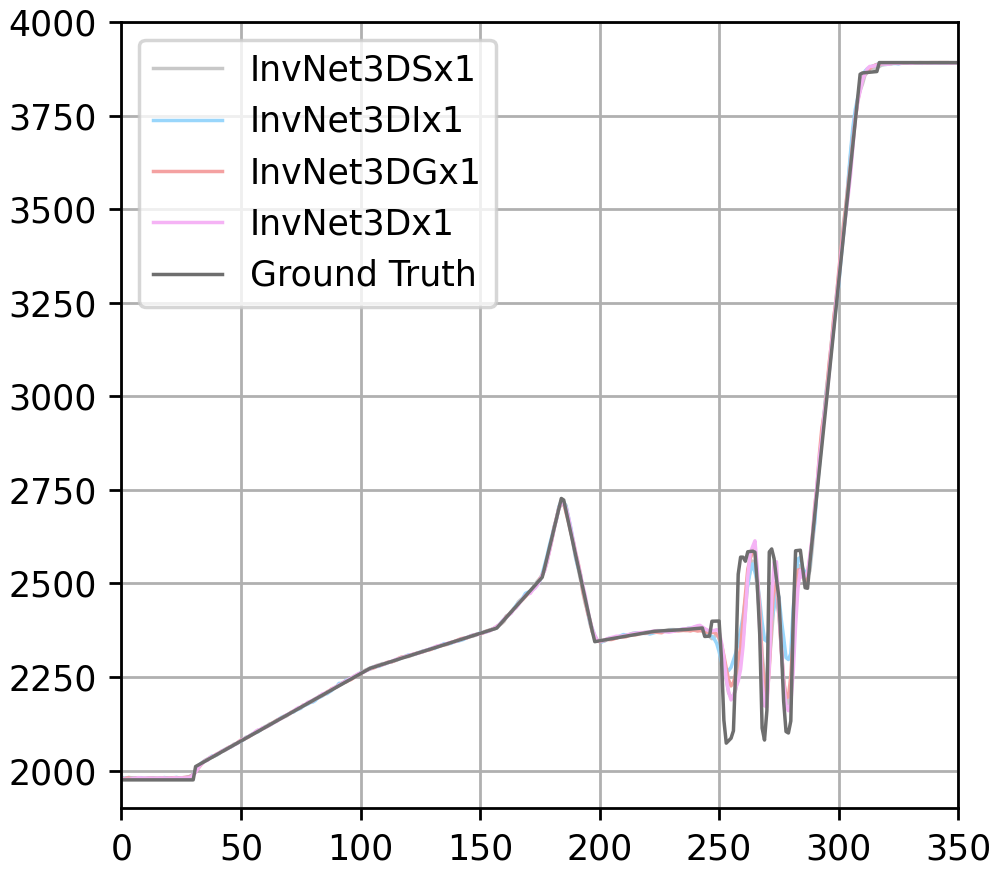} 
        \label{fig:vis-baseline-2}
        \end{minipage}
    }
    \subfloat[]{
        \begin{minipage}{0.18\linewidth}
        \centering
        \includegraphics[width=0.95\linewidth]{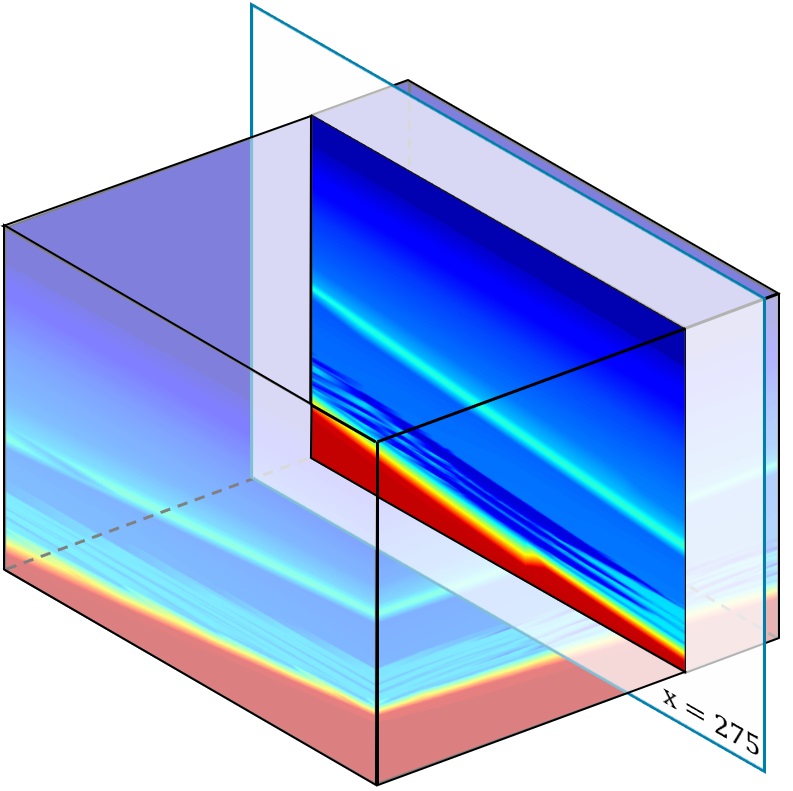}   \\\vspace{5pt}
        \includegraphics[width=0.98\linewidth]{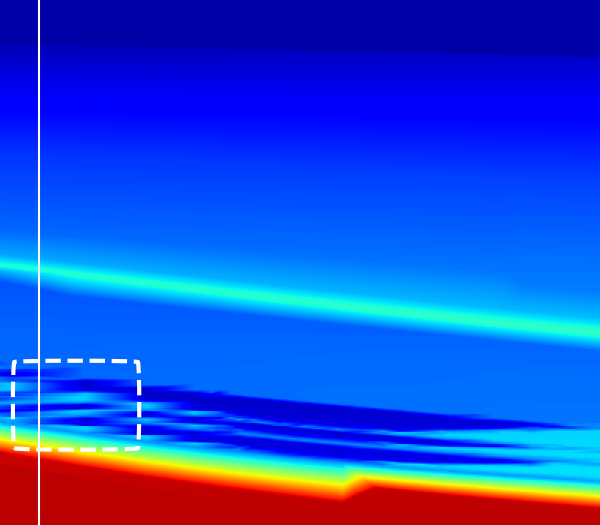}   \\\vspace{5pt}
        \includegraphics[width=0.98\linewidth]{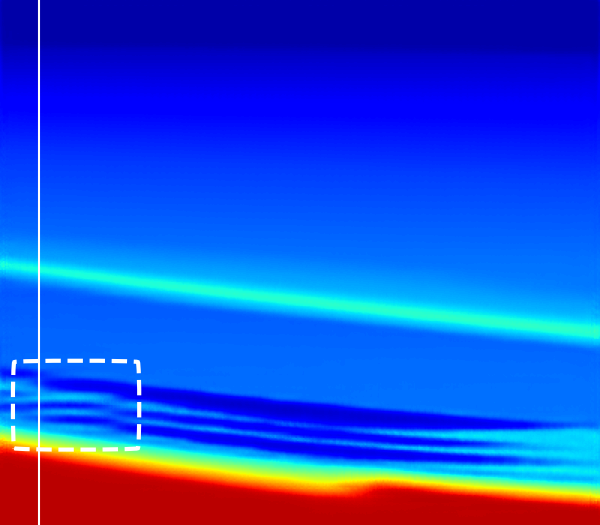}  \\\vspace{5pt}
        \includegraphics[width=0.98\linewidth]{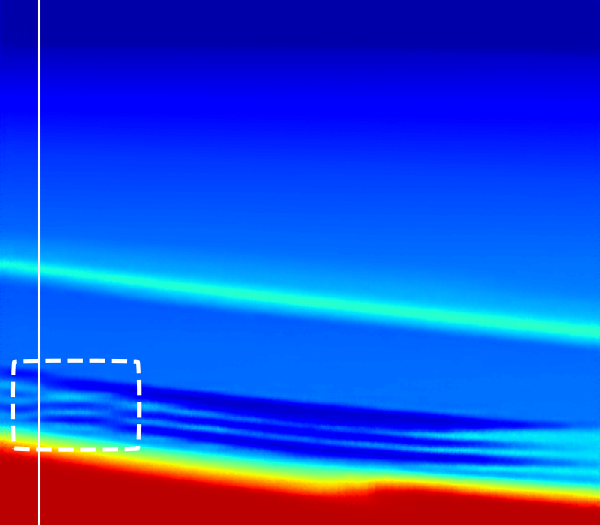}  \\\vspace{5pt}
        \includegraphics[width=0.98\linewidth]{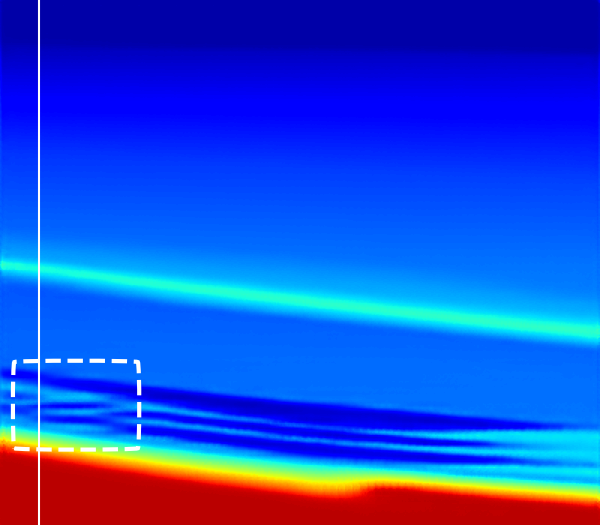}  \\\vspace{5pt}
        \includegraphics[width=0.98\linewidth]{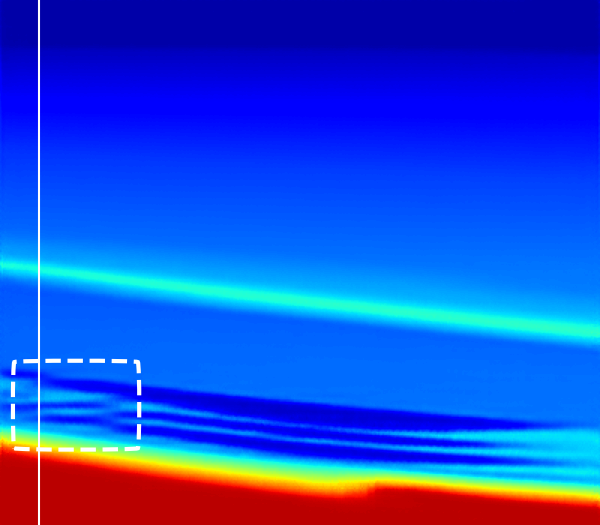} \\\vspace{5pt}
        \includegraphics[width=0.98\linewidth]{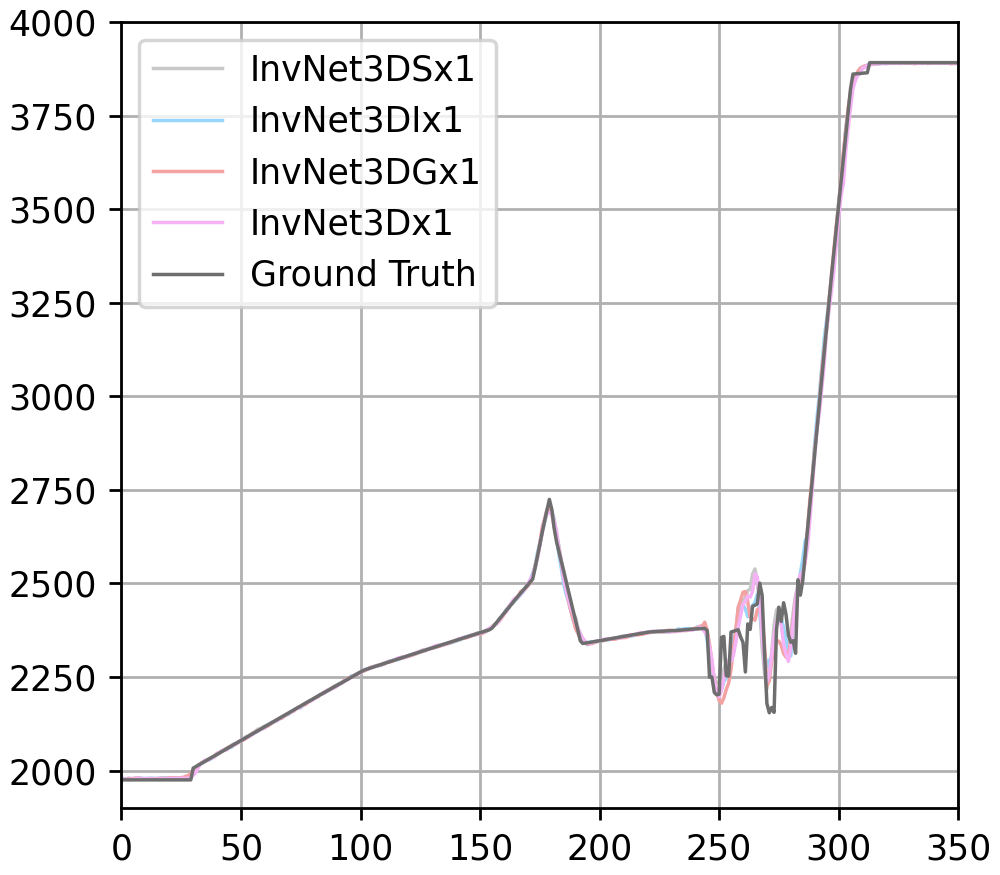}
        \label{fig:vis-baseline-3}
        \end{minipage}
    }
    \subfloat[]{
        \begin{minipage}{0.18\linewidth}
        \centering
        \includegraphics[width=0.99\linewidth]{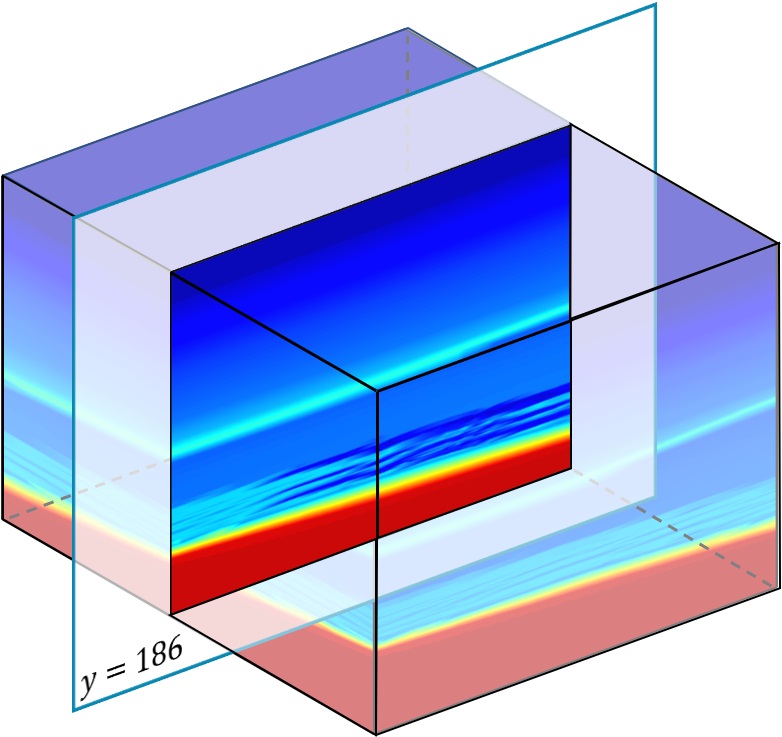}  \\\vspace{5pt}
        \includegraphics[width=0.98\linewidth]{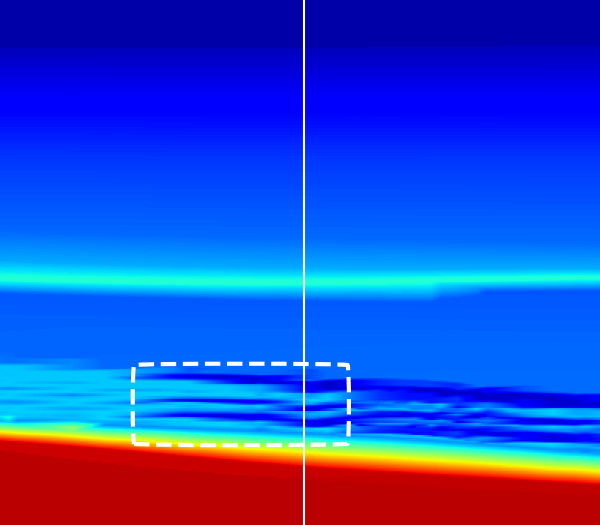}  \\\vspace{5pt}
        \includegraphics[width=0.98\linewidth]{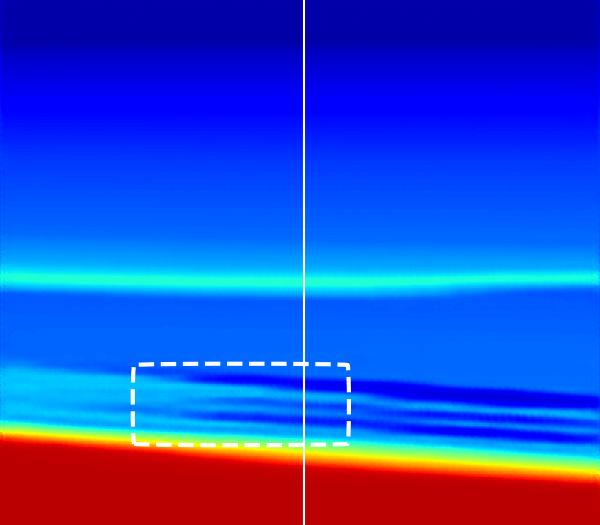}  \\\vspace{5pt}
        \includegraphics[width=0.98\linewidth]{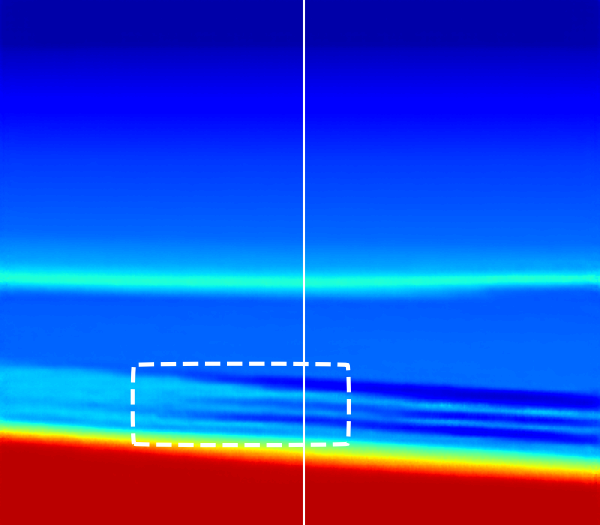}  \\\vspace{5pt}
        \includegraphics[width=0.98\linewidth]{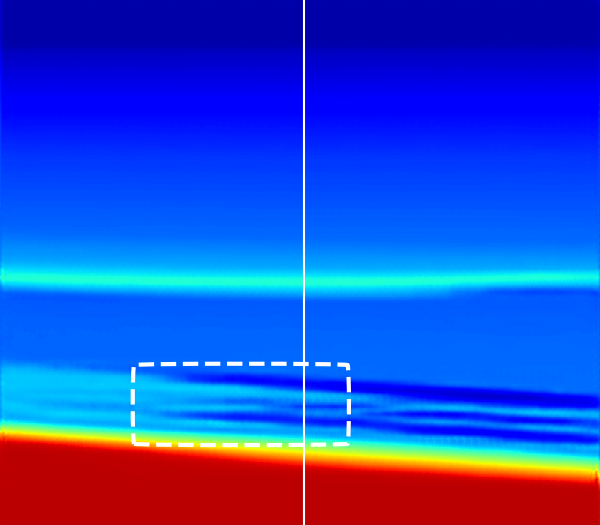}  \\\vspace{5pt}
        \includegraphics[width=0.98\linewidth]{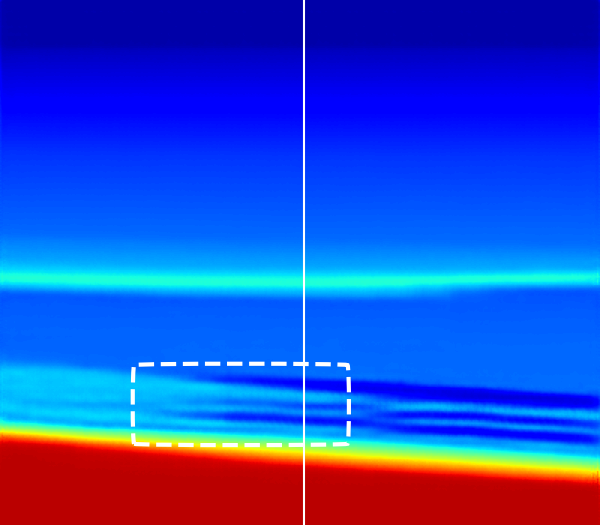} \\\vspace{5pt}
        \includegraphics[width=0.98\linewidth]{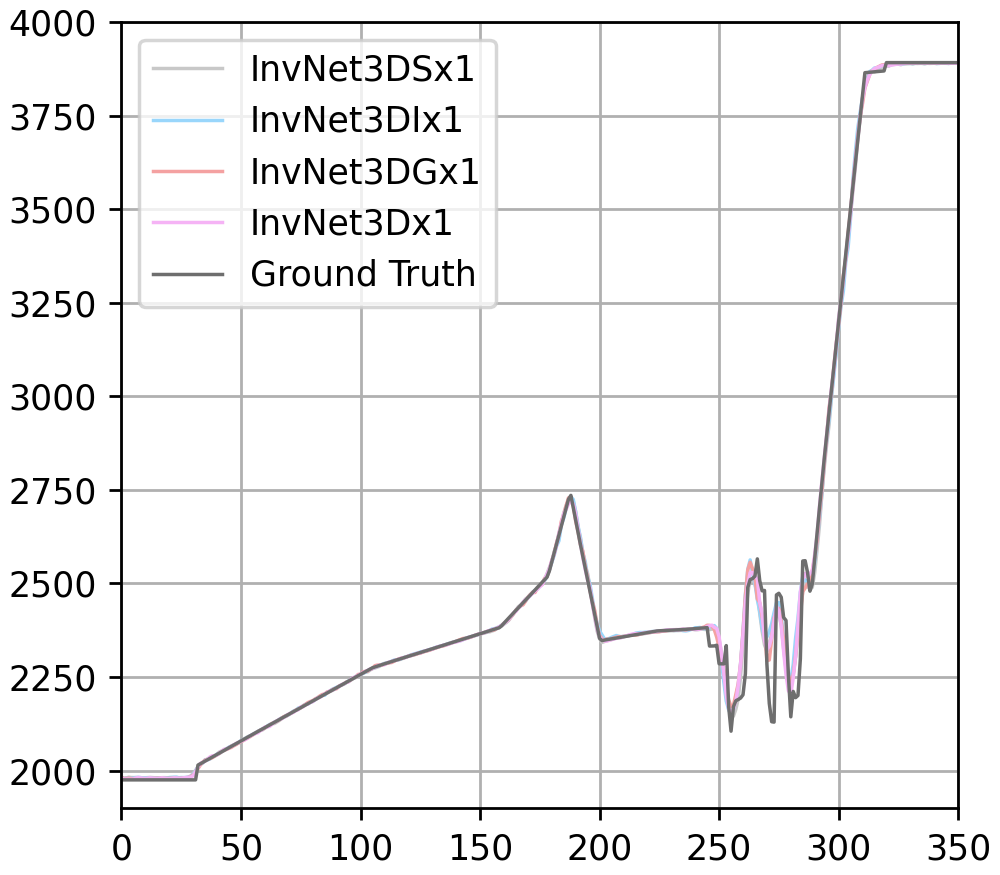}
        \label{fig:vis-baseline-4}
        \end{minipage}
    }
    \subfloat[]{
        \begin{minipage}{0.18\linewidth}
        \centering
        \includegraphics[width=0.98\linewidth]{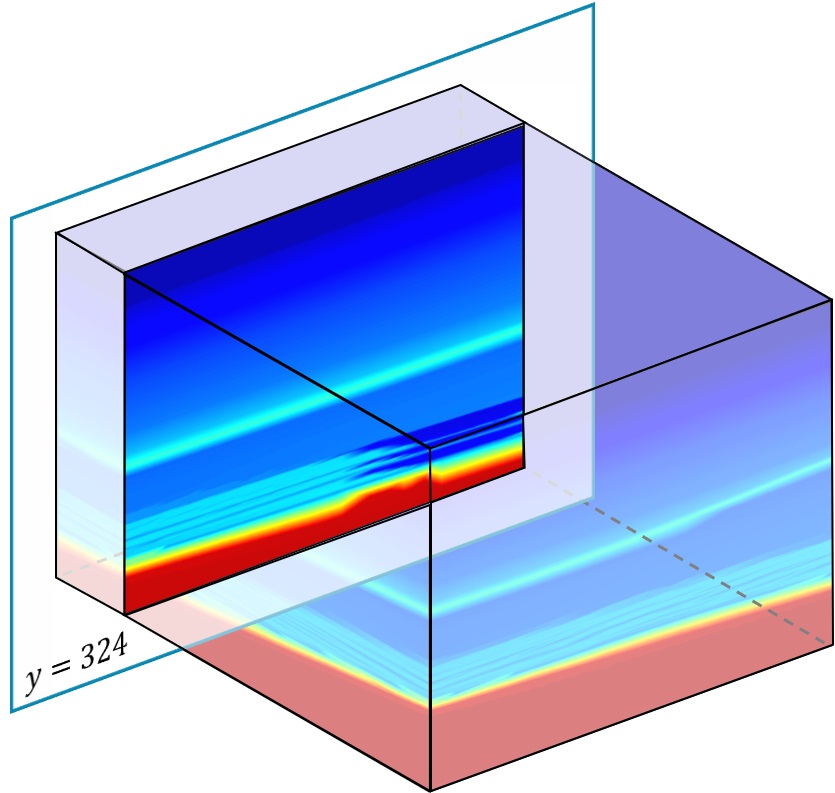}\\ \vspace{5pt}
        \includegraphics[width=0.98\linewidth]{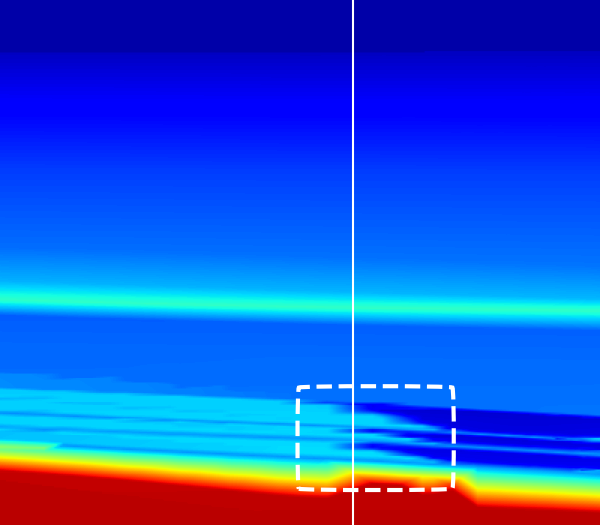} \\ \vspace{5pt}
        \includegraphics[width=0.98\linewidth]{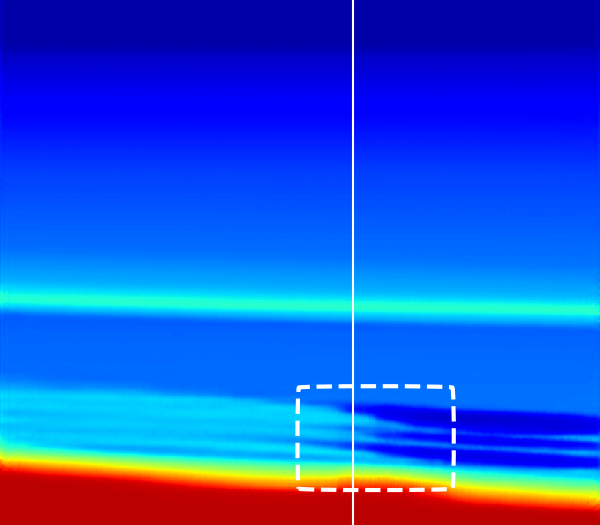}  \\\vspace{5pt}
        \includegraphics[width=0.98\linewidth]{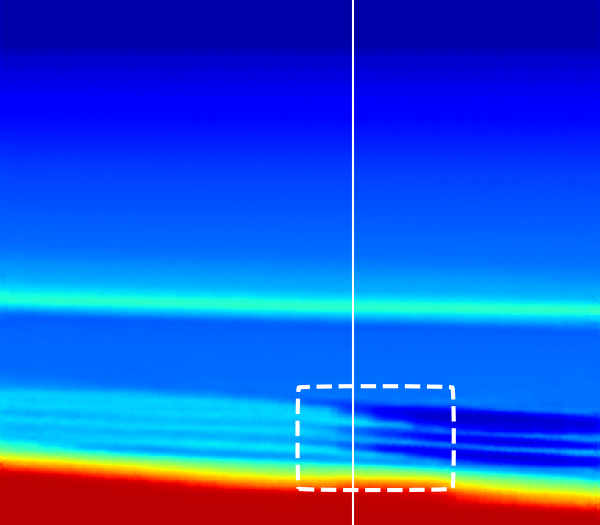}  \\\vspace{5pt}
        \includegraphics[width=0.98\linewidth]{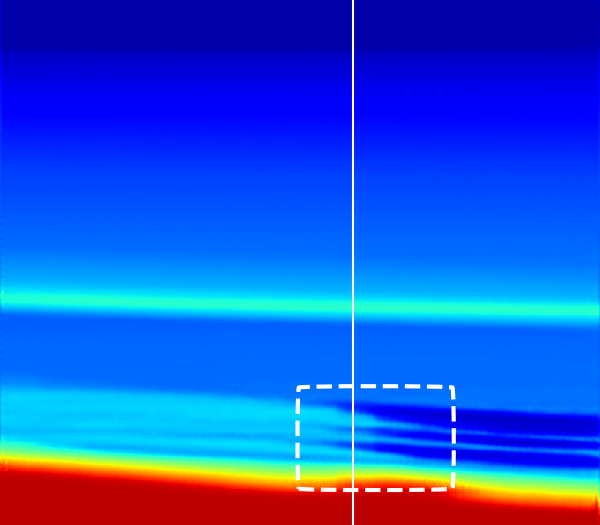}  \\\vspace{5pt}
        \includegraphics[width=0.98\linewidth]{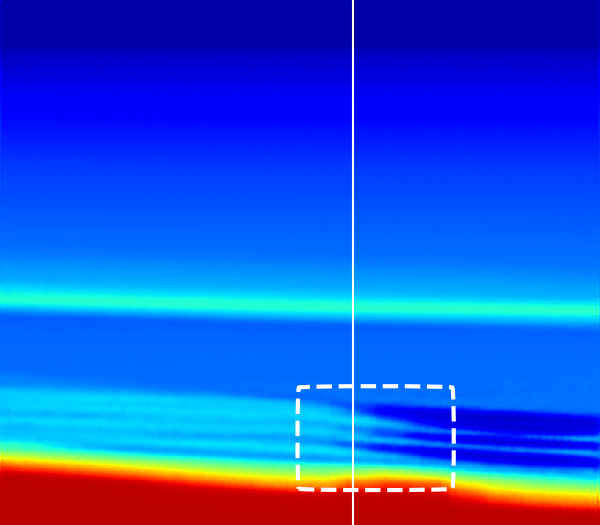} \\\vspace{5pt}
        \includegraphics[width=0.98\linewidth]{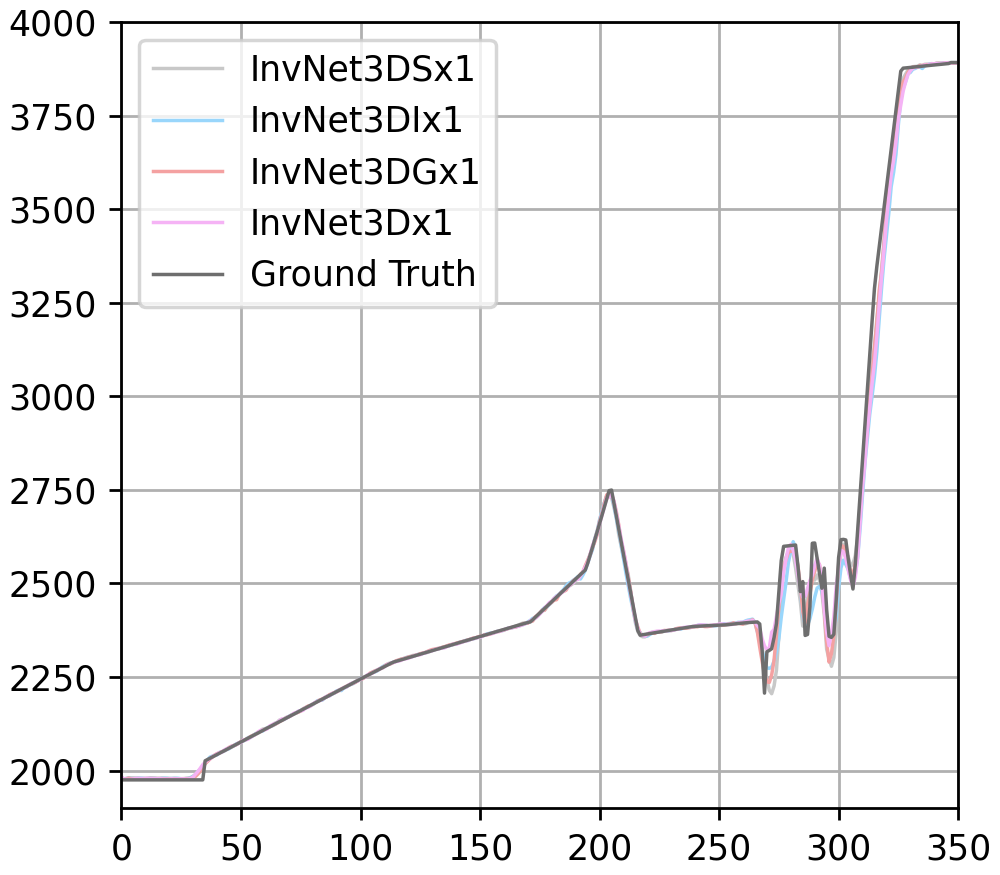}
        \label{fig:vis-baseline-5}
        \end{minipage}
    }
    \begin{minipage}{0.06\linewidth}
    \centering
    \includegraphics[width=0.95\linewidth]{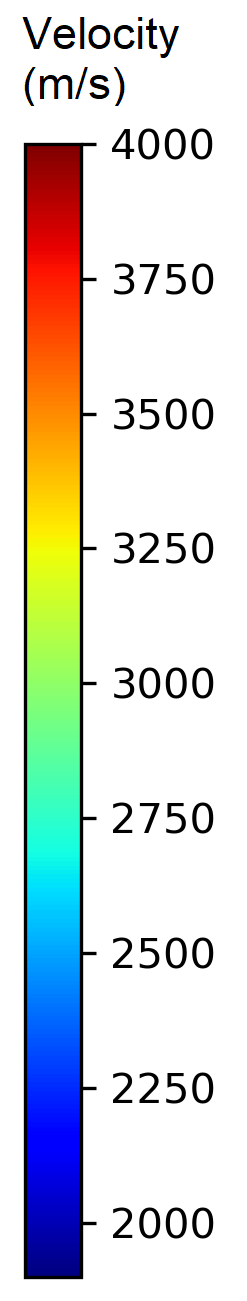}
    \end{minipage}
    
    \begin{minipage}{0.95\linewidth}
    \centering
    \caption{\textbf{Visualization of ground truth 3D velocity map (first row), a sampled 2D slice for comparison (second row), corresponding 2D slices extracted from predicted velocity map generated by InvNet3DS, InvNet3DI, InvNet3DG, and InvNet3D (third row to sixth row).} The seventh row displays vertical velocity profiles (velocity - depth) at the location of the white cursor on the 2D slices. Rectangles with white dashed borderline highlight the region of interest in comparison. All the models are built with 26 layers. Each column presents one sample. Zoom-in visualization of the highlighted areas can be found in the supplementary materials.}
    \label{fig:vis-baseline}
    \end{minipage}
\end{figure*}

We introduced two components, Channel-Separated Encoder and Invertible Module, to improve the baseline, InvNet3DS. The proposed InvNet3D is equipped with both components. For a deeper analysis of the impact of these components, our experiments also involve the comparison with models inserted with either of them. We denote the model with only a Channel-Separated Encoder as InvNet3DG and the model with only Invertible Modules as InvNet3DI. Table \ref{tab:network-name} provides a summary for the naming convention. In this experiment, for a fair comparison, all the models have the same number of layers as the baseline model InvNet3DS, described in Table \ref{tab:invnet3d-simple-arch}, i.e., 13 layers in encoder and 13-layer in decoder. Therefore, the number of layers in Invertible Modules is set to be $1$ for InvNet3DI and InvNet3D.

Table \ref{tab:main-comparison} shows the evaluation results. It is clear that InvNet3D outperforms InvNet3DS. The former is also 60\% smaller and 10.7\% faster than the latter. The results reported in this table can be further analyzed along two axes.

\begin{description}[font=\rmfamily, leftmargin=0pt, style=sameline]
  \item[The impact of Channel-Separated Encoder.] Comparing InvNet3DS and InvNet3DI with InvNet3DG and InvNet3D, performance measured by all the three metrics is improved due to the use of group convolution. This improvement can probably be attributed to the reduced number of pixels with high prediction error, as displayed in Figure \ref{fig:error-hist}. Models with Channel-Separated Encoder obviously have a much smaller maximum pixel-wise error compared to their counterparts with an ordinary encoder, which demonstrates that the former models are better at utilizing multiple seismic data records and generating representations of higher quality. Furthermore, the former ones are also much lighter-weight and computationally efficient due to smaller filters in the group convolution.
  \item[The impact of Invertible Module.] It can be observed that there are only minor differences between the performance of the models with invertible layers and that of those without invertible layers, which demonstrates that Invertible Modules, though learning representations in a workflow other than conventional convolution, would not heavily influence the capacity of models. Considering invertible layers implicitly introduce group convolution with a group size of 2, we believe the inferior performance of InvNet3DI compared to InvNet3DS results from such a non-ideal way of channel separation. In the comparison between InvNet3DG and InvNet3D, the influence of non-ideal channel separation  becomes very insignificant since invertible layers in InvNet3D are enforced to consist of two convolutional layers with a group size of 4 in order to be compatible with the overall framework built upon group convolution with a group size of 8.
\end{description}

The visualizations of the generated velocity maps provided in Figure \ref{fig:vis-baseline} intuitively illustrate the impact of the proposed components. These displayed samples mainly aim at comparing the models' ability to reconstruct the detailed subsurface structure. For a better visibility, we enclose the distinct region of interest by rectangles with a white dashed borderline. It is worth mentioning that within this region, there are several low-velocity reflectors, which correspond to the shale layers existed in the actual geologic formation at the Kimberlina site~\cite{wagoner2009}.   

\subsubsection{Experiments with deeper networks}
\label{sec:n-blocks}
While deeper models can be expected to reconstruct velocity maps of higher quality due to the increased model capacity, there is an associated computational cost because the memory consumption by intermediate representations grows linearly with the number of layers.
Invertible networks, whose memory usage is irrelevant to the volume of activations, can be made very deep if needed. The same is also true for partially reversible networks like InvNet3D. For a demonstration of the benefits of invertibility and deeper networks, we expand InvNet3D by stacking more layers in its Invertible Module. More precisely, we train and test InvNet3D models with the number of layers ($\mathrm{\#Blocks}$) in all Invertible Modules of 2, 3, and 4, respectively and compare them with the minimum version with $\mathrm{\#Blocks}=1$. Accordingly, we also establish expanded InvNet3DG models, which employ Channel-Separated Encoder but are not partially reversible, for a fair comparison. In the following experiments, we denote these models by "[model]x[$\mathrm{\#Blocks}$]". 

\begin{figure*}
\begin{minipage}{0.34\linewidth}
\centering
    \centering
    \includegraphics[width=0.9\linewidth]{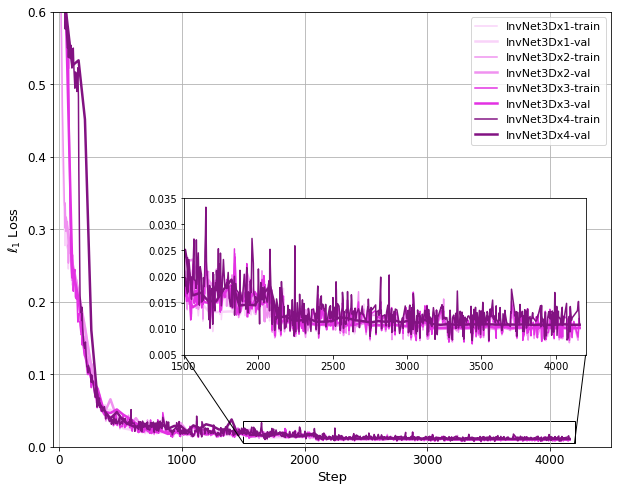}
    \begin{minipage}{0.9\linewidth}
    \centering
    \caption{\textbf{Training and validation $\ell_1$ loss of InversionNet3D with \#blocks varying from one to four.} Separated loss curves for each model can be found in the supplementary materials.}
    \label{fig:block-loss}
    \end{minipage}
\end{minipage}
\begin{minipage}{0.65\linewidth}
\centering

    \begin{minipage}{0.98\linewidth}
    \centering
    \captionof{table}{\textbf{Performance Comparison between InvNet3DG and InvNet3D.} Memory column presents maximum memory consumption in training stage.}
    \label{tab:block-comparison}
    \end{minipage}
\footnotesize
    \renewcommand\arraystretch{1.2}
    \begin{tabular}{c|c|c|c|c|c|c|c}
    \hline
    Model                       & \#Blocks & \#Params & GFLOPs   & Memory \footnote{Maximum memory consumption is obtained via \texttt{torch.cuda.max\_memory\_allocated()}, 
    which provides statistics of memory used by tensors. Actual memory usage would be larger than the reported value due to memory reserved by caching memory allocator and context manager.} & MAE $\downarrow$    & RMSE $\downarrow$   & SSIM $\uparrow$ \\ \hline
    \multirow{4}{*}{InvNet3DG}  & 1        & 15.60M   & 2760.88  & 7.50GB & 9.82   & 26.00  & 0.9831 \\
                                & 2        & 17.99M   & 2824.14  & 9.77GB & \textbf{9.45}   & \textbf{25.43}  & \textbf{0.9841} \\
                                & 3        & 20.38M   & 2885.42  & 12.03GB & 11.25  & 32.55  & 0.9798 \\
                                & 4        & 22.77M   & 2946.68  & 14.31GB  & -      & -      & -\\ \hhline{=:=:=:=:=:=:=:=}
    \multirow{4}{*}{InvNet3D}   & 1        & 14.42M   & 2734.54  & 9.93GB   & 9.83   & 26.11  & 0.9826 \\
                                & 2        & 15.63M   & 2767.48  & 9.93GB   & \textbf{9.52}   & \textbf{25.35}  & \textbf{0.9838} \\
                                & 3        & 16.85M   & 2800.40  & 9.95GB   & 9.74   & 26.25  & 0.9826 \\
                                & 4        & 18.06M   & 2833.34  & 9.98GB   & 10.33  & 28.01  & 0.9804\\ \hline
    \end{tabular}
\end{minipage}
\end{figure*}

Table \ref{tab:block-comparison} shows the evaluation results. Since P100-16GB GPUs  do not support the training of InvNet3DGx4 due to memory limitations, the model's performance is not reported. In contrast, InvNet3Dx4 can be successfully trained on the same hardware. Although the maximum memory consumption InvNet3Dx1/2 is larger than that of InvNet3DGx1/2 since backward pass in the former requires extra memory for temporary results, deeper-version InvNet3DGs consumes much more memory than InvNet3Ds of the same scale. The growth of peak memory usage of InvNet3D is fairly stable as the extra memory is taken by model parameters instead of activations. Both models achieve the best performance with $\mathrm{\#Blocks}=2$ but start to perform worse when $\mathrm{\#Blocks}$ becomes larger than 2. We believe it is because the limited scale of our dataset (with only 1664 training samples in total) and the relatively homogeneous geologic structure described by these data make it difficult to train very deep networks. Should there be a dataset with more samples and more complex or diversified subsurface layers, the advantage of deeper networks could be better demonstrated and the tendency of improvement could continue with $\mathrm{\#Blocks}\textgreater2$. That being said, InvNet3Dx3 still outperforms InvNet3DGx3, which should result from the reduced $\mathrm{\#Params}$ due to the use of Invertible Module.

\begin{figure*}[]
\centering
    \subfloat[]{
        \begin{minipage}{0.18\linewidth}
        \centering
        \includegraphics[width=0.98\linewidth]{Figure/vis_masked/year30_cut37_x108/4-InvNet3DSRCx1.png} \\\vspace{5pt}
        \includegraphics[width=0.98\linewidth]{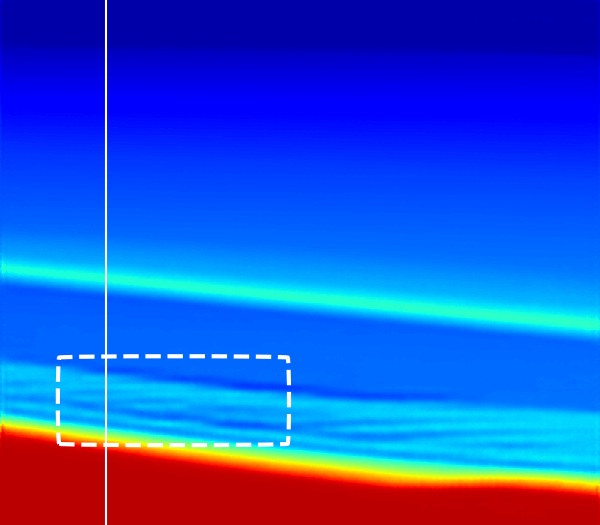} \\\vspace{5pt}
        \includegraphics[width=0.98\linewidth]{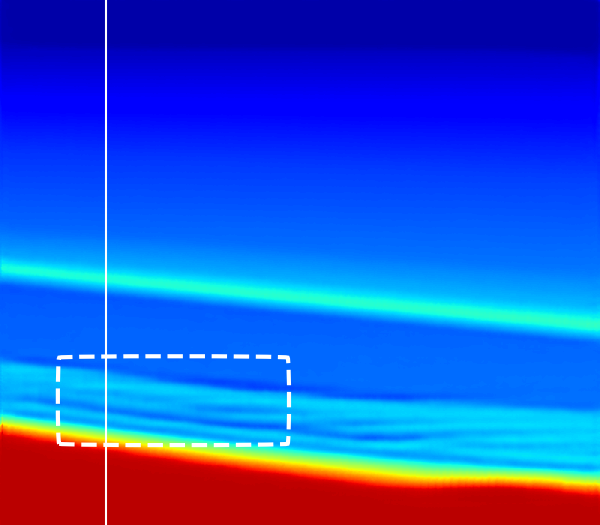} \\\vspace{5pt}
        \includegraphics[width=0.98\linewidth]{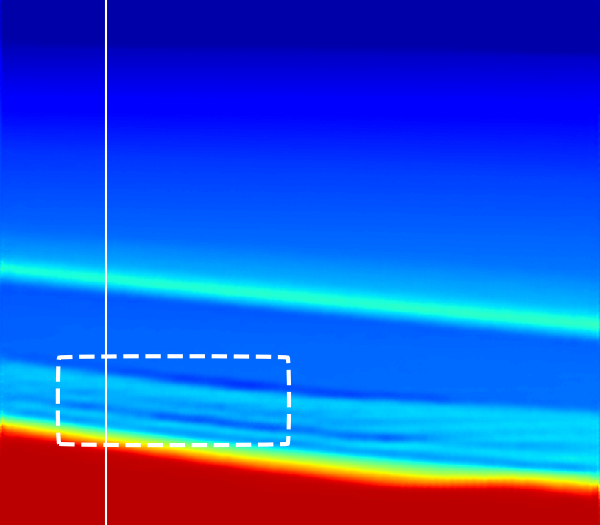} \\\vspace{5pt}
        \includegraphics[width=0.98\linewidth]{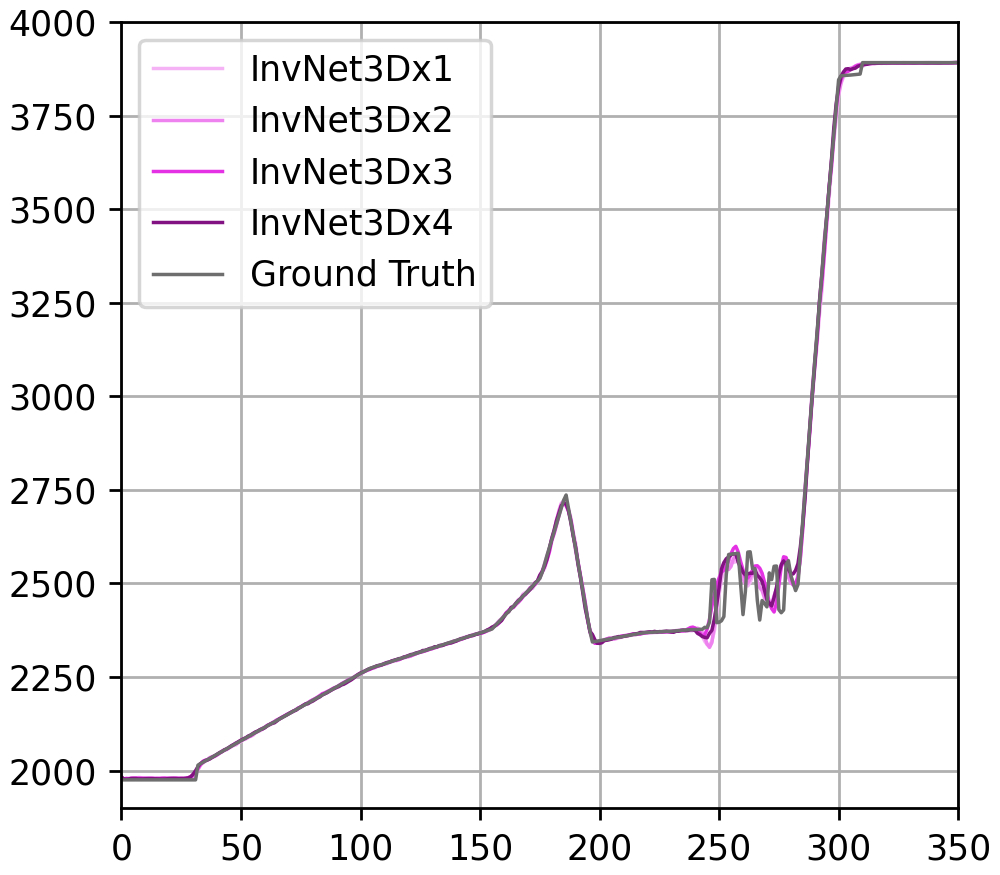}
        \end{minipage}
    }
    \subfloat[]{
        \begin{minipage}{0.18\linewidth}
        \centering
        \includegraphics[width=0.98\linewidth]{Figure/vis_masked/year25_cut7_x135/4-InvNet3DSRCx1.png} \\\vspace{5pt}
        \includegraphics[width=0.98\linewidth]{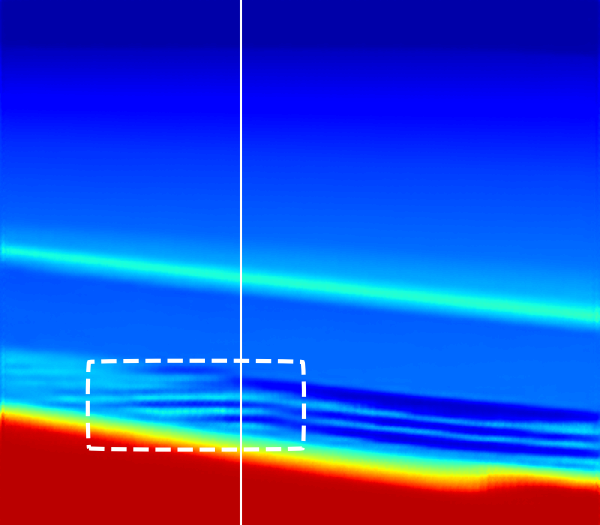} \\\vspace{5pt}
        \includegraphics[width=0.98\linewidth]{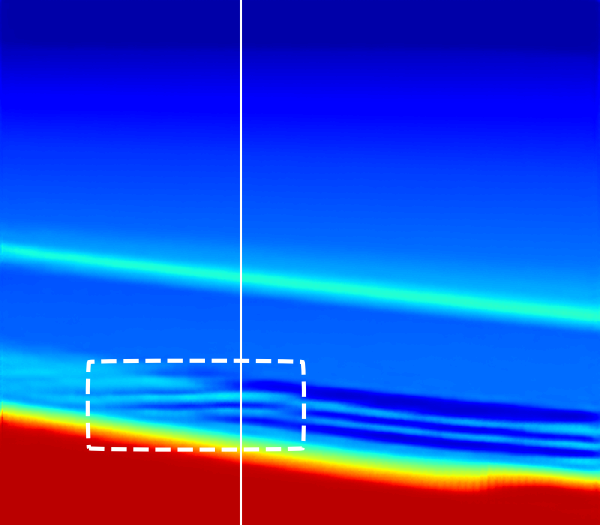} \\\vspace{5pt}
        \includegraphics[width=0.98\linewidth]{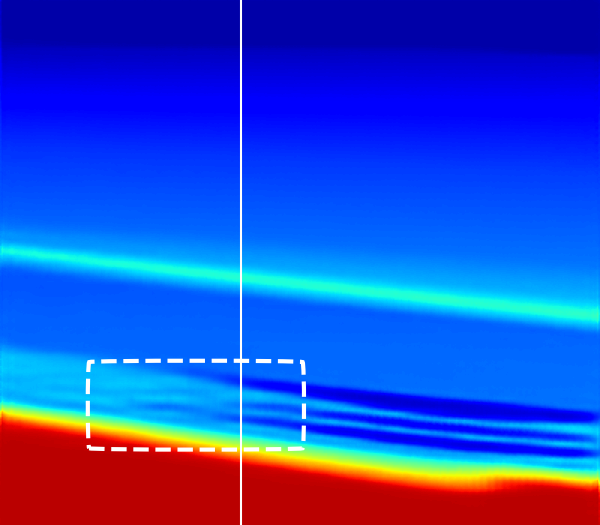} \\ \vspace{5pt}
        \includegraphics[width=0.98\linewidth]{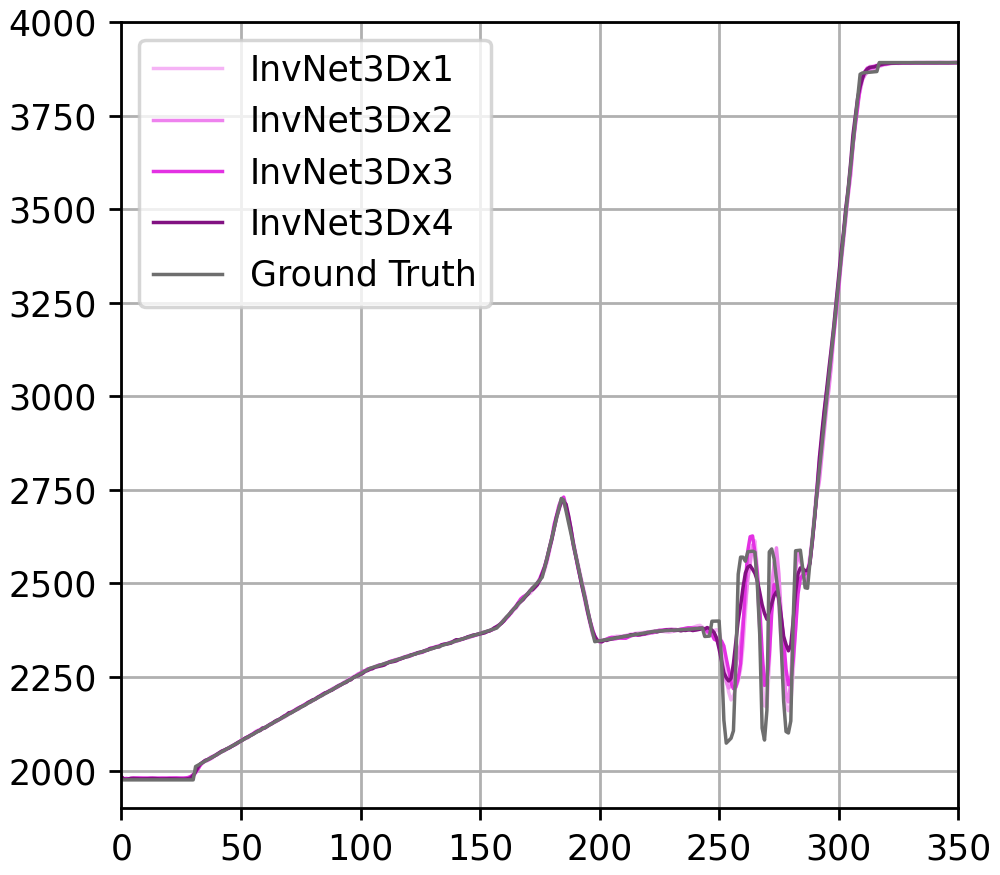}
        \end{minipage}
    }
    \subfloat[]{
        \begin{minipage}{0.18\linewidth}
        \centering
        \includegraphics[width=0.98\linewidth]{Figure/vis_masked/year30_cut37_x275/4-InvNet3DSRCx1.png} \\\vspace{5pt}
        \includegraphics[width=0.98\linewidth]{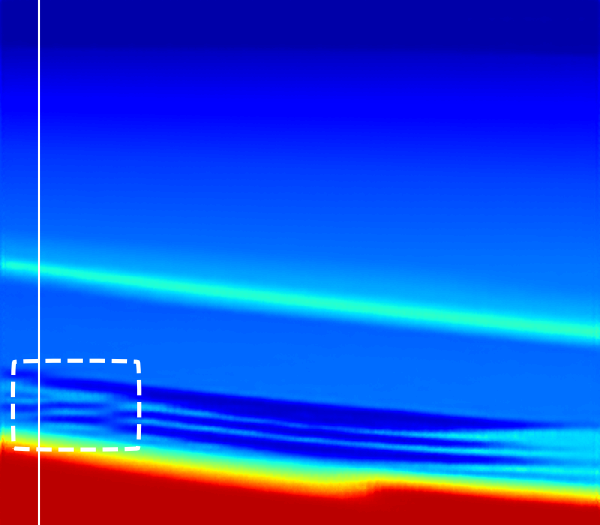} \\\vspace{5pt}
        \includegraphics[width=0.98\linewidth]{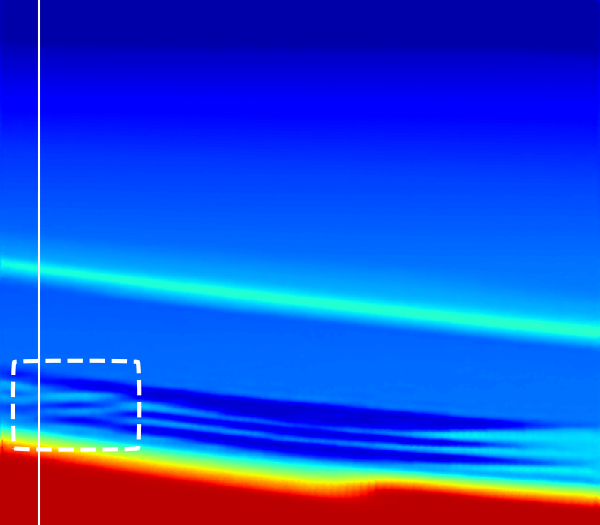} \\\vspace{5pt}
        \includegraphics[width=0.98\linewidth]{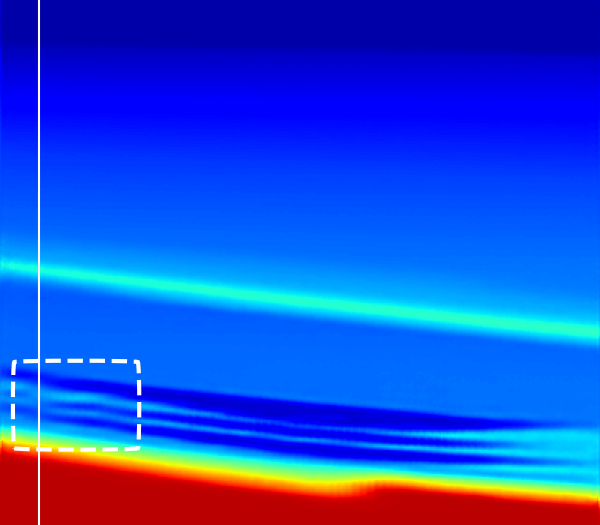} \\ \vspace{5pt}
        \includegraphics[width=0.98\linewidth]{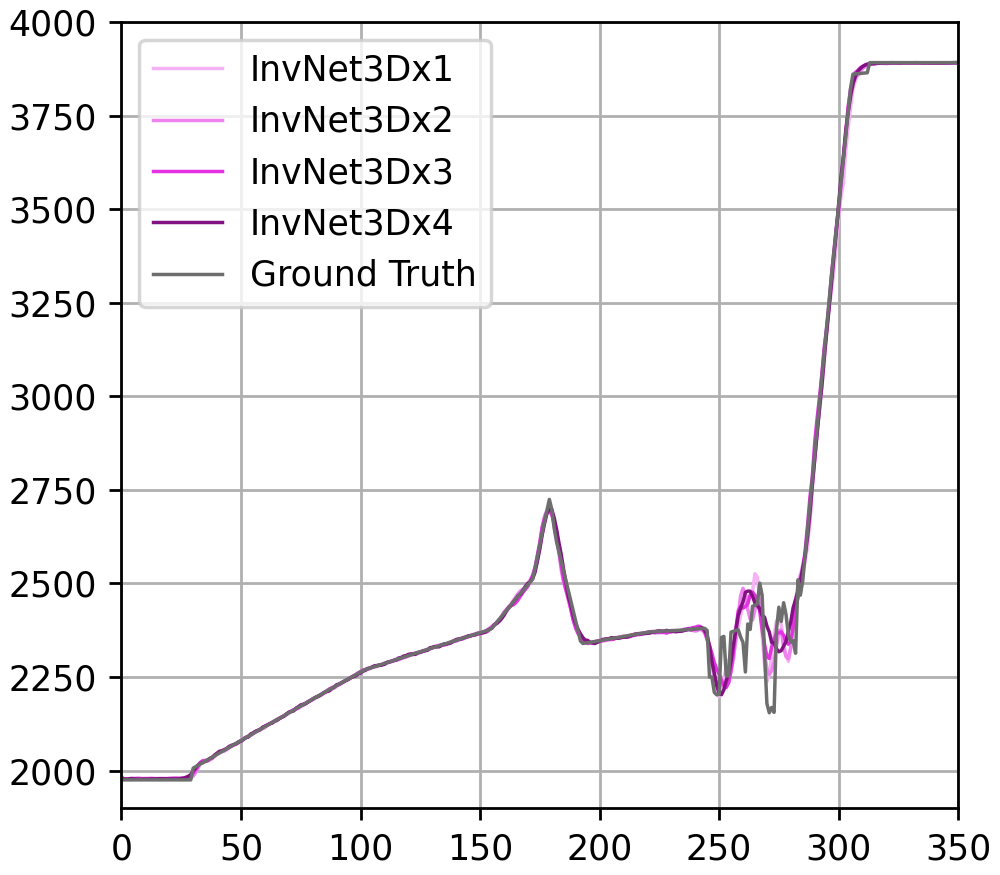}
        \end{minipage}
    }
    \subfloat[]{
        \begin{minipage}{0.18\linewidth}
        \centering
        \includegraphics[width=0.98\linewidth]{Figure/vis_masked/year30_cut42_y186/4-InvNet3DSRCx1.png} \\\vspace{5pt}
        \includegraphics[width=0.98\linewidth]{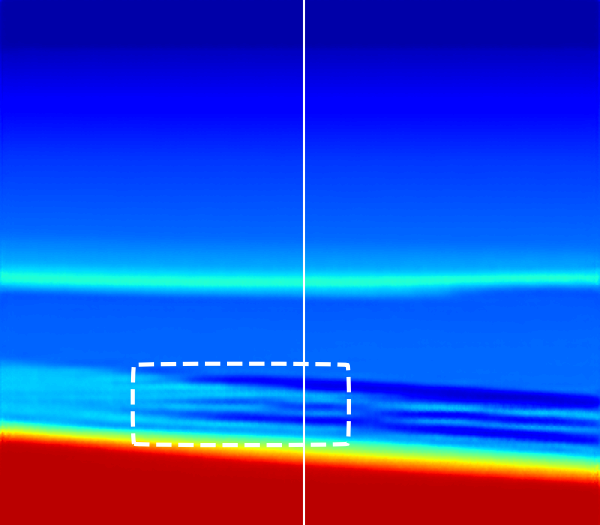} \\\vspace{5pt}
        \includegraphics[width=0.98\linewidth]{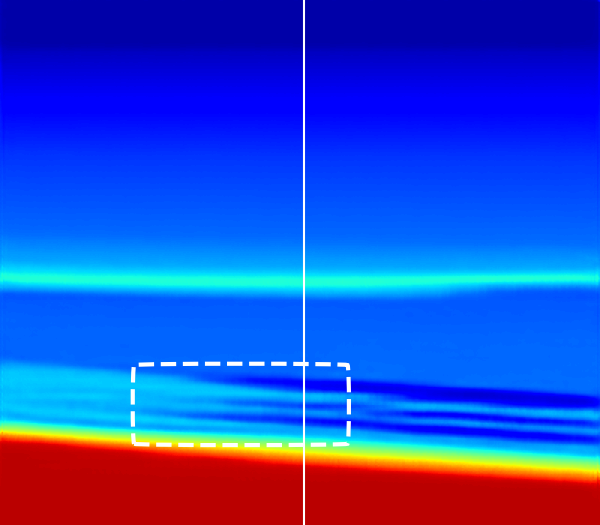} \\\vspace{5pt}
        \includegraphics[width=0.98\linewidth]{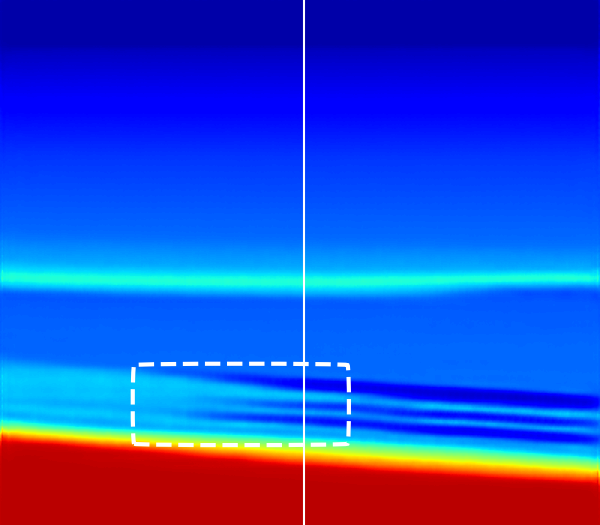} \\\vspace{5pt}
        \includegraphics[width=0.98\linewidth]{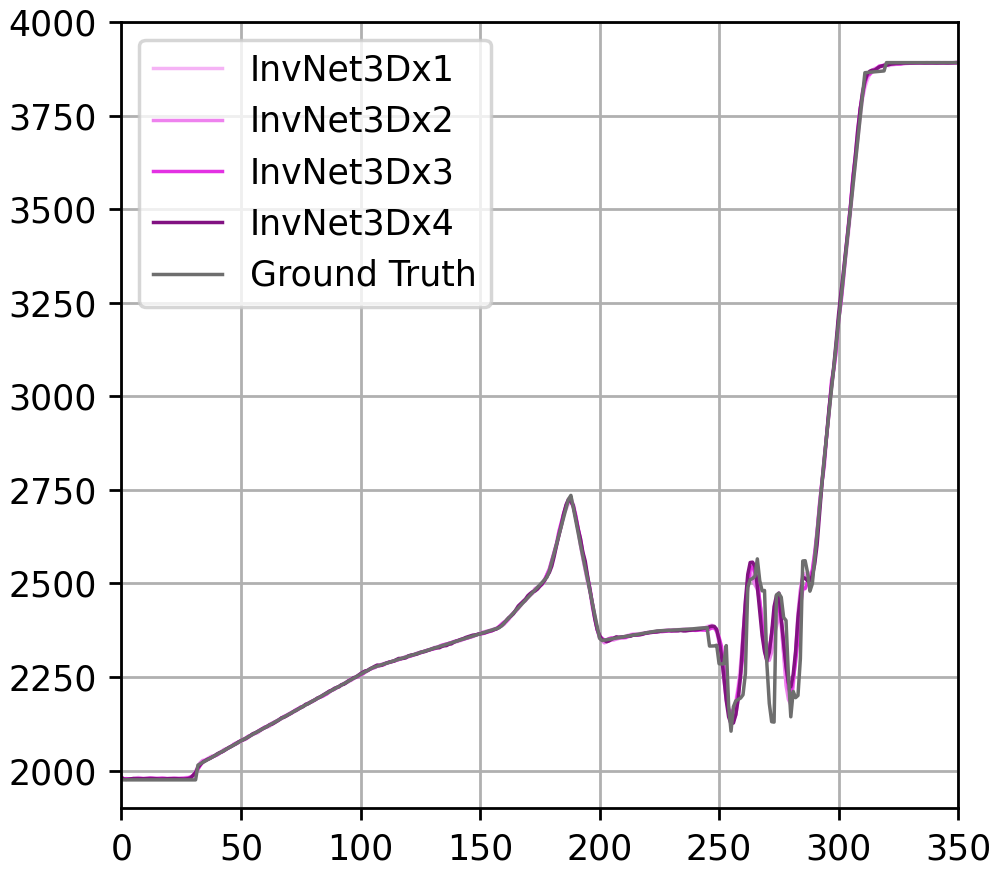}
        \end{minipage}
    }
    \subfloat[]{
        \begin{minipage}{0.18\linewidth}
        \centering
        \includegraphics[width=0.98\linewidth]{Figure/vis_masked/year10_cut48_y324/4-InvNet3DSRCx1.png} \\\vspace{5pt}
        \includegraphics[width=0.98\linewidth]{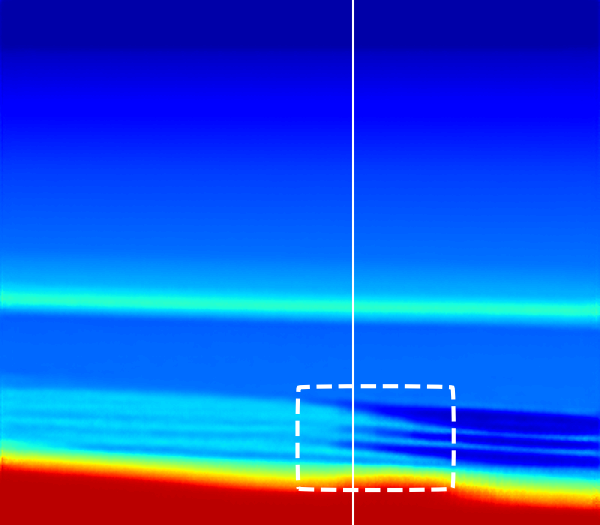} \\\vspace{5pt}
        \includegraphics[width=0.98\linewidth]{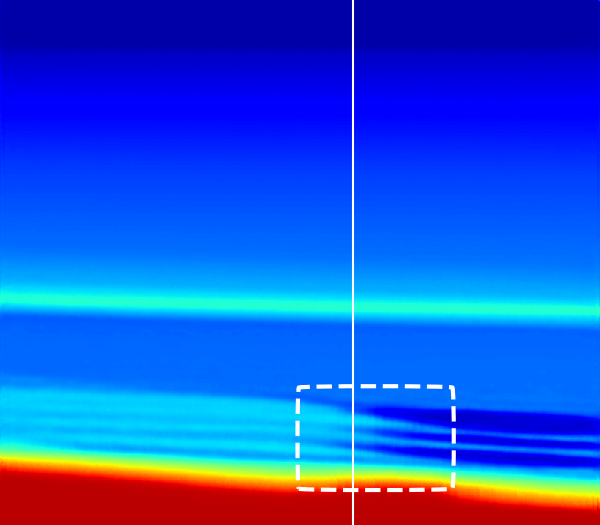} \\\vspace{5pt}
        \includegraphics[width=0.98\linewidth]{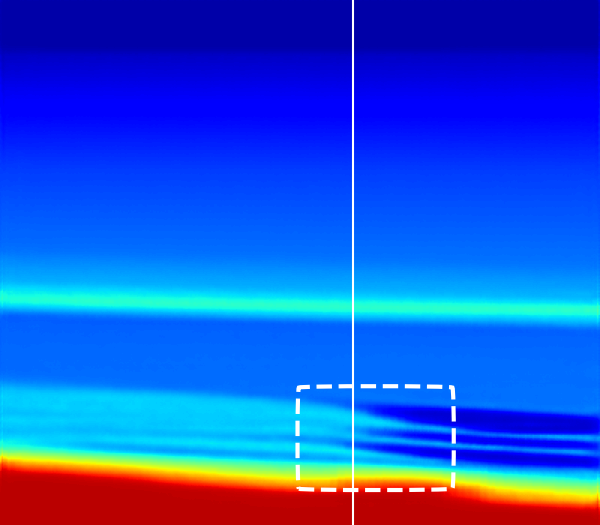} \\\vspace{5pt}
        \includegraphics[width=0.98\linewidth]{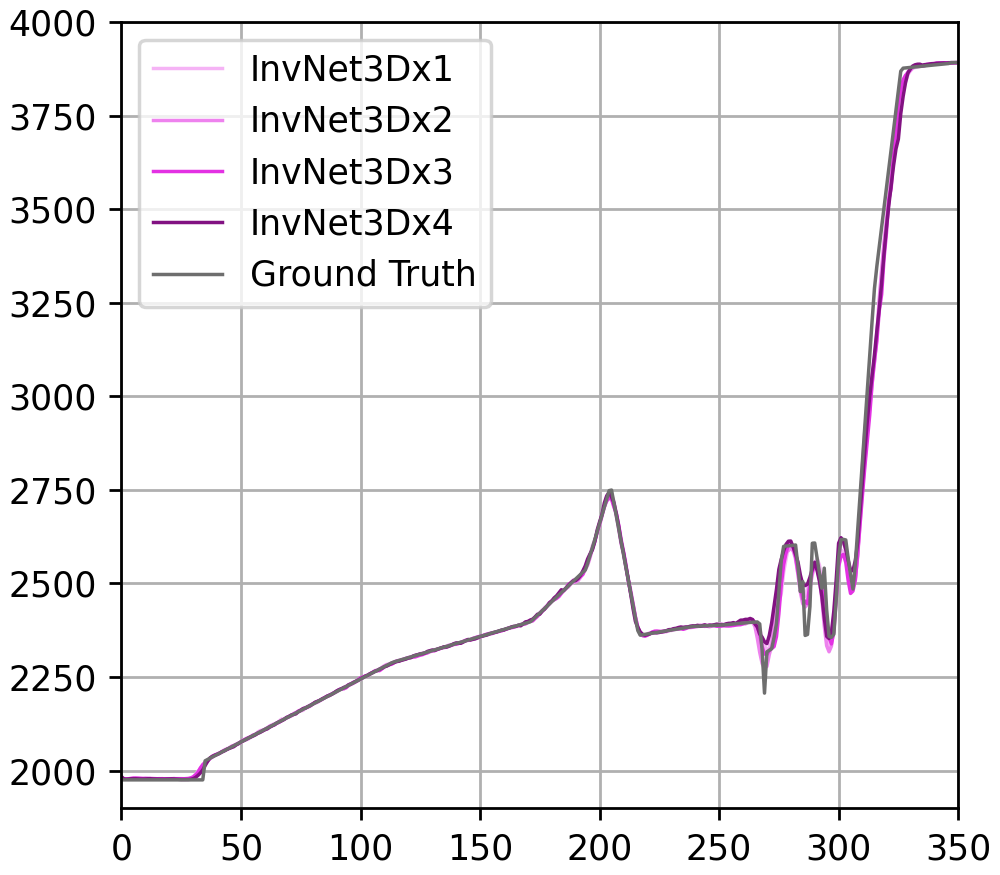} \\
        \end{minipage}
    }
    \begin{minipage}{0.06\linewidth}
    \centering
    \includegraphics[width=0.95\linewidth]{Figure/color_bar.png}
    \end{minipage}
    
    \begin{minipage}{0.97\linewidth}
    \centering
    \caption{\textbf{Visualization of predicted velocity map samples generated by InvNet3Dx1, InvNet3Dx2, InvNet3Dx3 and InvNet3Dx4 (first row to fourth row).} The fifth row displays vertical velocity profiles (velocity - depth) at the location of the white cursor on 2D slices. Rectangles with white dashed borderline highlight the region of interest in comparison. Each column presents one sample. 2D Ground truth velocity maps are provided in the second row of Figure \ref{fig:vis-baseline}. Zoom-in visualization of the highlighted areas can be found in the supplementary materials.}
    \label{fig:vis-nblocks}
    \end{minipage}
\vspace{-0.75em}
\end{figure*}

We also visualize the predicted velocity map in Figure \ref{fig:vis-nblocks}. It is clear that the highest-quality prediction is obtained with $\mathrm{\#Blocks}=2$ and deterioration starts at $\mathrm{\#Blocks}=3$, which follows the same tendency as numerical results.

\section{Ablation Study}
\label{sec:as}
In section \ref{sec:experiment}, we briefly describe the subsampling strategy on the temporal and channel dimensions of input seismic data. We provide more detailed analysis of the influence of different strategies on training and testing and seek more insight into the proposed model. Furthermore, the generalizability of a network to new environment other than the training settings is one of the critical topics in deep-learning-based FWI. Here we also conduct detailed experiments on the robustness of our network to noise and seismic frequency contents as well as the generalization of it to various source signatures and out-of-distribution data. For simplicity, all the following experiments are conducted with a InvNet3Dx1 model. 

\subsection{Influence of Input Temporal Resolution}
\label{sec:as-temp-downsample}
In prior research on 2D FWI \cite{wu2020a, zhang2019c}, the temporal length ($T$) of input data was set to 1000. In this paper, we adopt a similar value, 896 as the default configuration. Although the raw seismic data in 3D Kimberlina dataset have a temporal length of 5,001 samples, which means uniformly extracting 896 slices would lead to more than $4/5$ information on the temporal dimension being discarded, the extracted data sequences are still visibly redundant. In order to understand how sparse the seismic data could be subsampled temporally without affecting the model's performance, we continue to decrease the temporal subsampling rate and the results are shown in Table \ref{tab:as-temp-perf}.

\begin{figure*}[]
\centering
    \subfloat[]{ 
        \begin{minipage}{0.18\linewidth}
        \centering
        \includegraphics[width=0.98\linewidth]{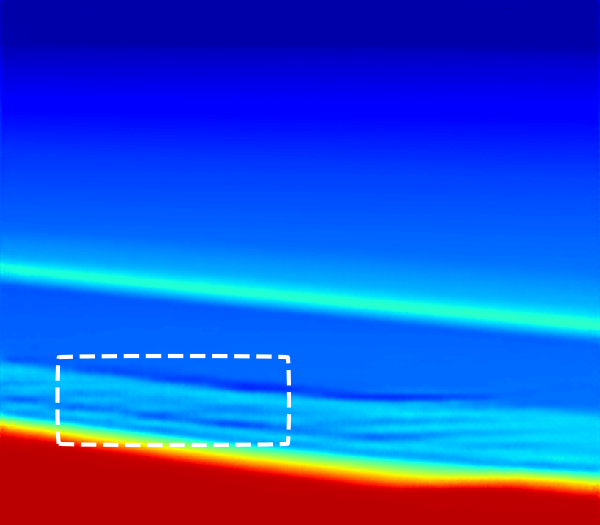} \\ \vspace{5pt}
        \includegraphics[width=0.98\linewidth]{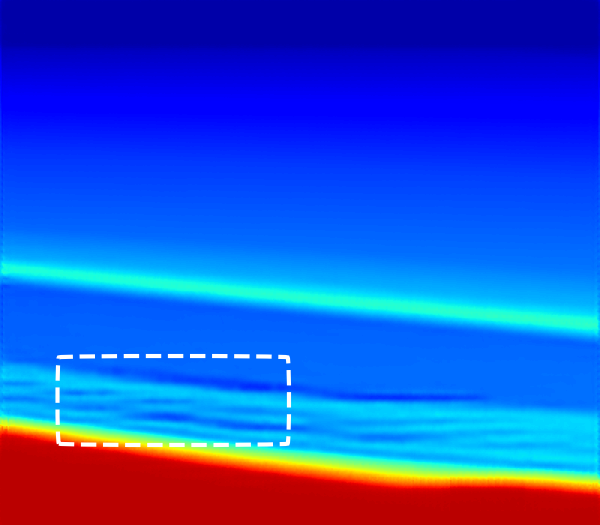} \\ \vspace{5pt}
        \includegraphics[width=0.98\linewidth]{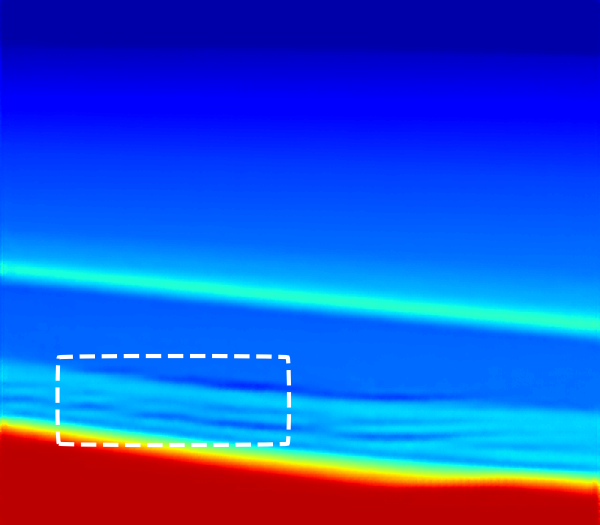} 
        \end{minipage}
    }
    \subfloat[]{
        \begin{minipage}{0.18\linewidth}
        \centering
        \includegraphics[width=0.98\linewidth]{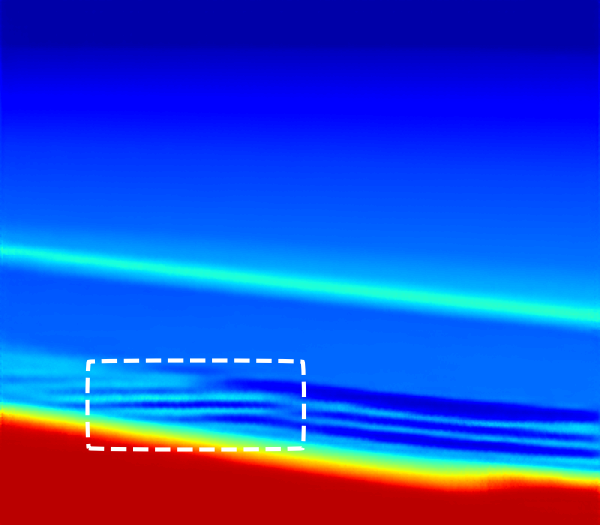} \\ \vspace{5pt}
        \includegraphics[width=0.98\linewidth]{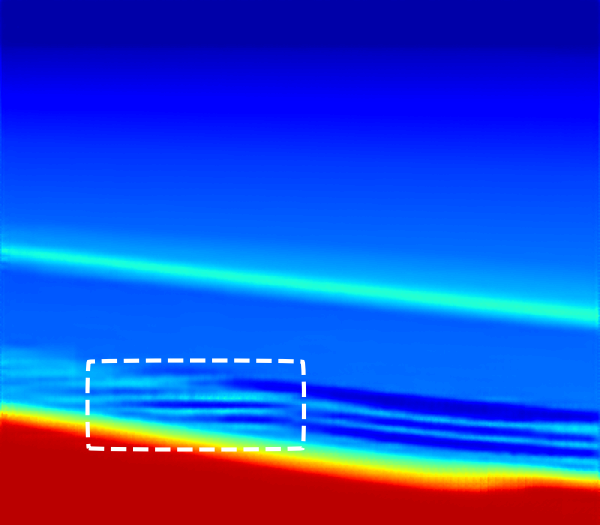} \\ \vspace{5pt}
        \includegraphics[width=0.98\linewidth]{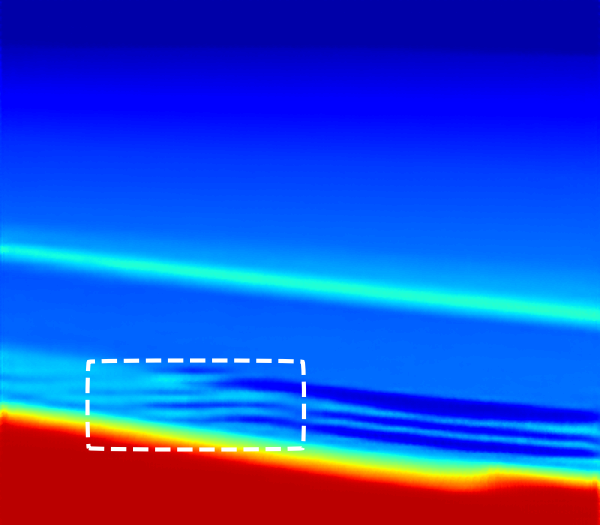} 
        \end{minipage}
    }
    \subfloat[]{ 
        \begin{minipage}{0.18\linewidth}
        \centering
        \includegraphics[width=0.98\linewidth]{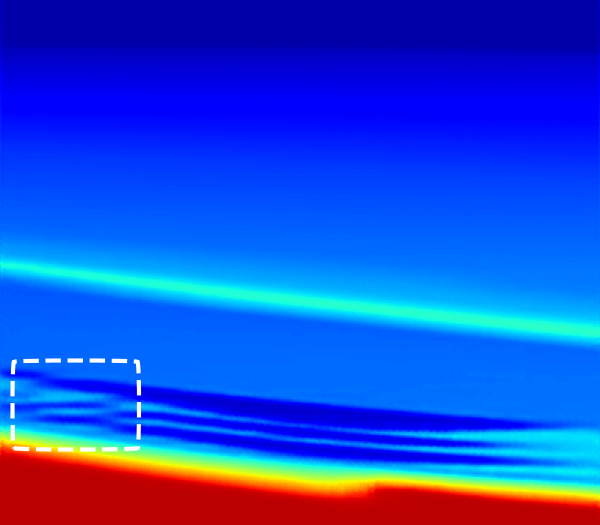} \\ \vspace{5pt}
        \includegraphics[width=0.98\linewidth]{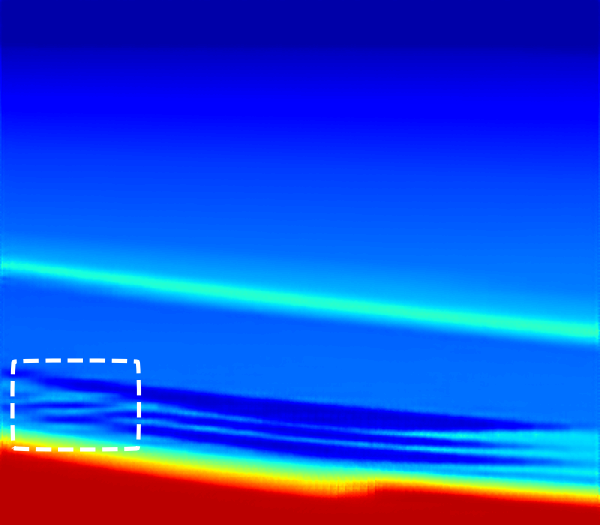} \\ \vspace{5pt}
        \includegraphics[width=0.98\linewidth]{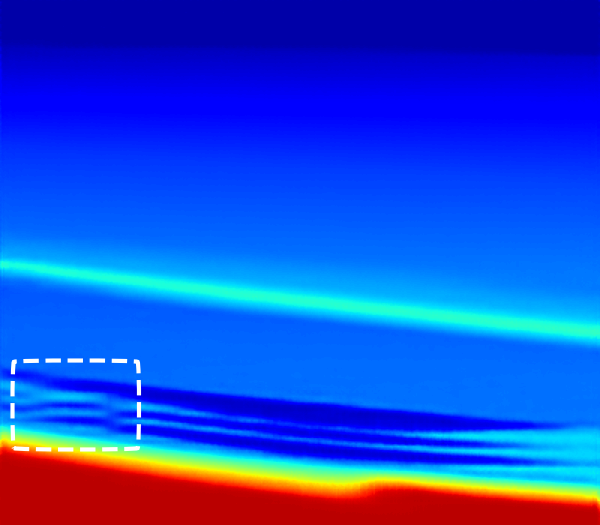} 
        \end{minipage}
    }
    \subfloat[]{ 
        \begin{minipage}{0.18\linewidth}
        \centering
        \includegraphics[width=0.98\linewidth]{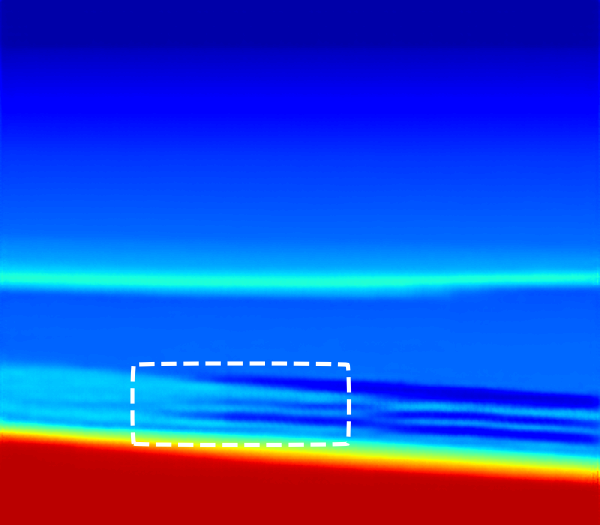} \\ \vspace{5pt}
        \includegraphics[width=0.98\linewidth]{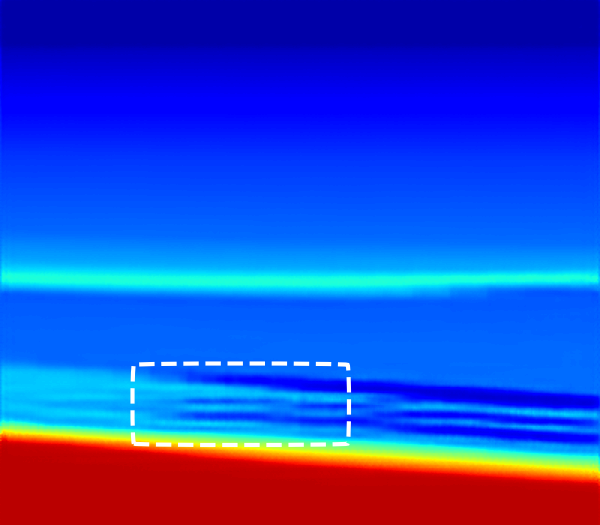} \\ \vspace{5pt}
        \includegraphics[width=0.98\linewidth]{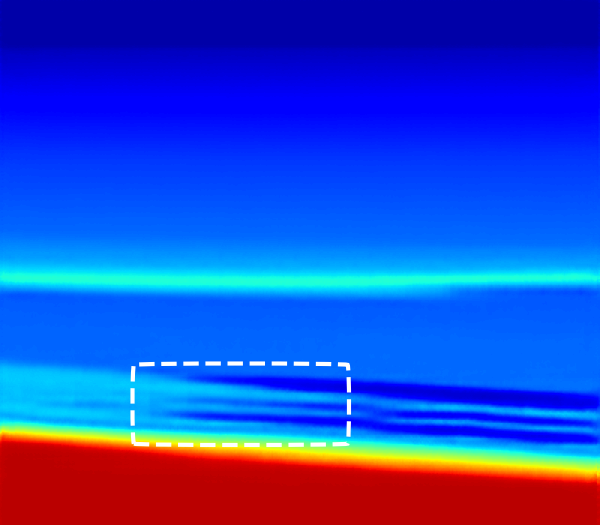} 
        \end{minipage}
    }
    \subfloat[]{
        \begin{minipage}{0.18\linewidth}
        \centering
        \includegraphics[width=0.98\linewidth]{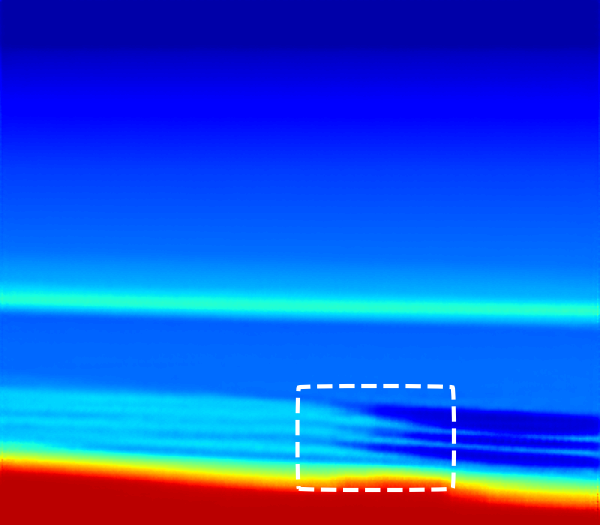} \\ \vspace{5pt}
        \includegraphics[width=0.98\linewidth]{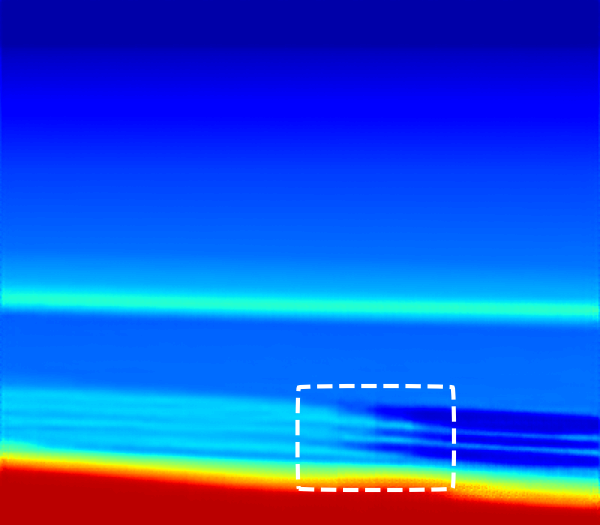} \\ \vspace{5pt}
        \includegraphics[width=0.98\linewidth]{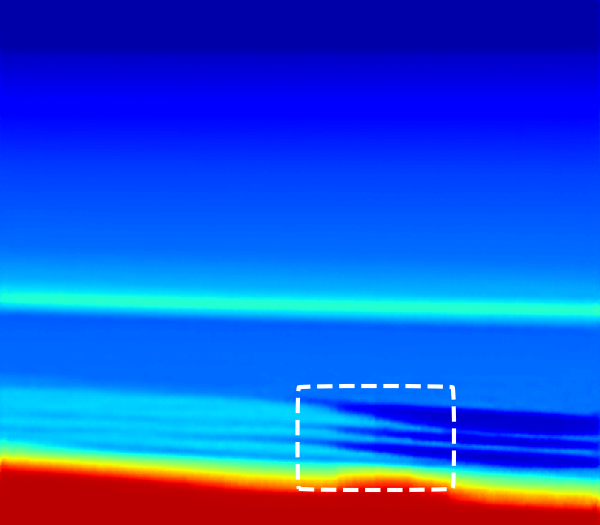} 
        \end{minipage}
    }
    \begin{minipage}{0.06\linewidth}
    \centering
    \includegraphics[width=0.95\linewidth]{Figure/color_bar.png}
    \end{minipage}
    
    \begin{minipage}{0.98\linewidth}
    \centering
    \caption{\textbf{Visualization of predicted velocity map samples generated by InvNet3Dx1 with the input temporal length of 896, 448, and 224 (first row to third row).} Rectangles with white dashed borderline highlight the region of interest in comparison. Each column presents one sample. Ground truth velocity maps are shown in the second row of Figure \ref{fig:vis-baseline}. Zoom-in visualization of the highlighted areas are provided in the supplementary materials.}
    \label{fig:vis-as-temporal}
    \end{minipage}
\vspace{-0.75em}
\end{figure*}

It can be observed that the model with $T=448$ performs very similar to that with $T=896$, which indicates that there is high redundancy on temporal dimension even when only fewer than $1/5$ frames being sampled. However, $T=224$ leads to performance deterioration, though the influence of such highly sparsely sampled input seems to be even slighter than that of the model itself, considering the results shown in Table \ref{tab:main-comparison} and \ref{tab:block-comparison}. Visualization in Figure \ref{fig:vis-as-temporal} reflects the same tendency. Due to the temporal downsampling scheme of InvNet3D, feeding the network with the input of $T=112$ is infeasible, but from these results, we could already conclude that the raw seismic data does have high redundancy on temporal dimension and the performance of our model is stable within a wide range of temporal subsampling rate.

\begin{table}[]
\centering
\caption{\textbf{The influence of input temporal length ($T$) on reconstruction performance}. All the experiments are implemented with an InvNet3Dx1 model.}
\label{tab:as-temp-perf}
\begin{tabular}{c|ccc}
\hline
$T$         & MAE $\downarrow$      & RMSE $\downarrow$      & SSIM $\uparrow$   \\ \hline
896         & 9.83     & 26.11     & 0.9826  \\
448         & 9.86     & 25.99     & 0.9832     \\
224         & 10.12    & 26.50     & 0.9823    \\\hline
\end{tabular}
\vspace{-0.75em}
\end{table}

\subsection{Selection of Sources}
\label{sec:as-source}

\begin{figure*}
\centering
\begin{minipage}{0.25\linewidth}
    \centering
    \vspace{25pt}
    \includegraphics[height=3.5cm]{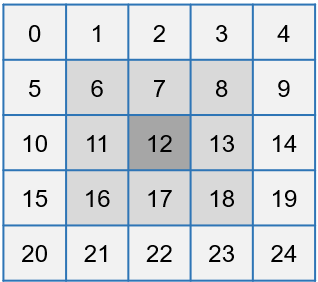}
    \captionof{figure}{\textbf{Spatial Placement of Sources.} Each source independently stimulates seismic wave propagating in subsurface layers. The direct wave and the reflected wave are then captured by receivers on the ground and recorded as seismic data of one channel in the input.}
    \label{fig:source-distribution}
\end{minipage} 
\begin{minipage}{0.72\linewidth}
    \centering
    \begin{minipage}[]{1.0\linewidth}
    \centering
      \includegraphics[width=0.22\linewidth]{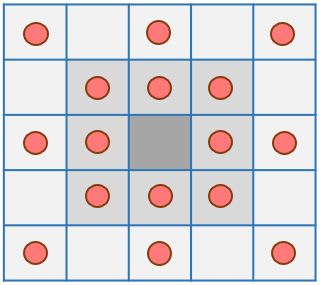}
      \includegraphics[width=0.22\linewidth]{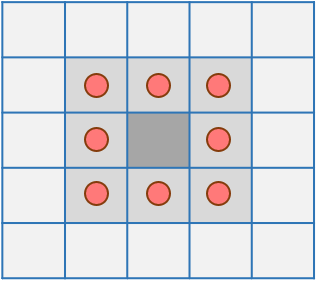}
      \includegraphics[width=0.22\linewidth]{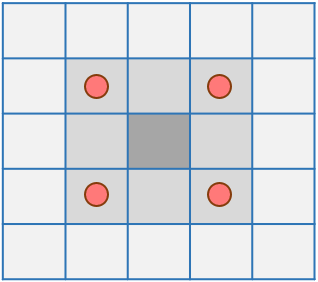} 
      \includegraphics[width=0.22\linewidth]{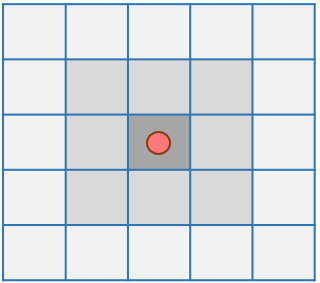}
      \footnotesize
      \begin{tabular}{cccc}
             10.87 / 28.99 / 0.9788%
            &10.38 / 26.95 / 0.9819%
            &11.11 / 29.56 / 0.9784%
            &11.24 / 29.91 / 0.9779\\
      \end{tabular}
      \normalsize
      \vspace{5pt}
   \end{minipage}
   \begin{minipage}[]{1.0\linewidth}
    \centering
      \includegraphics[width=0.22\linewidth]{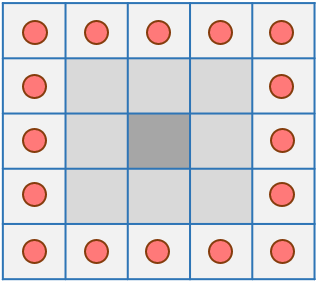}
      \includegraphics[width=0.22\linewidth]{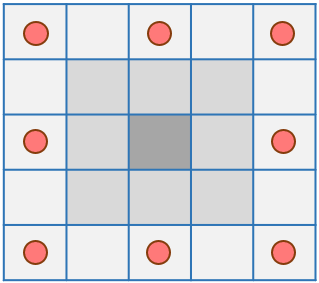}
      \includegraphics[width=0.22\linewidth]{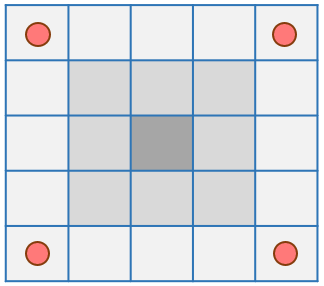} 
      \includegraphics[width=0.22\linewidth]{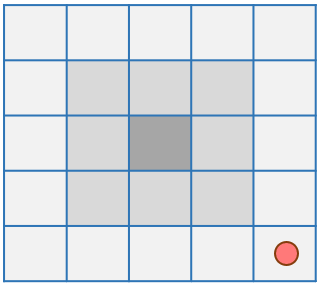}
      \footnotesize
      \begin{tabular}{cccc}
             10.26 / 27.30 / 0.9812%
            &10.19 / 27.27 / 0.9816%
            &10.11 / 26.65 / 0.9822%
            &10.29 / 27.90 / 0.9803\\
      \end{tabular}
      \normalsize
   \end{minipage}
   \begin{minipage}[]{0.9\linewidth}
   \centering
   \captionof{figure}{\textbf{Influence of different source selection strategy on reconstruction performance.} The grid with a red dot indicates the seismic data generated by the source placed at the topological location is selected as one of the input channels for training and testing InvNet3D. The numbers below each figure provides the performance resulted from the corresponding strategy, listed in the format of MAE / RMSE / SSIM.  Further analysis on the source illumination of the top-left and the bottom-right strategies are included in the supplementary materials.}
   \label{fig:source-selection}
   \end{minipage}

\end{minipage}%
\vspace{-0.75em}
\end{figure*}

There are 25 seismic records per sample in the 3D Kimberlina dataset. Figure \ref{fig:source-distribution} displays the placement of corresponding seismic sources on a 2D grid. Our default training and testing configurations make use of 8 out of 25 records, namely records 1, 2, 14, 15, 16, 20, 23 and 24, which is a randomly selected trivial placement scheme. The number of channels is determined based on the consideration of the capacity and characteristics of Channel-Separated Encoder, for the reasons outlined below.

We first study the impact of the number of selected channels ($\mathrm{\#Channels}$) by training and testing InvNet3Dx1 with 5 randomly generated strategies of channel selection for every $\mathrm{\#Channels}$ in ${1,4,8,16}$. Note that when $\mathrm{\#Channels}=1$, InvNet3Dx1 is equivalent to InvNet3DGx1. The averaged results for each $\mathrm{\#Channels}$ value are reported in Table \ref{tab:as-nsource-perf}. It is clear that the model performs the best with $\mathrm{\#Channels}=8$. It can be easily understood that a smaller number of records contain fewer information of subsurface structure, thereby resulting in inferior performance. However, the model with $\mathrm{\#Channels}=16$ also performs worse. This result can be explained based on the workflow of Channel-Separated Encoder. Since we always set the group size equal to $\mathrm{\#Channels}$, the number of filters corresponding to each input channel would be larger when using a smaller $\mathrm{\#Channels}$, which means more features per record will be extracted at the very first layer and fewer low-level information would be discarded.

\begin{table}[]
\centering
\caption{\textbf{The influence of number of selected sources on reconstruction performance}. The reported results are averaged from five experiments with different channel selection strategy of each \#channels configuration. All the experiments are implemented with an InvNet3Dx1 model. }
\label{tab:as-nsource-perf}
\begin{tabular}{c|ccc}
\hline
\#Channels & MAE $\downarrow$      & RMSE $\downarrow$      & SSIM $\uparrow$ \\ \hline
16         & 10.73    & 28.49     & 0.9798     \\
8          & 10.41    & 27.57     & 0.9811     \\
4          & 10.93    & 29.12     & 0.9795     \\
1          & 11.25    & 29.94     & 0.9786    \\\hline
\end{tabular}
\vspace{-0.75em}
\end{table}

The choice of $\mathrm{\#Channels}$ also influence the stage in the network where filters start to possess a global receptive field over all the raw input channels, which will further impact the learning of high-level representations from the input. When $\mathrm{\#Channels}$ becomes excessively large, filters receiving information from all the raw input channels exist in very deeper layers and thus there may not be sufficient subsequent layers for processing global information on channel dimension. Increasing the number of filters may help to improve the performance when $\mathrm{\#Channels}=16$ or even larger, but it will also enlarge the model capacity as well. Therefore, for a given model capacity, more seismic data records being sampled for input do not necessarily contribute to higher performance. 

We further analyze the influence of the detailed strategy of channel selection. For $\mathrm{\#Channels}$ in $\{1,4,8,16\}$, we select two typical sampling plans each, which are intuitively described in Figure \ref{fig:source-selection}. Strategies in the same column sample the same number of sources, while those in the first-row sample sources with an inner-ward arrangement compared to those in the second row. Performance evaluation results are listed below the corresponding figure. With regard to the influence of $\mathrm{\#Channels}$, these detailed experiments lead to conclusions in accordance with Table \ref{tab:as-nsource-perf}. However, there are some surprising results in terms of the placement of chosen sources. It can be observed that the outer-ward selection strategies of sources lead to higher performance, especially when $\mathrm{\#Channels}\leq4$. 
We believe this phenomenon implies that the reconstruction performance at the boundary area is sensitive to the coverage of seismic sources. When $\mathrm{\#Channels}$ is large, critical information for reconstructing this marginal can be well presented. However, when $\mathrm{\#Channel}$ becomes smaller, outward strategies have to be adopted in order to alleviate the deterioration of prediction for this area. This can be also inferred from the higher stability of performance with outer-ward strategies.  

\subsection{Robustness to Noise}
\label{sec:noise}

\begin{figure}
    \centering
    \includegraphics[width=0.75\linewidth]{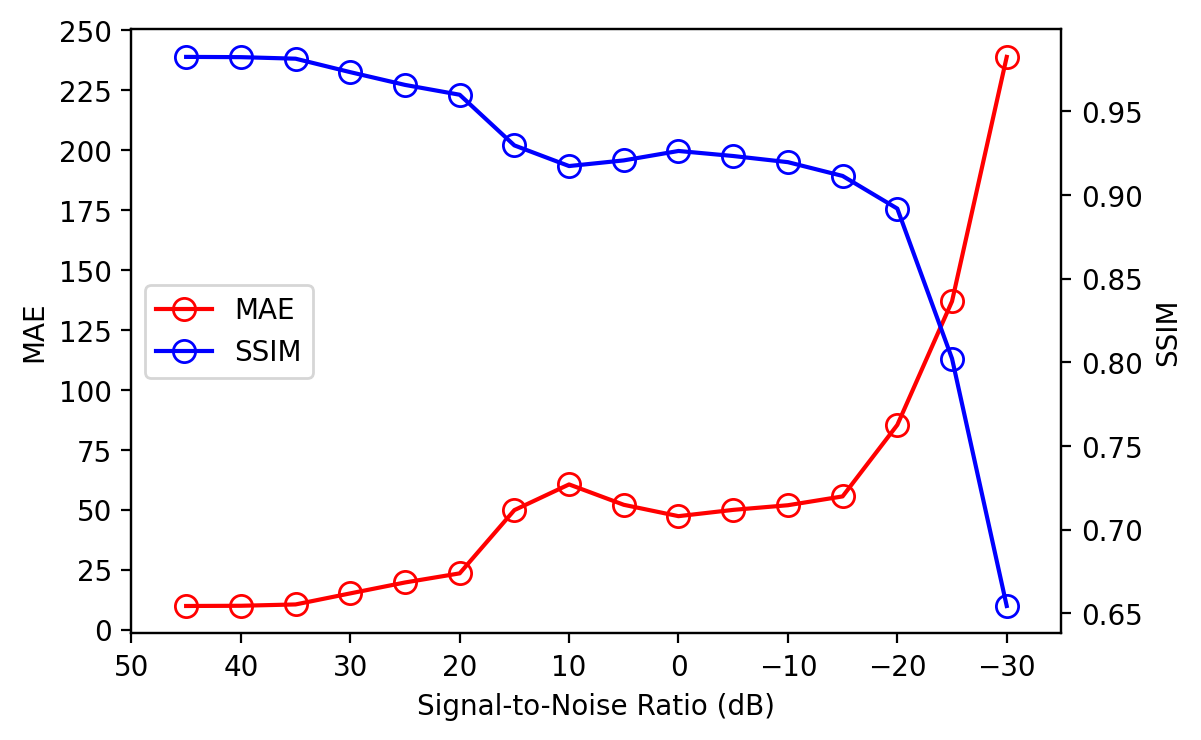}
    \caption{\textbf{Robustness of InvNet3Dx1 to gaussian noise.} The InvNet3Dx1 model used in this experiment is pre-trained on noise-free data. Note that RMSE, which follows the same tendency as MAE, is not included in this figure due to its larger value scale. Smaller SNR indicates stronger noise.}
    \label{fig:as-noise}
\vspace{-0.75em}
\end{figure}

Seismic data collected in the field may contain the noise of certain levels. To demonstrate the robustness of the proposed InvNet3D against various noise levels, we manually impose Gaussian noise to the validation dataset and test the InvNet3Dx1 model under different signal-to-noise ratios~(SNR). Note that the model is trained in a noise-free environment and its performance on the original noise-free validation dataset is reported in Table~\ref{tab:main-comparison}. 

\begin{figure*}[]
\centering
    \begin{minipage}{0.96\linewidth}
        \subfloat[$\mathrm{SNR}=30\mathrm{dB}$]{
            \centering
            \includegraphics[width=0.18\linewidth]{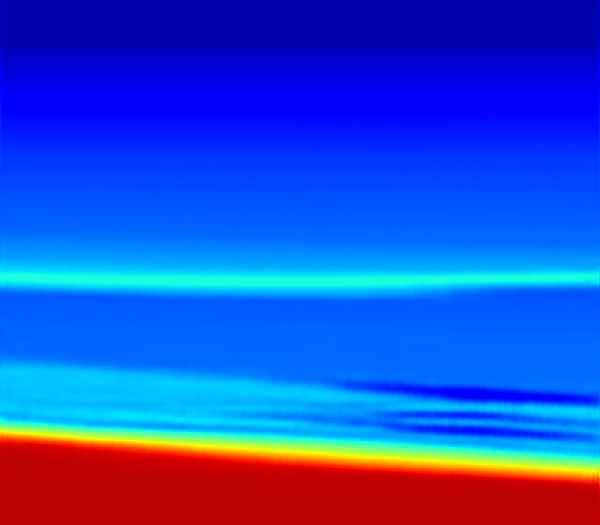}
        }
        \subfloat[$\mathrm{SNR}=20\mathrm{dB}$]{
            \centering
            \includegraphics[width=0.18\linewidth]{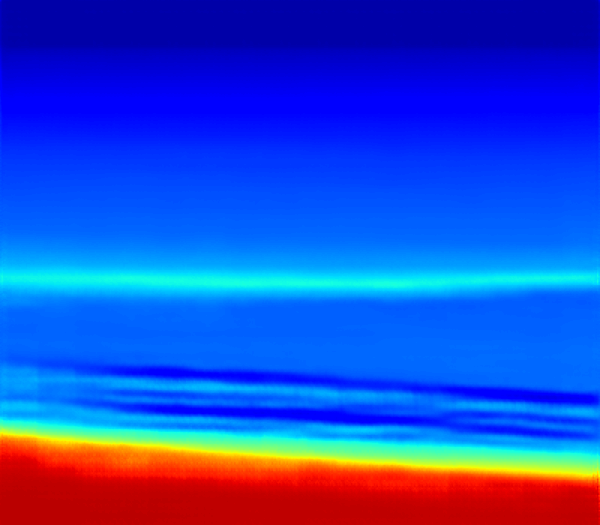}
        }
        \subfloat[$\mathrm{SNR}=10\mathrm{dB}$]{
            \centering
            \includegraphics[width=0.18\linewidth]{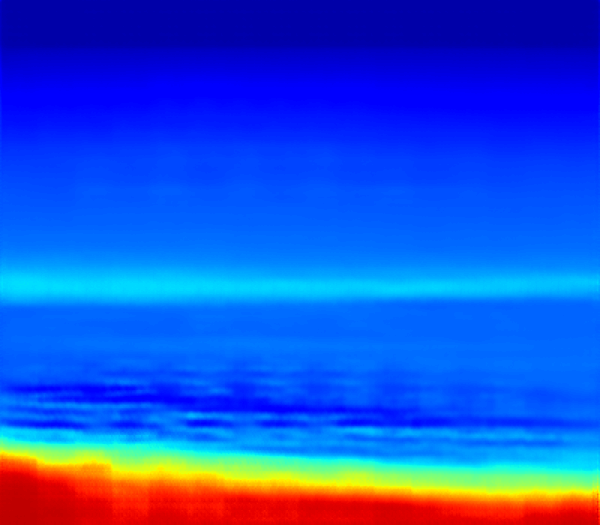}
        }
        \subfloat[$\mathrm{SNR}=0\mathrm{dB}$]{
            \centering
            \includegraphics[width=0.18\linewidth]{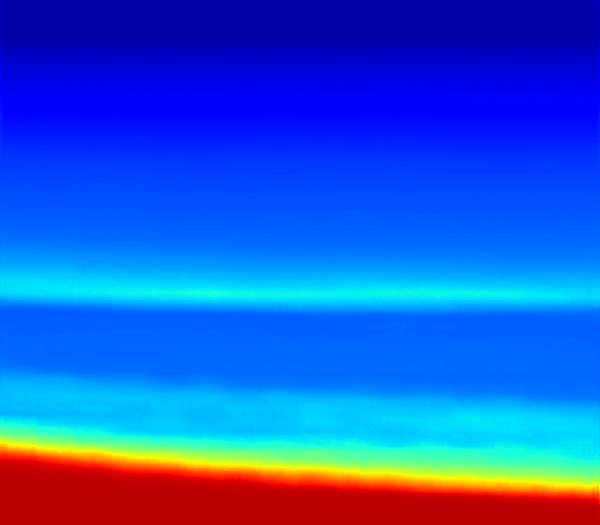} 
        }
        \subfloat[$\mathrm{SNR}=-10\mathrm{dB}$]{ 
            \includegraphics[width=0.18\linewidth]{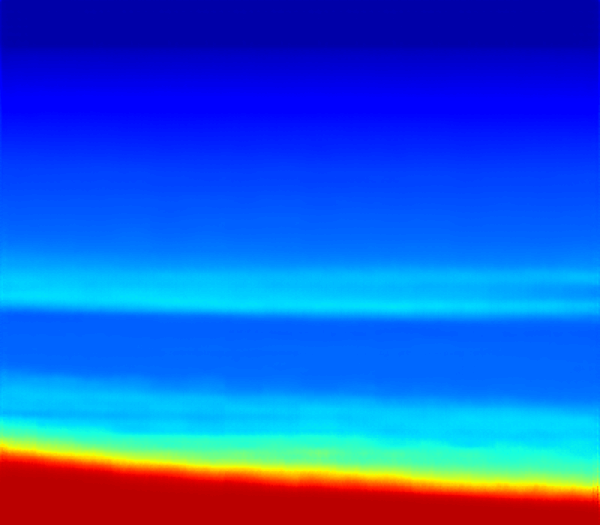} 
        }
    \end{minipage}
    \begin{minipage}{0.95\linewidth}
    \centering
    \caption{\textbf{Visualization of predicted velocity map samples (2D slice) generated by InvNet3Dx1 under various noise levels.} The ground truth velocity map is visualized in the second row of Figure~\ref{fig:vis-baseline-4} and the color map of these two Figures are also the same.}
    \label{fig:vis-as-noise}
    \end{minipage}
\vspace{-0.75em}
\end{figure*}

Figure~\ref{fig:as-noise} reports the results of this experiment, from which it can be observed that InvNet3Dx1 maintains competitive performance until $\mathrm{SNR}=20\mathrm{dB}$. The performance, though deteriorates, is still stable within a wide $\mathrm{SNR}$ range of $[-15, 20]\mathrm{dB}$. However, a real-world environment with a constant negative SNR is rare and thus it is shown that InvNet3Dx1 is robust to noise in typical application scenarios. Some sampled visualization can be found in Figure~\ref{fig:vis-as-noise}.

\subsection{Contribution of Low-frequency Components in Prediction}

\begin{figure}
    \centering
    \includegraphics[width=0.75\linewidth]{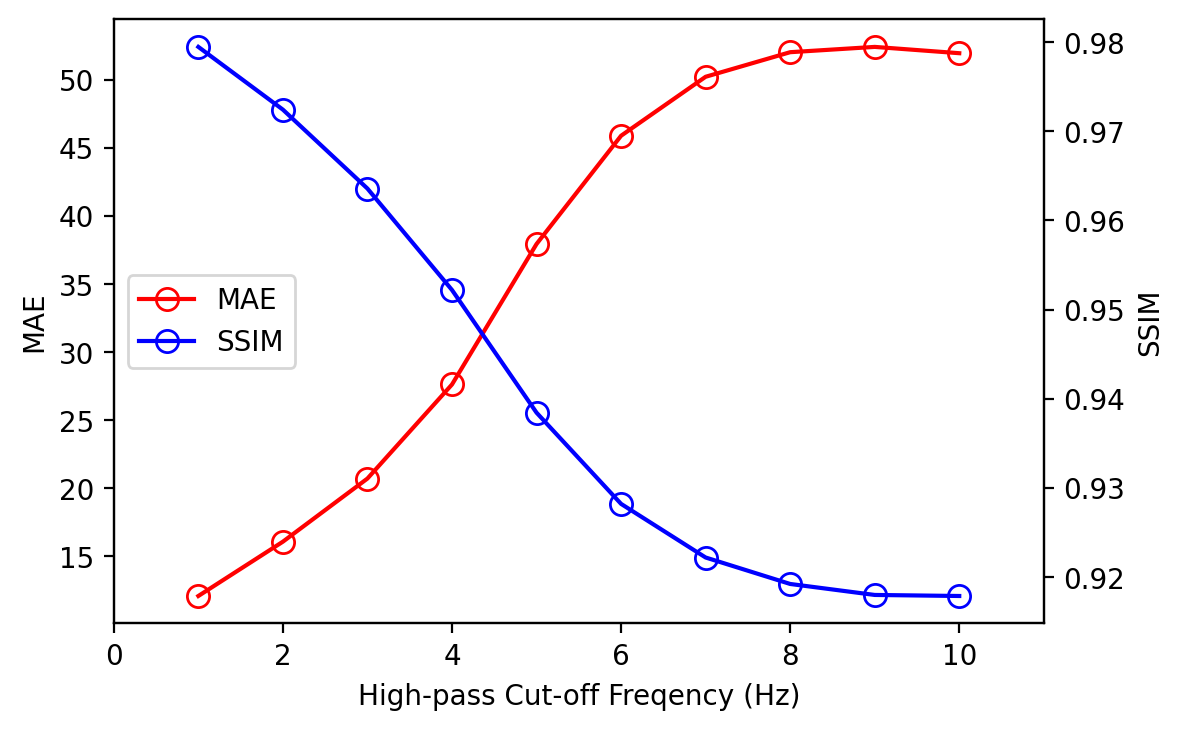}
    \caption{\textbf{Influence of the suppression on low-frequency components.} A second-order high-pass Butterworth filter is used for removing low frequencies. The InvNet3Dx1 model used in this test is pre-trained on full-frequency data. Note that RMSE, which follows the same tendency as MAE, is not included in this figure due to its larger value scale.}
    \label{fig:as-freq}
\vspace{-0.75em}
\end{figure}

Low-frequency components in seismic data, though can be helpful in the reconstruction of background velocity, is hard to obtain in real-world measurement. To understand the contribution of these components in FWI and analyze the dependency of our networks on them, we use high-pass filters with cut-off frequencies ranging from 1Hz to 10Hz to remove low frequencies from the data and directly test the pre-trained network on the resulted datasets. Note that higher cut-off frequency indicates more low-frequency components being removed and the central frequency of the original data is around 15Hz.

The results are provided in Figure \ref{fig:as-freq}, which shows the suppression of low frequencies significantly influences the performance. However, the model produces fairly high-quality images when cut-off frequency $\leq$ 4Hz, where SSIM is above 0.95. Moreover, we believe these results are reasonable since a network pre-trained on full-frequency data is not trained to handle the highly deteriorated environment.

\subsection{Adaptation to Different Source Signatures}

\begin{figure*}[]
\centering
    \begin{minipage}{0.72\linewidth}
    \centering
        \subfloat[5Hz Ricker]{
            \centering
            \includegraphics[width=0.24\linewidth]{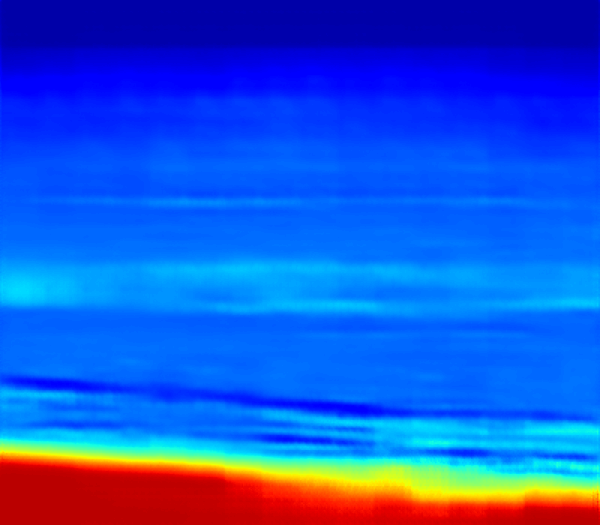}
        }
        \subfloat[10Hz Ricker]{
            \centering
            \includegraphics[width=0.24\linewidth]{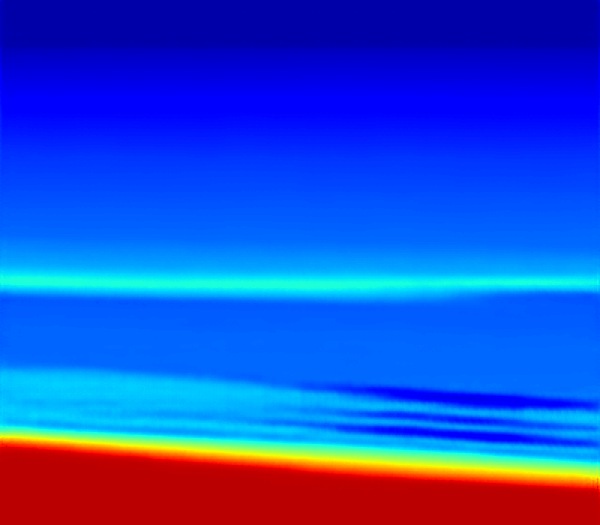} 
        }
        \subfloat[\textbf{15Hz Ricker}]{ 
           \centering
           \includegraphics[width=0.24\linewidth]{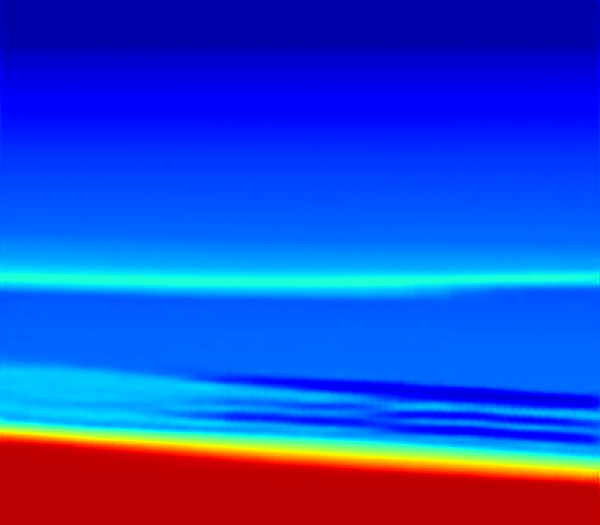} 
        }
        \subfloat[20Hz Ricker]{ 
            \centering
            \includegraphics[width=0.24\linewidth]{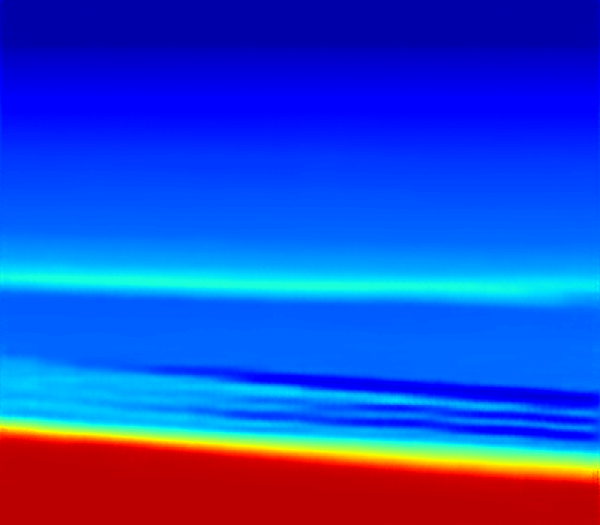}
        }\\[-2ex]%
        \subfloat[25Hz Ricker]{ 
            \centering
            \includegraphics[width=0.24\linewidth]{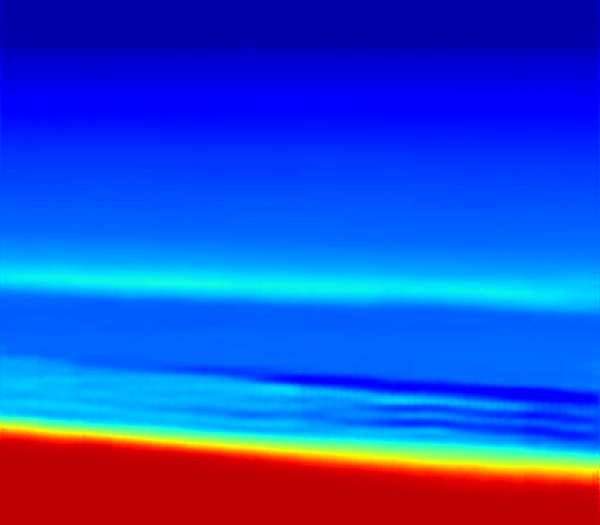} 
        }
         \subfloat[Gaussian-4]{ 
            \centering
            \includegraphics[width=0.24\linewidth]{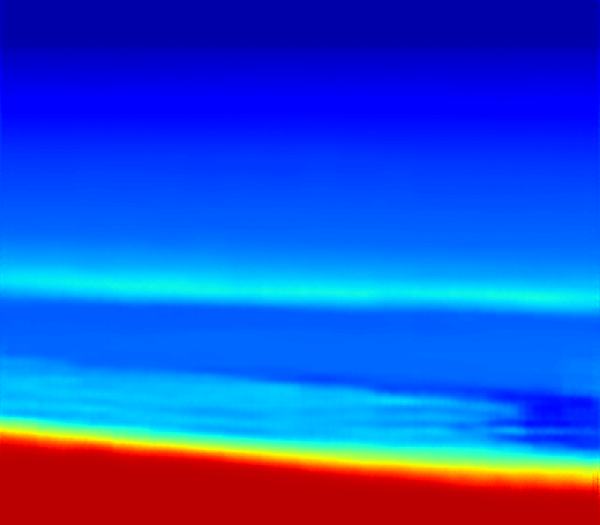} 
        }
        \subfloat[Gaussian-8]{
            \centering
            \includegraphics[width=0.24\linewidth]{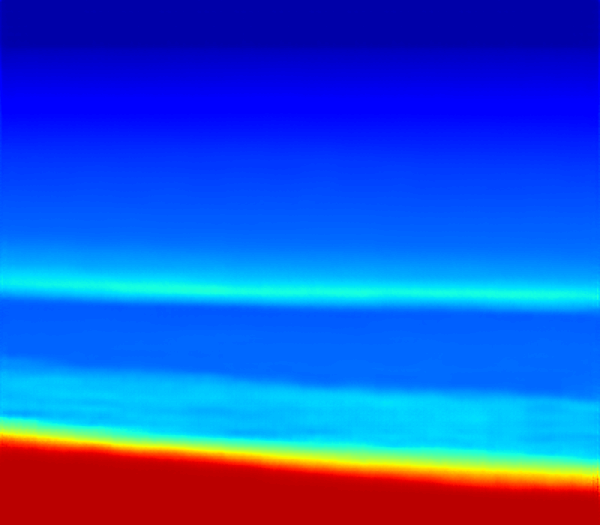}
        }
        \subfloat[Meyer]{
            \centering
            \includegraphics[width=0.24\linewidth]{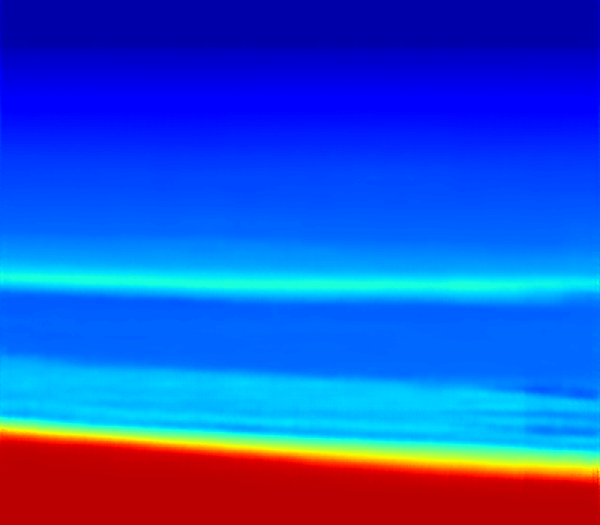} 
        }
    \end{minipage}
    \hspace{2em}
    \begin{minipage}{0.06\linewidth}
    \centering
    \includegraphics[width=0.95\linewidth]{Figure/color_bar.png}
    \end{minipage}
    
    \begin{minipage}{0.95\linewidth}
    \centering
    \caption{\textbf{Visualization of predicted velocity map samples (2D slice) generated by InvNet3Dx1 on seismic data with different source signatures.} The ground truth velocity map is visualized in the second row of Figure~\ref{fig:vis-baseline-4}.}
    \label{fig:vis-as-signature}
    \end{minipage}
\vspace{-1.0em}
\end{figure*}

\begin{table*}[]
\centering
\begin{minipage}{0.98\linewidth}
\caption{\textbf{Testing performance of InvNet3Dx1 on sampled seismic data with different source sigantures}. The model is trained on data generated with a 15Hz Ricker wavelet. Gaussian-4 and Gaussian-8 refer to order 4 and order 8 Gaussian wavelets.}
\label{tab:as-source-signature}
\end{minipage}
\begin{tabular}{l|c|c|c|c|c|c|c|c}
\hline
     & 5Hz Ricker    & 10Hz Ricker   & \textbf{15Hz Ricker}   & 20Hz Ricker   & 25Hz Ricker   & Gaussian-4 & Gaussian-8 & Meyer  \\ \hline
MAE $\downarrow$  & 112.04 & 30.83  & \textbf{9.74}   & 28.45  & 34.74  & 41.65     & 45.65     & 29.89  \\
RMSE $\downarrow$ & 228.76 & 75.04  & \textbf{26.35}  & 70.84  & 81.67  & 98.88     & 111.12    & 83.18  \\
SSIM $\uparrow$ & 0.8680 & 0.9444 & \textbf{0.9819} & 0.9437 & 0.9347 & 0.9274    & 0.9239    & 0.9464 \\ \hline
\end{tabular}
\vspace{-0.75em}
\end{table*}

Source uncertainty has been a long-term challenge for FWI, where the accurate estimation of source signature is of great importance. Although explicit source estimation is not required in deconvolution- and convolution-based source-independent approaches \cite{xu2006,song2020}, a proper reference trace has to be determined to formulate the objective function. In the main experiments, our models are trained on seismic data generated using a Ricker wavelet with acentral frequency of 15Hz. To test whether the model can handle different source signatures in the testing environment, we directly evaluate the pre-trained InvNet3Dx1 on seismic data generated with various source central frequencies and wavelets. The ground truth velocity model in this experiment is randomly selected from the validation subset used in the main experiment. Table~\ref{tab:as-source-signature} shows the numerical results and Figure~\ref{fig:vis-as-signature} visualizes the predictions of a sampled velocity map under different source signatures. The sources wavelets involved in this experiments are intuitively displayed in the supplementary materials.

It is clear that a direct migration of the model into a new environment with a different source signature can be over-optimistic. As expected, the more the source central frequency in testing is shifted from that in training, the larger performance drop is present. That being said, the results from such direct adaption should still be better than traditional methods when source estimation is not accurate enough and the source invariance of a network can be further improved if it is trained on a larger dataset comprised of seismic data generated with different source signatures.

\subsection{Out-of-Distribution Reconstruction}

\begin{figure*}[]
\centering
    \subfloat[]{ 
        \begin{minipage}{0.18\linewidth}
        \centering
        \includegraphics[width=0.98\linewidth]{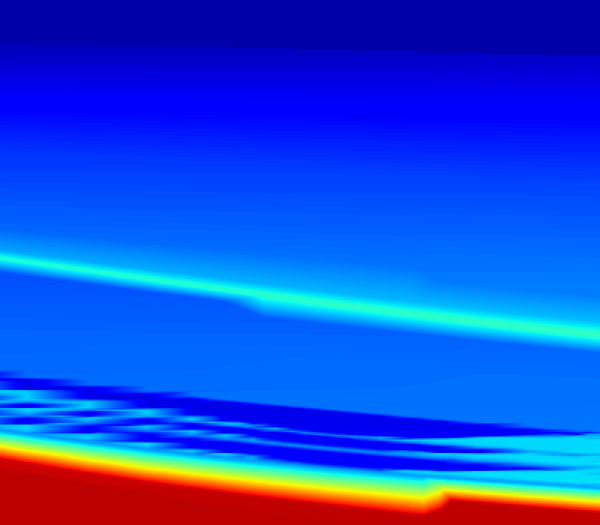} \\ \vspace{5pt}
        \includegraphics[width=0.98\linewidth]{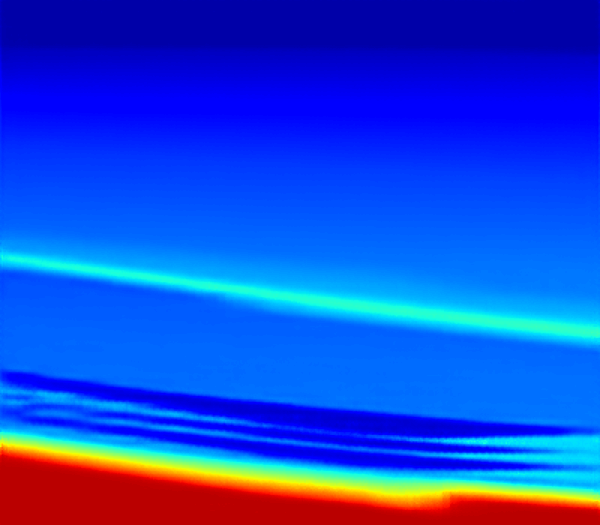} 
        \end{minipage}
    }
    \subfloat[]{
        \begin{minipage}{0.18\linewidth}
        \centering
        \includegraphics[width=0.98\linewidth]{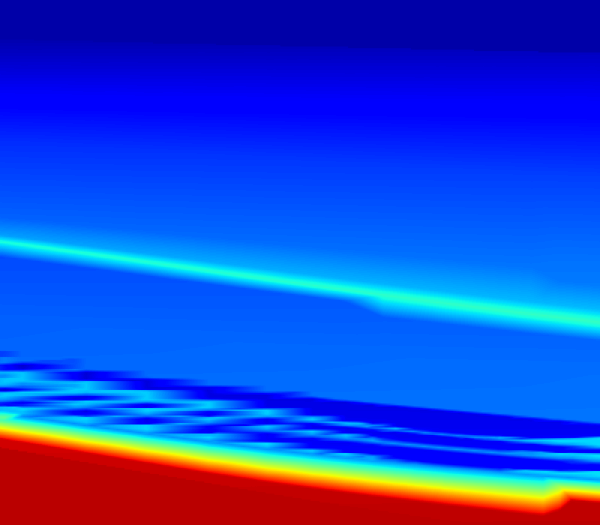} \\ \vspace{5pt}
        \includegraphics[width=0.98\linewidth]{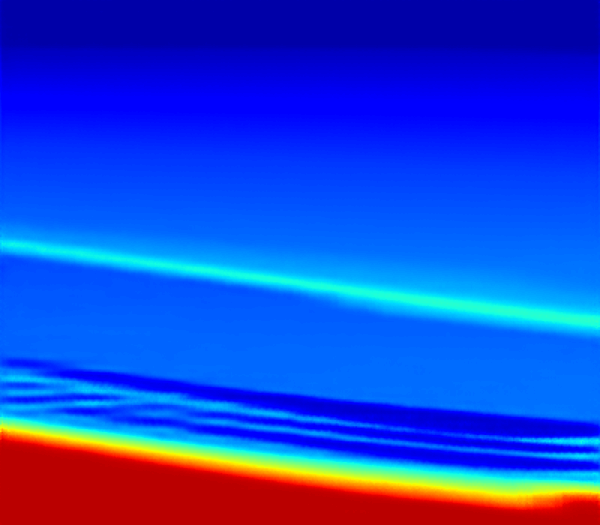} 
        \end{minipage}
    }
    \subfloat[]{ 
        \begin{minipage}{0.18\linewidth}
        \centering
        \includegraphics[width=0.98\linewidth]{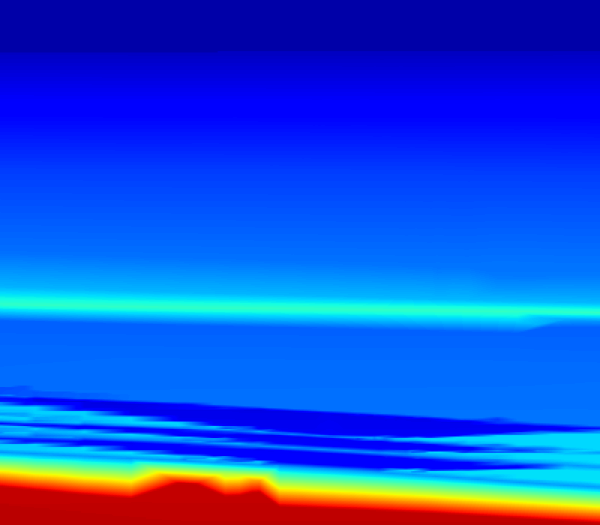} \\ \vspace{5pt}
        \includegraphics[width=0.98\linewidth]{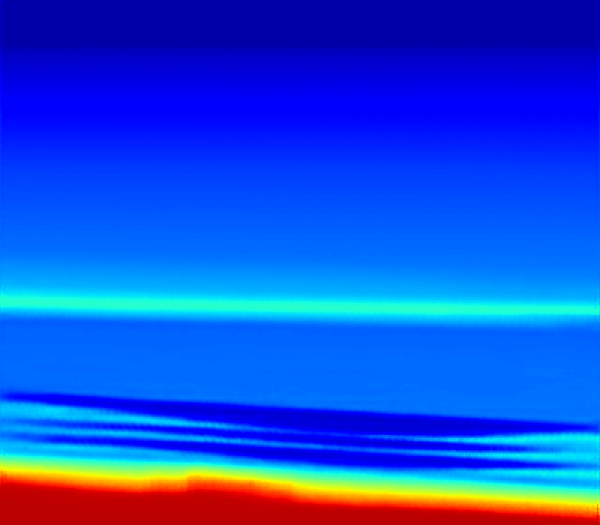} 
        \end{minipage}
    }
    \subfloat[]{ 
        \begin{minipage}{0.18\linewidth}
        \centering
        \includegraphics[width=0.98\linewidth]{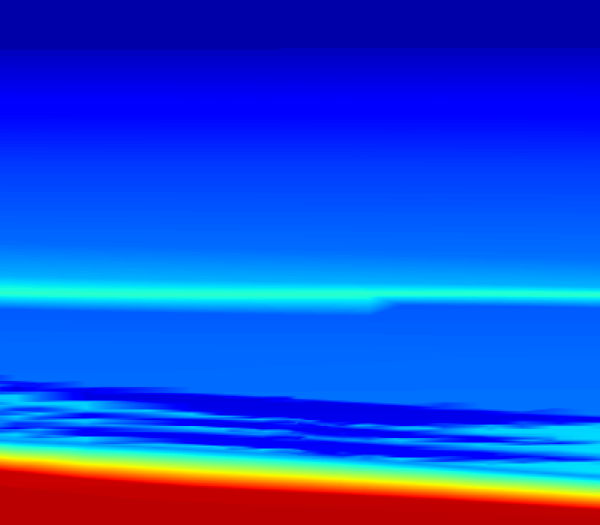} \\ \vspace{5pt}
        \includegraphics[width=0.98\linewidth]{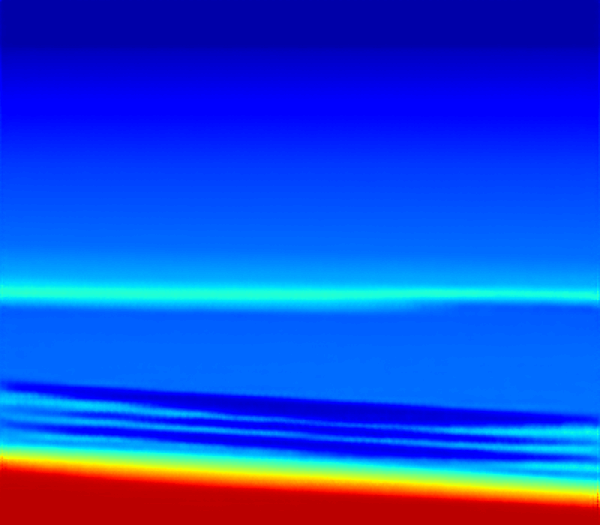} 
        \end{minipage}
    }
    \subfloat[]{
        \begin{minipage}{0.18\linewidth}
        \centering
        \includegraphics[width=0.98\linewidth]{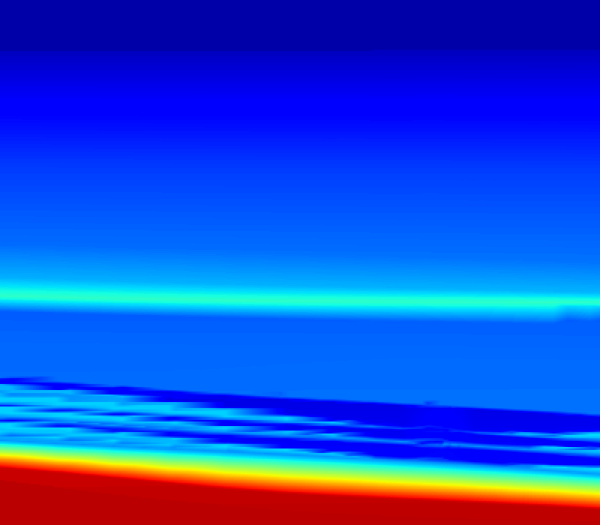} \\ \vspace{5pt}
        \includegraphics[width=0.98\linewidth]{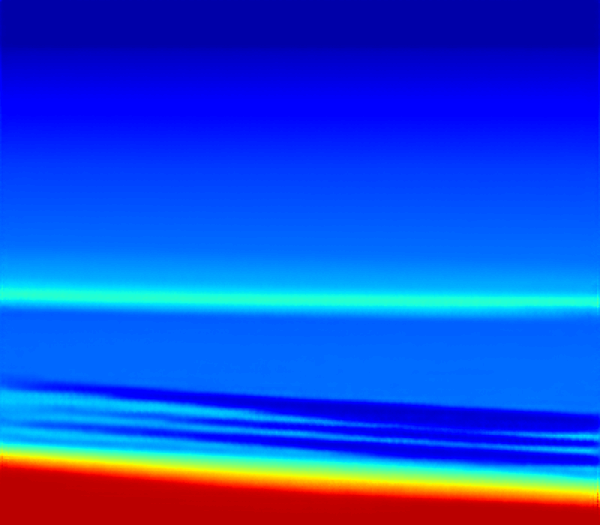} 
        \end{minipage}
    }
    \begin{minipage}{0.06\linewidth}
    \centering
    \includegraphics[width=0.95\linewidth]{Figure/color_bar.png}
    \end{minipage}
    
    \begin{minipage}{0.95\linewidth}
    \centering
    \caption{\textbf{Visualization of predicted velocity map samples (2D slice) generated by InvNet3Dx1 on out-of-distribution testing dataset.} Ground truth and the corresponding prediction are placed in the first and second row, respectively. Each column presents one sample. }
    \label{fig:vis-as-ood}
    \end{minipage}
\vspace{-0.75em}
\end{figure*}

\begin{figure*}[]
\centering
    \subfloat[]{ 
        \begin{minipage}{0.18\linewidth}
        \centering
        \includegraphics[width=0.98\linewidth]{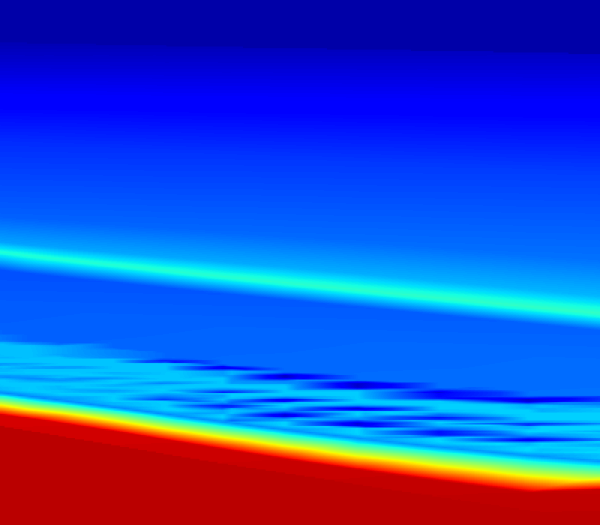} \\ \vspace{5pt}
        \includegraphics[width=0.98\linewidth]{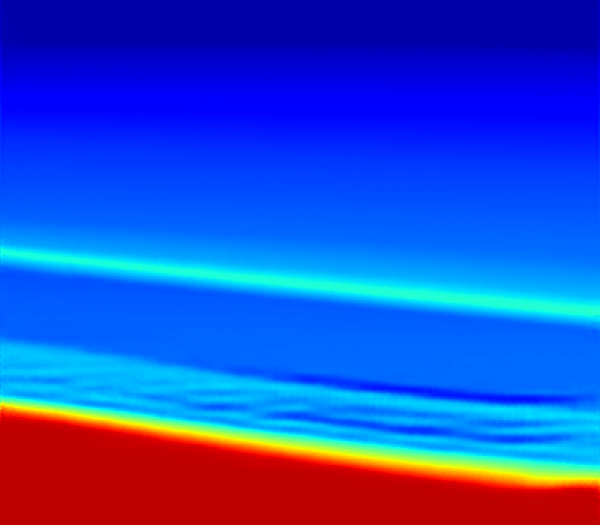} \\ \vspace{5pt}
        \includegraphics[width=0.98\linewidth]{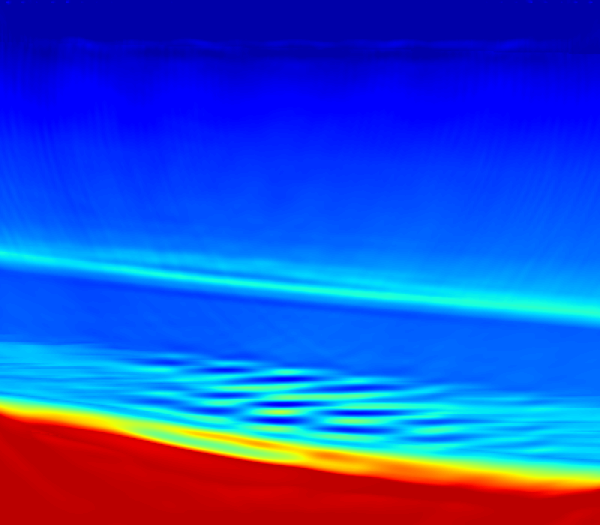} 
        \end{minipage}
    }
    \subfloat[]{
        \begin{minipage}{0.18\linewidth}
        \centering
        \includegraphics[width=0.98\linewidth]{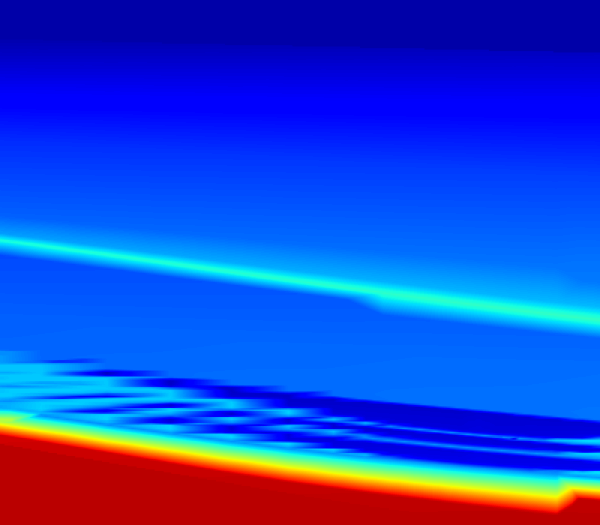} \\ \vspace{5pt}
        \includegraphics[width=0.98\linewidth]{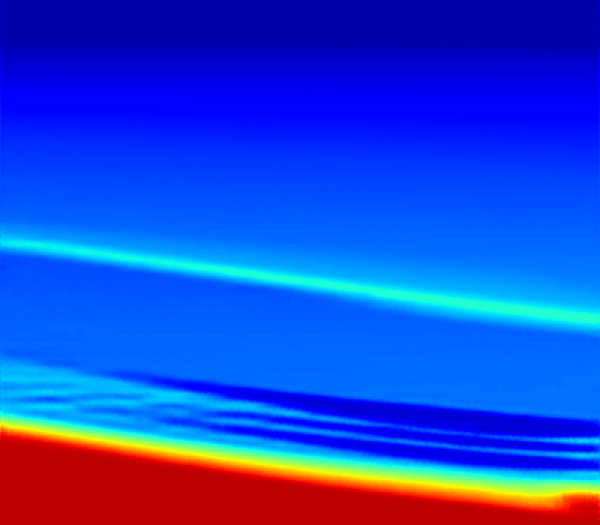} \\ \vspace{5pt}
        \includegraphics[width=0.98\linewidth]{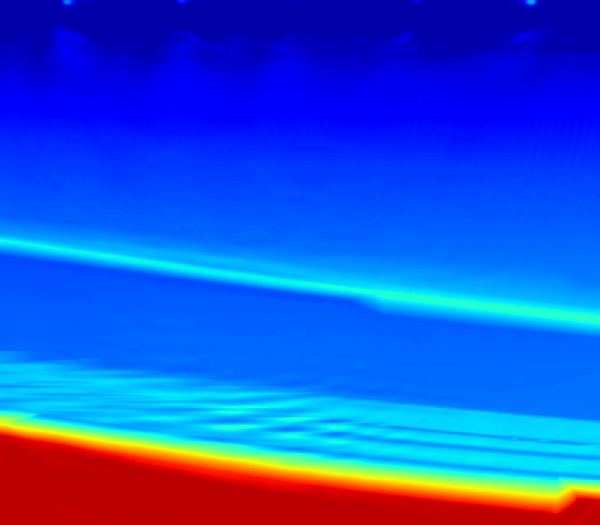} 
        \end{minipage}
    }
    \subfloat[]{ 
        \begin{minipage}{0.18\linewidth}
        \centering
        \includegraphics[width=0.98\linewidth]{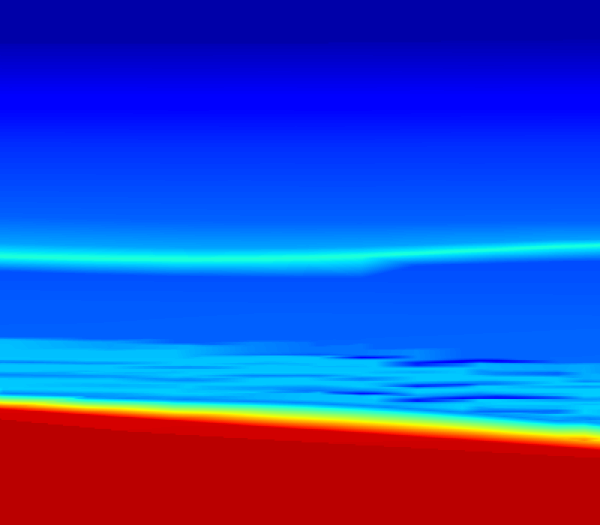} \\ \vspace{5pt}
        \includegraphics[width=0.98\linewidth]{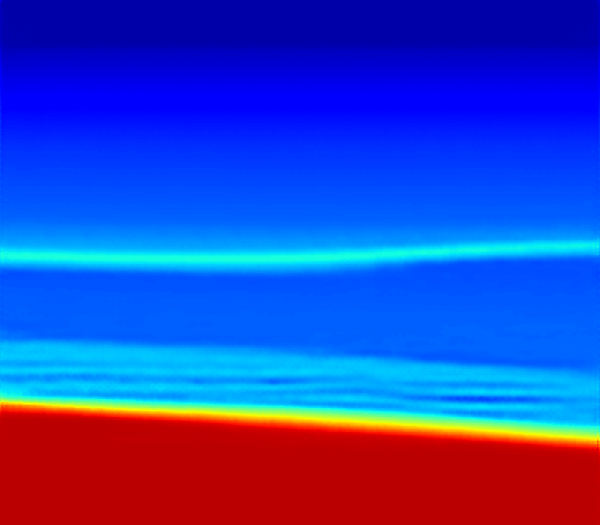} \\ \vspace{5pt}
        \includegraphics[width=0.98\linewidth]{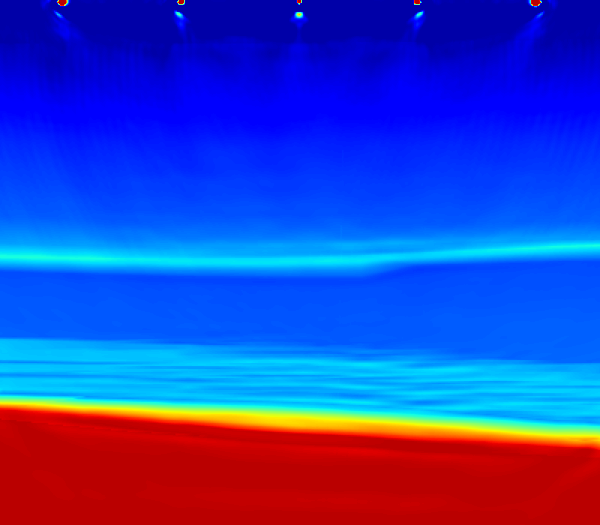} 
        \end{minipage}
    }
    \subfloat[]{ 
        \begin{minipage}{0.18\linewidth}
        \centering
        \includegraphics[width=0.98\linewidth]{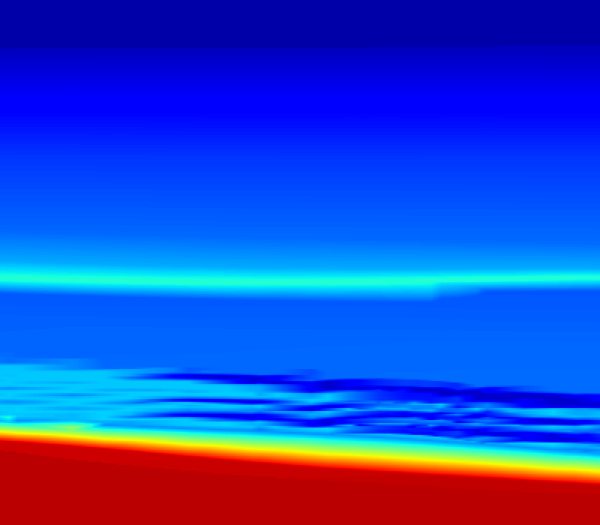} \\ \vspace{5pt}
        \includegraphics[width=0.98\linewidth]{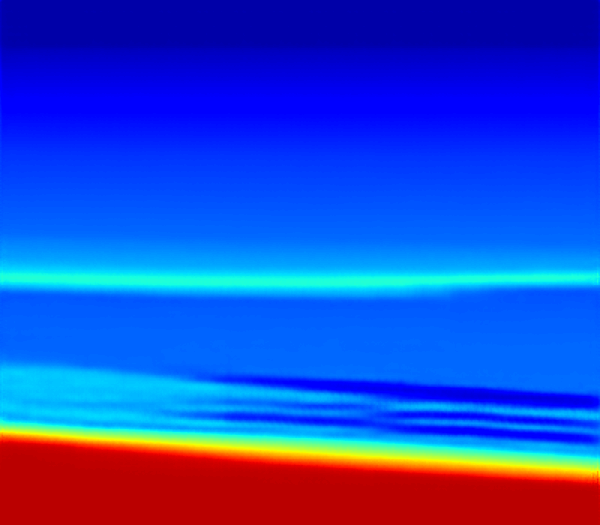} \\ \vspace{5pt}
        \includegraphics[width=0.98\linewidth]{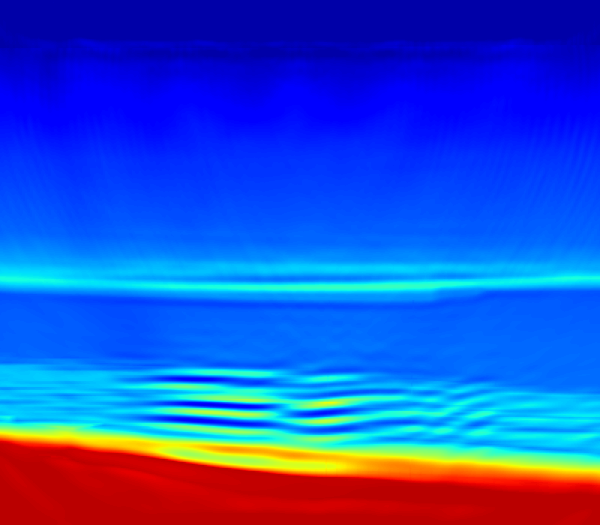} 
        \end{minipage}
    }
    \subfloat[]{
        \begin{minipage}{0.18\linewidth}
        \centering
        \includegraphics[width=0.98\linewidth]{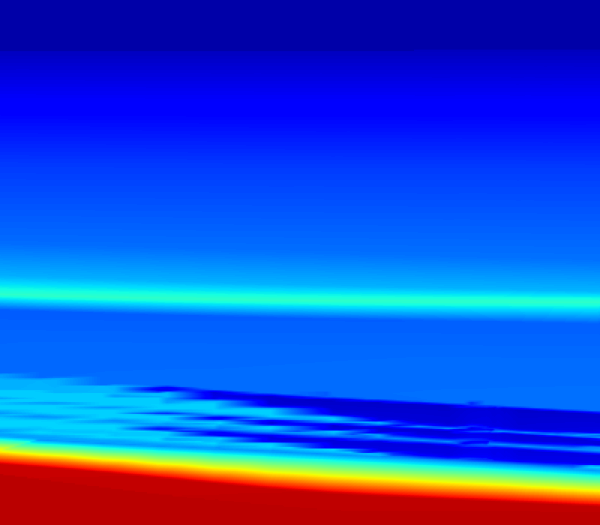} \\ \vspace{5pt}
        \includegraphics[width=0.98\linewidth]{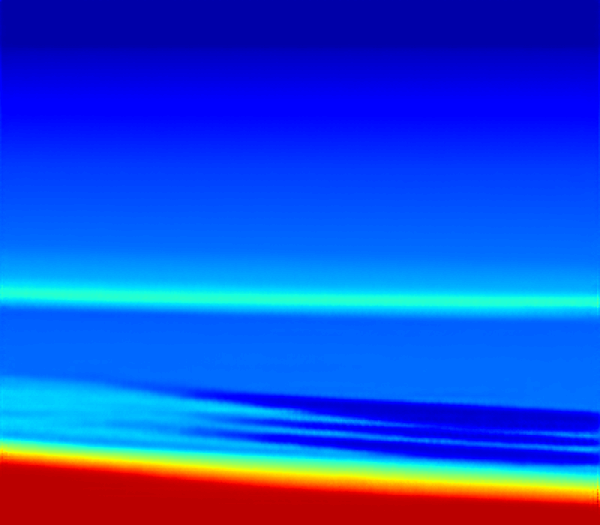} \\ \vspace{5pt}
        \includegraphics[width=0.98\linewidth]{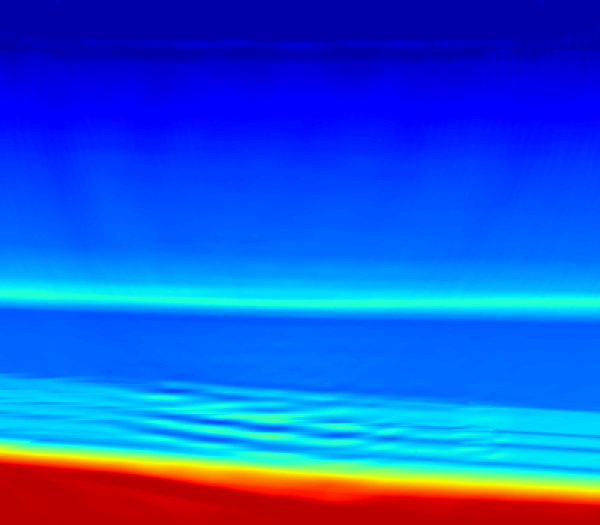} 
        \end{minipage}
    }
    \begin{minipage}{0.06\linewidth}
    \centering
    \includegraphics[width=0.95\linewidth]{Figure/color_bar.png}
    \end{minipage}
    
    \begin{minipage}{0.95\linewidth}
    \centering
    \caption{\textbf{Visualization of ground truth 2D slice sample (first row), corresponding prediction generated by InvNet3Dx1 and physics-based method (second and third row).} Each column presents one sample. }
    \label{fig:vis-phys}
    \end{minipage}
\vspace{-0.75em}
\end{figure*}

\begin{table}[]
\centering
\begin{minipage}{1.0\linewidth}
\caption{\textbf{Testing performance of InvNet3Dx1 on out-of-distribution (OOD) samples}. OOD samples are generated with a timestamp which is not used in the generation of default datasets.}
\label{tab:as-ood}
\end{minipage}
\begin{tabular}{l|c|c|c|c|c}
\hline
        &\#Train &\#Test & MAE $\downarrow$   & RMSE $\downarrow$  & SSIM $\uparrow$   \\ \hline
default (overall) &1663 &164 & 9.83  & 26.11 & 0.9826 \\
default (year 150) &53 &7 & 10.43 &28.36 &0.9809 \\
default (year 200) &58 &2 & 9.68  &25.95 &0.9833 \\
OOD (year 175)    &0 &49 & 15.22 & 37.41 & 0.9731 \\ \hline
\end{tabular}
\vspace{-0.75em}
\end{table}

There has been a long-term expectation that deep networks trained within an experimental environment can solve real-world FWI. Although this work is not particularly designed for high ability of out-of-distribution (OOD) generalization, we evaluate the pre-trained InvNet3Dx1 model on testing samples generated with an unused timestamp to provide basic insights of the generalizability of InvNet3D. We select testing samples correspond to year 175, whose nearest neighboring timestamps in the default datasets used in the main experiments are year~150 and year~200. Since the distances between these two adjacent timestamps to year 175 are fairly large, we believe this testing environment is very different from the default one.

Table~\ref{tab:as-ood} reports the results, where InvNet3Dx1 does perform slightly worse on out-of-distribution samples as expected. However, such a performance degradation is insignificant on any of the three criteria, which indicates that InvNet3Dx1 can generalize to out-of-distribution data to a certain degree. Figure~\ref{fig:vis-as-ood} visualizes randomly selected examples to further demonstrate that an acceptable reconstruction quality is maintained in an unseen environment.

\section{Conclusion}
\label{sec:conclusion}
We developed an efficient and scalable encoder-decoder network, InversionNet3D, for 3D full-waveform inversion. This model is partially reversible due to the use of invertible modules, which enables training very deep networks on limited memory. The encoder of this model is built upon group convolution and thus is able to learn representations from seismic data resulted from multiple sources with better hierarchical information flow, a smaller amount of parameters, and lower computational cost. We evaluate the model on the 3D Kimberlina dataset and demonstrate its superior performance over the baseline. We believe our InversionNet3D model not only serves as an effective solution for 3D full-waveform inversion, but also implies more possibilities of future deep-learning-based approaches with greater robustness, better interpretability, and improved generalization ability.


%

\appendices
\section{Comparison with Physics-based Methods}

In order to compare the data-driven FWI with the physics-based FWI, we apply 3D physics-based FWI on the Kimberlina data. The velocity map without CO$_2$ storage is used as the initial model and the propagating wavefields are simulated using finite difference method~\cite{carcione2002seismic}. The hybrid parallel implementation using MPI \cite{mpi1993} and OpenMPI \cite{gabriel04} are applied to increase the computational efficiency. However, the snapshot of the forward wavefields must be stored in the local file system due to the limitation of memory; thus the cross-correlation of the forward and backward wavefields heavily relies on the I/O (input/output) system~\cite{imbert2011tips}.  The iterative updating of the velocity map is achieved by the conjugated gradient method and each iteration costs around 33 minutes.  As the comparison with predicted velocity maps generated by InvNet3Dx1, the inverted results after 20 iterations are given in Figure~\ref{fig:vis-phys}. Different from InvNet3D, which only require the data from a few sources, the 3D physics-based FWI require dense acquisition configuration to remove the acquisition footprint. However, only 25 sources are aligned on the surface in Kimberlina dataset; thus there are clear footprint artifacts near the surface in the inverted result. Moreover, only the center parts of the velocity map can be updated by the physics-based FWI due to the illumination of the ray paths.


\section*{Acknowledgment}

This work was co-funded by the Laboratory Directed Research and Development program of LANL under project numbers 20210542MFR and 20200061DR, and by the U.S. DOE Office of Fossil Energy SMART Program. We also thank two anonymous reviewers and the Associate Editor for their constructive suggestions and comments that improved the quality of this work.

\ifCLASSOPTIONcaptionsoff
  \newpage
\fi



%
\bibliographystyle{IEEEtran}
\bibliography{IEEEabrv.bib, LANL-InvNet3D.bib}

\begin{thebibliography}{10}
\providecommand{\url}[1]{#1}
\csname url@samestyle\endcsname
\providecommand{\newblock}{\relax}
\providecommand{\bibinfo}[2]{#2}
\providecommand{\BIBentrySTDinterwordspacing}{\spaceskip=0pt\relax}
\providecommand{\BIBentryALTinterwordstretchfactor}{4}
\providecommand{\BIBentryALTinterwordspacing}{\spaceskip=\fontdimen2\font plus
\BIBentryALTinterwordstretchfactor\fontdimen3\font minus
  \fontdimen4\font\relax}
\providecommand{\BIBforeignlanguage}[2]{{%
\expandafter\ifx\csname l@#1\endcsname\relax
\typeout{** WARNING: IEEEtran.bst: No hyphenation pattern has been}%
\typeout{** loaded for the language `#1'. Using the pattern for}%
\typeout{** the default language instead.}%
\else
\language=\csname l@#1\endcsname
\fi
#2}}
\providecommand{\BIBdecl}{\relax}
\BIBdecl

\bibitem{tarantola2005inverse}
A.~Tarantola, \emph{Inverse Problem Theory and Methods for Model Parameter
  Estimation}.\hskip 1em plus 0.5em minus 0.4em\relax {SIAM}, 2005.

\bibitem{feng2019transmission+}
S.~Feng and G.~T. Schuster, ``Transmission+ reflection anisotropic
  wave-equation traveltime and waveform inversion,'' \emph{Geophysical
  Prospecting}, vol.~67, no.~2, pp. 423--442, 2019.

\bibitem{virieux2009overview}
J.~Virieux and S.~Operto, ``An overview of full-waveform inversion in
  exploration geophysics,'' \emph{Geophysics}, vol.~74, no.~6, pp. WCC1--WCC26,
  2009.

\bibitem{datta2016estimating}
D.~Datta and M.~K. Sen, ``Estimating a starting model for full-waveform
  inversion using a global optimization method,'' \emph{Geophysics}, vol.~81,
  no.~4, pp. R211--R223, 2016.

\bibitem{mazzotti2016two}
A.~Mazzotti, N.~Bienati, E.~Stucchi, A.~Tognarelli, M.~Aleardi, and A.~Sajeva,
  ``Two-grid genetic algorithm full-waveform inversion,'' \emph{The Leading
  Edge}, vol.~35, no.~12, pp. 1068--1075, 2016.

\bibitem{zheng20183d}
Y.~Zheng, Y.~Wang, and X.~Chang, ``{{3D}} forward modeling of upgoing and
  downgoing wavefields using {{Hilbert}} transform,'' \emph{Geophysics},
  vol.~83, no.~1, pp. F1--F8, 2018.

\bibitem{wang20193d}
Z.-Y. Wang, J.-P. Huang, D.-J. Liu, Z.-C. Li, P.~Yong, and Z.-J. Yang, ``{{3D}}
  variable-grid full-waveform inversion on {{GPU}},'' \emph{Petroleum Science},
  vol.~16, no.~5, pp. 1001--1014, 2019.

\bibitem{sajeva2016estimation}
A.~Sajeva, M.~Aleardi, E.~Stucchi, N.~Bienati, and A.~Mazzotti, ``Estimation of
  acoustic macro models using a genetic full-waveform inversion: Applications
  to the marmousi modelgenetic fwi for acoustic macro models,''
  \emph{Geophysics}, vol.~81, no.~4, pp. R173--R184, 2016.

\bibitem{gatys2015}
L.~A. Gatys, A.~S. Ecker, and M.~Bethge, ``A {{Neural Algorithm}} of {{Artistic
  Style}},'' \emph{arXiv:1508.06576}, Sep. 2015.

\bibitem{gatys2016}
------, ``Image {{Style Transfer Using Convolutional Neural Networks}},'' in
  \emph{IEEE Conf. Comput. Vis. Pattern Recog.}\hskip 1em plus 0.5em minus
  0.4em\relax {Las Vegas, NV, USA}: {IEEE}, 2016, pp. 2414--2423.

\bibitem{luan2017}
F.~Luan, S.~Paris, E.~Shechtman, and K.~Bala, ``Deep {{Photo Style
  Transfer}},'' in \emph{IEEE Conf. Comput. Vis. Pattern Recog.}\hskip 1em plus
  0.5em minus 0.4em\relax {Honolulu, HI}: {IEEE}, 2017, pp. 4990--4998.

\bibitem{johnson2016}
J.~Johnson, A.~Alahi, and L.~{Fei-Fei}, ``\BIBforeignlanguage{en}{Perceptual
  {{Losses}} for {{Real}}-{{Time Style Transfer}} and
  {{Super}}-{{Resolution}}},'' in \emph{\BIBforeignlanguage{en}{Eur. Conf.
  Comput. Vis.}}\hskip 1em plus 0.5em minus 0.4em\relax {Amsterdam, The
  Netherlands}: {Springer International Publishing}, 2016, pp. 694--711.

\bibitem{isola2017}
P.~Isola, J.-Y. Zhu, T.~Zhou, and A.~A. Efros, ``Image-{{To}}-{{Image
  Translation With Conditional Adversarial Networks}},'' in \emph{IEEE Conf.
  Comput. Vis. Pattern Recog.}\hskip 1em plus 0.5em minus 0.4em\relax
  {Honolulu, HI}: {IEEE}, 2017, pp. 1125--1134.

\bibitem{zhu2017b}
J.-Y. Zhu, R.~Zhang, D.~Pathak, T.~Darrell, A.~A. Efros, O.~Wang, and
  E.~Shechtman, ``\BIBforeignlanguage{en}{Toward {{Multimodal
  Image}}-to-{{Image Translation}}},'' in \emph{\BIBforeignlanguage{en}{Adv.
  Neural Inform. Process. Syst.}}\hskip 1em plus 0.5em minus 0.4em\relax {Long
  Beach, CA}: {Curran Associates, Inc.}, Dec. 2017, p.~12.

\bibitem{zhu2017c}
J.-Y. Zhu, T.~Park, P.~Isola, and A.~A. Efros, ``Unpaired
  {{Image}}-{{To}}-{{Image Translation Using Cycle}}-{{Consistent Adversarial
  Networks}},'' in \emph{Int. Conf. Comput. Vis.}\hskip 1em plus 0.5em minus
  0.4em\relax {Venice}: {IEEE}, 2017, pp. 2223--2232.

\bibitem{reed2016}
S.~Reed, Z.~Akata, X.~Yan, L.~Logeswaran, B.~Schiele, and H.~Lee,
  ``\BIBforeignlanguage{en}{Generative {{Adversarial Text}} to {{Image
  Synthesis}}},'' in \emph{\BIBforeignlanguage{en}{Int. Conf. Mach.
  Learn.}}\hskip 1em plus 0.5em minus 0.4em\relax {PMLR}, Jun. 2016, pp.
  1060--1069.

\bibitem{zhang2017b}
H.~Zhang, T.~Xu, H.~Li, S.~Zhang, X.~Wang, X.~Huang, and D.~Metaxas,
  ``\BIBforeignlanguage{en}{{{StackGAN}}: {{Text}} to {{Photo}}-{{Realistic
  Image Synthesis}} with {{Stacked Generative Adversarial Networks}}},'' in
  \emph{\BIBforeignlanguage{en}{Int. Conf. Comput. Vis.}}\hskip 1em plus 0.5em
  minus 0.4em\relax {Venice}: {IEEE}, Oct. 2017, pp. 5908--5916.

\bibitem{zhang2019a}
H.~Zhang, T.~Xu, H.~Li, S.~Zhang, X.~Wang, X.~Huang, and D.~N. Metaxas,
  ``{{StackGAN}}++: {{Realistic Image Synthesis}} with {{Stacked Generative
  Adversarial Networks}},'' \emph{IEEE Trans. Pattern Anal. Mach. Intell.},
  vol.~41, no.~8, pp. 1947--1962, Aug. 2019.

\bibitem{xu2018}
T.~Xu, P.~Zhang, Q.~Huang, H.~Zhang, Z.~Gan, X.~Huang, and X.~He,
  ``\BIBforeignlanguage{en}{{{AttnGAN}}: {{Fine}}-{{Grained Text}} to {{Image
  Generation}} with {{Attentional Generative Adversarial Networks}}},'' in
  \emph{\BIBforeignlanguage{en}{IEEE Conf. Comput. Vis. Pattern Recog.}}\hskip
  1em plus 0.5em minus 0.4em\relax {Salt Lake City, UT, USA}: {IEEE}, Jun.
  2018, pp. 1316--1324.

\bibitem{fabien-ouellet2019a}
G.~{Fabien-Ouellet} and R.~Sarkar, ``Seismic velocity estimation: {{A}} deep
  recurrent neural-network approach,'' \emph{Geophysics}, vol.~85, no.~1, pp.
  U21--U29, Dec. 2019.

\bibitem{li2020c}
S.~Li, B.~Liu, Y.~Ren, Y.~Chen, S.~Yang, Y.~Wang, and P.~Jiang,
  ``Deep-{{Learning Inversion}} of {{Seismic Data}},'' \emph{IEEE Trans.
  Geosci. Remote Sens.}, vol.~58, no.~3, pp. 2135--2149, Mar. 2020.

\bibitem{wang2018}
W.~Wang, F.~Yang, and J.~Ma, ``Velocity model building with a modified fully
  convolutional network,'' in \emph{SEG Tech. Prog. Exp. Abs.}, 2018, pp.
  2086--2090.

\bibitem{yang2019}
F.~Yang and J.~Ma, ``Deep-learning inversion: {{A}} next-generation seismic
  velocity model building method,'' \emph{Geophysics}, vol.~84, no.~4, pp.
  R583--R599, 2019.

\bibitem{wu2020a}
Y.~Wu and Y.~Lin, ``{{InversionNet}}: {{An Efficient}} and {{Accurate
  Data}}-{{Driven Full Waveform Inversion}},'' \emph{IEEE Trans. Comput.
  Imag.}, vol.~6, pp. 419--433, 2020.

\bibitem{zhang2019c}
Z.~Zhang, Y.~Wu, Z.~Zhou, and Y.~Lin, ``{{VelocityGAN}}: {{Subsurface Velocity
  Image Estimation Using Conditional Adversarial Networks}},'' in \emph{IEEE
  Winter Conf. Appl. Comput. Vis.}, Jan. 2019, pp. 705--714.

\bibitem{aminzadeh1996three}
F.~Aminzadeh, N.~Burkhard, J.~Long, T.~Kunz, and P.~Duclos, ``Three dimensional
  {{SEG}}/{{EAEG}} models\textemdash{{An}} update,'' \emph{The Leading Edge},
  vol.~15, no.~2, pp. 131--134, 1996.

\bibitem{vigh2014elastic}
D.~Vigh, K.~Jiao, D.~Watts, and D.~Sun, ``Elastic full-waveform inversion
  application using multicomponent measurements of seismic data collection,''
  \emph{Geophysics}, vol.~79, no.~2, pp. R63--R77, 2014.

\bibitem{guo2017elastic}
Q.~Guo and T.~Alkhalifah, ``Elastic reflection-based waveform inversion with a
  nonlinear approach,'' \emph{Geophysics}, vol.~82, no.~6, pp. R309--R321,
  2017.

\bibitem{simonyan2015}
K.~Simonyan and A.~Zisserman, ``\BIBforeignlanguage{en}{Very {{Deep
  Convolutional Networks}} for {{Large}}-{{Scale Image Recognition}}},'' in
  \emph{\BIBforeignlanguage{en}{Int. Conf. Learn. Represent.}}, May 2015.

\bibitem{he2016}
K.~He, X.~Zhang, S.~Ren, and J.~Sun, ``\BIBforeignlanguage{en}{Deep {{Residual
  Learning}} for {{Image Recognition}}},'' in
  \emph{\BIBforeignlanguage{en}{IEEE Conf. Comput. Vis. Pattern Recog.}}\hskip
  1em plus 0.5em minus 0.4em\relax {Las Vegas, NV, USA}: {IEEE}, Jun. 2016, pp.
  770--778.

\bibitem{li2020b}
Y.~Li, J.~Song, W.~Lu, P.~Monkam, and Y.~Ao, ``Multitask {{Learning}} for
  {{Super}}-{{Resolution}} of {{Seismic Velocity Model}},'' \emph{IEEE Trans.
  Geosci. Remote Sens.}, pp. 1--12, 2020.

\bibitem{leong2019}
Z.~X. Leong and T.~Zhu, ``Multi-scale deep learning seismic full waveform
  inversion,'' in \emph{{{AGU}} Fall Meeting Abstracts}, vol. 2019, Dec. 2019,
  pp. S53D--0480.

\bibitem{araya-polo2018}
M.~{Araya-Polo}, J.~Jennings, A.~Adler, and T.~Dahlke,
  ``\BIBforeignlanguage{en}{Deep-learning tomography},''
  \emph{\BIBforeignlanguage{en}{The Leading Edge}}, vol.~37, no.~1, pp. 58--66,
  Jan. 2018.

\bibitem{hochreiter1997}
S.~Hochreiter and J.~Schmidhuber, ``Long {{Short}}-{{Term Memory}},''
  \emph{Neural Computation}, vol.~9, no.~8, pp. 1735--1780, Nov. 1997.

\bibitem{zhu2020a}
W.~Zhu, K.~Xu, E.~Darve, B.~Biondi, and G.~C. Beroza, ``Integrating {{Deep
  Neural Networks}} with {{Full}}-waveform {{Inversion}}:
  {{Reparametrization}}, {{Regularization}}, and {{Uncertainty
  Quantification}},'' \emph{arXiv:2012.11149}, Dec. 2020.

\bibitem{Adler2021Deep}
A.~Adler, M.~Araya-Polo, and T.~Poggio, ``Deep learning for seismic inverse
  problems: Toward the acceleration of geophysical analysis workflows,''
  \emph{IEEE Signal Process. Mag.}, vol.~38, pp. 89--119, 2021.

\bibitem{xu2021}
Z.~Xu, A.~Desai, M.~Gupta, A.~Chandran, A.~{Vial-Aussavy}, and A.~Shrivastava,
  ``Beyond {{Convolutions}}: {{A Novel Deep Learning Approach}} for {{Raw
  Seismic Data Ingestion}},'' \emph{arXiv:2102.13631}, Feb. 2021.

\bibitem{dinh2015a}
L.~Dinh, D.~Krueger, and Y.~Bengio, ``{{NICE}}: {{Non}}-linear {{Independent
  Components Estimation}},'' in \emph{Int. Conf. Learn. Represent. Worksh.},
  May 2015.

\bibitem{dinh2017}
L.~Dinh, J.~{Sohl-Dickstein}, and S.~Bengio, ``\BIBforeignlanguage{en}{Density
  estimation using {{Real NVP}}},'' in \emph{\BIBforeignlanguage{en}{Int. Conf.
  Learn. Represent.}}, {Toulon, France}, Apr. 2017.

\bibitem{kingma2018}
D.~P. Kingma and P.~Dhariwal, ``\BIBforeignlanguage{en}{Glow: {{Generative
  Flow}} with {{Invertible}} $1\times1$ {{Convolutions}}},'' in
  \emph{\BIBforeignlanguage{en}{Adv. Neural Inform. Process. Syst.}}\hskip 1em
  plus 0.5em minus 0.4em\relax {Montr\'eal, Canada}: {Curran Associates, Inc.},
  Dec. 2018, p.~10.

\bibitem{gomez2017}
A.~N. Gomez, M.~Ren, R.~Urtasun, and R.~B. Grosse,
  ``\BIBforeignlanguage{en}{The {{Reversible Residual Network}}:
  {{Backpropagation Without Storing Activations}}},'' in
  \emph{\BIBforeignlanguage{en}{Adv. Neural Inform. Process. Syst.}},
  vol.~30.\hskip 1em plus 0.5em minus 0.4em\relax {Long Beach, CA}: {Curran
  Associates, Inc.}, 2017, pp. 2214--2224.

\bibitem{jacobsen2018}
J.-H. Jacobsen, A.~Smeulders, and E.~Oyallon,
  ``\BIBforeignlanguage{en}{I-{{RevNet}}: {{Deep Invertible Networks}}},'' in
  \emph{\BIBforeignlanguage{en}{Int. Conf. Learn. Represent.}}, {Vancouver,
  BC}, May 2018, p.~11.

\bibitem{papamkarios2017}
G.~Papamakarios, T.~Pavlakou, and I.~Murray, ``Masked autoregressive flow for
  density estimation,'' in \emph{Adv. Neural Inform. Process. Syst.},
  vol.~30.\hskip 1em plus 0.5em minus 0.4em\relax {Long Beach, CA}: {Curran
  Associates, Inc.}, Dec. 2017.

\bibitem{kingma2016}
D.~P. Kingma, T.~Salimans, R.~Jozefowicz, X.~Chen, I.~Sutskever, and
  M.~Welling, ``Improved variational inference with inverse autoregressive
  flow,'' in \emph{Adv. Neural Inform. Process. Syst.}, vol.~29.\hskip 1em plus
  0.5em minus 0.4em\relax {Curran Associates, Inc.}, 2016.

\bibitem{chen2018}
R.~T.~Q. Chen, Y.~Rubanova, J.~Bettencourt, and D.~K. Duvenaud, ``Neural
  ordinary differential equations,'' in \emph{Adv. Neural Inform. Process.
  Syst.}, vol.~31.\hskip 1em plus 0.5em minus 0.4em\relax {Montr\'eal, Canada}:
  {Curran Associates, Inc.}, 2018.

\bibitem{behrmann2019}
J.~Behrmann, W.~Grathwohl, R.~T.~Q. Chen, D.~Duvenaud, and J.-H. Jacobsen,
  ``\BIBforeignlanguage{en}{Invertible {{Residual Networks}}},'' in
  \emph{\BIBforeignlanguage{en}{Int. Conf. Mach. Learn.}}\hskip 1em plus 0.5em
  minus 0.4em\relax {PMLR}, May 2019, pp. 573--582.

\bibitem{chen2019}
R.~T.~Q. Chen, J.~Behrmann, D.~Duvenaud, and J.-H. Jacobsen,
  ``\BIBforeignlanguage{en}{Residual {{Flows}} for {{Invertible Generative
  Modeling}}},'' in \emph{\BIBforeignlanguage{en}{Adv. Neural Inform. Process.
  Syst.}}\hskip 1em plus 0.5em minus 0.4em\relax {Vancouver, Canada}: {Curran
  Associates, Inc.}, Dec. 2019, p.~11.

\bibitem{nalisnick2019}
E.~Nalisnick, A.~Matsukawa, Y.~W. Teh, D.~Gorur, and B.~Lakshminarayanan,
  ``\BIBforeignlanguage{en}{Hybrid {{Models}} with {{Deep}} and {{Invertible
  Features}}},'' in \emph{\BIBforeignlanguage{en}{Int. Conf. Mach.
  Learn.}}\hskip 1em plus 0.5em minus 0.4em\relax {PMLR}, May 2019, pp.
  4723--4732.

\bibitem{rumelhart1986}
D.~E. Rumelhart, G.~E. Hinton, and R.~J. Williams,
  ``\BIBforeignlanguage{en}{Learning representations by back-propagating
  errors},'' \emph{\BIBforeignlanguage{en}{Nature}}, vol. 323, no. 6088, pp.
  533--536, Oct. 1986.

\bibitem{brugger2019}
R.~Br{\"u}gger, C.~F. Baumgartner, and E.~Konukoglu,
  ``\BIBforeignlanguage{en}{A {{Partially Reversible U}}-{{Net}} for
  {{Memory}}-{{Efficient Volumetric Image SEGmentation}}},'' in
  \emph{\BIBforeignlanguage{en}{Med. Image Comput. Comput. Assist.
  Interv.}}\hskip 1em plus 0.5em minus 0.4em\relax {Cham}: {Springer
  International Publishing}, 2019, pp. 429--437.

\bibitem{ioffe}
S.~Ioffe and C.~Szegedy, ``\BIBforeignlanguage{en}{Batch {{Normalization}}:
  {{Accelerating Deep Network Training}} by {{Reducing Internal Covariate
  Shift}}},'' p.~9.

\bibitem{maas2013}
A.~L. Maas, A.~Y. Hannun, and A.~Y. Ng, ``\BIBforeignlanguage{en}{Rectifier
  {{Nonlinearities Improve Neural Network Acoustic Models}}},'' in
  \emph{\BIBforeignlanguage{en}{Int. Conf. Mach. Learn.}}\hskip 1em plus 0.5em
  minus 0.4em\relax {Atlanta, Georgia}: {PMLR}, Jun. 2013, p.~6.

\bibitem{howard2017}
A.~G. Howard, M.~Zhu, B.~Chen, D.~Kalenichenko, W.~Wang, T.~Weyand,
  M.~Andreetto, and H.~Adam, ``{{MobileNets}}: {{Efficient Convolutional Neural
  Networks}} for {{Mobile Vision Applications}},'' \emph{arXiv:1704.04861},
  Apr. 2017.

\bibitem{sandler2018}
M.~Sandler, A.~Howard, M.~Zhu, A.~Zhmoginov, and L.-C. Chen, ``{{MobileNetV2}}:
  {{Inverted Residuals}} and {{Linear Bottlenecks}},'' in \emph{IEEE Conf.
  Comput. Vis. Pattern Recog.}\hskip 1em plus 0.5em minus 0.4em\relax {Salt
  Lake City, UT}: {IEEE}, 2018, pp. 4510--4520.

\bibitem{zhang2018b}
X.~Zhang, X.~Zhou, M.~Lin, and J.~Sun, ``{{ShuffleNet}}: {{An Extremely
  Efficient Convolutional Neural Network}} for {{Mobile Devices}},'' in
  \emph{IEEE Conf. Comput. Vis. Pattern Recog.}\hskip 1em plus 0.5em minus
  0.4em\relax {IEEE}, 2018, pp. 6848--6856.

\bibitem{ma2018a}
N.~Ma, X.~Zhang, H.-T. Zheng, and J.~Sun, ``{{ShuffleNet V2}}: {{Practical
  Guidelines}} for {{Efficient CNN Architecture Design}},'' in \emph{Eur. Conf.
  Comput. Vis.}, 2018, pp. 116--131.

\bibitem{tan2019b}
M.~Tan, B.~Chen, R.~Pang, V.~Vasudevan, M.~Sandler, A.~Howard, and Q.~V. Le,
  ``{{MnasNet}}: {{Platform}}-{{Aware Neural Architecture Search}} for
  {{Mobile}},'' in \emph{IEEE Conf. Comput. Vis. Pattern Recog.}\hskip 1em plus
  0.5em minus 0.4em\relax {Long Beach, CA}: {IEEE}, 2019, pp. 2820--2828.

\bibitem{tan2019a}
M.~Tan and Q.~Le, ``\BIBforeignlanguage{en}{{{EfficientNet}}: {{Rethinking
  Model Scaling}} for {{Convolutional Neural Networks}}},'' in
  \emph{\BIBforeignlanguage{en}{Int. Conf. Mach. Learn.}}\hskip 1em plus 0.5em
  minus 0.4em\relax {PMLR}, May 2019, pp. 6105--6114.

\bibitem{wu2018a}
B.~Wu, A.~Wan, X.~Yue, P.~Jin, S.~Zhao, N.~Golmant, A.~Gholaminejad,
  J.~Gonzalez, and K.~Keutzer, ``\BIBforeignlanguage{en}{Shift: {{A Zero
  FLOP}}, {{Zero Parameter Alternative}} to {{Spatial Convolutions}}},'' in
  \emph{\BIBforeignlanguage{en}{IEEE Conf. Comput. Vis. Pattern Recog.}}\hskip
  1em plus 0.5em minus 0.4em\relax {Salt Lake City, UT}: {IEEE}, Jun. 2018, pp.
  9127--9135.

\bibitem{he2019}
Y.~He, X.~Liu, H.~Zhong, and Y.~Ma, ``{{AddressNet}}: {{Shift}}-{{Based
  Primitives}} for {{Efficient Convolutional Neural Networks}},'' in \emph{IEEE
  Winter Conf. Appl. Comput. Vis.}, Jan. 2019, pp. 1213--1222.

\bibitem{lin2019}
J.~Lin, C.~Gan, and S.~Han, ``{{TSM}}: {{Temporal Shift Module}} for
  {{Efficient Video Understanding}},'' in \emph{Int. Conf. Comput. Vis.}\hskip
  1em plus 0.5em minus 0.4em\relax {Seoul, Korea}: {IEEE}, 2019, pp.
  7083--7093.

\bibitem{liu2019a}
Z.~Liu, H.~Tang, Y.~Lin, and S.~Han, ``\BIBforeignlanguage{en}{Point-{{Voxel
  CNN}} for {{Efficient 3D Deep Learning}}},'' in
  \emph{\BIBforeignlanguage{en}{Adv. Neural Inform. Process. Syst.}},
  vol.~32.\hskip 1em plus 0.5em minus 0.4em\relax {Vancouver, Canada}: {Curran
  Associates, Inc.}, 2019, pp. 965--975.

\bibitem{chen2016a}
T.~Chen, B.~Xu, C.~Zhang, and C.~Guestrin, ``Training {{Deep Nets}} with
  {{Sublinear Memory Cost}},'' \emph{arXiv:1604.06174}, Apr. 2016.

\bibitem{chollet2017}
F.~Chollet, ``\BIBforeignlanguage{en}{Xception: {{Deep Learning}} with
  {{Depthwise Separable Convolutions}}},'' in
  \emph{\BIBforeignlanguage{en}{IEEE Conf. Comput. Vis. Pattern Recog.}}\hskip
  1em plus 0.5em minus 0.4em\relax {Honolulu, HI}: {IEEE}, Jul. 2017, pp.
  1800--1807.

\bibitem{sweldens1998}
W.~Sweldens, ``\BIBforeignlanguage{en}{The {{Lifting Scheme}}: {{A
  Construction}} of {{Second Generation Wavelets}}},''
  \emph{\BIBforeignlanguage{en}{SIMA}}, vol.~29, no.~2, pp. 511--546, Mar.
  1998.

\bibitem{wagoner2009}
J.~Wagoner, ``\BIBforeignlanguage{English}{{{3D Geologic Modeling}} of the
  {{Southern San Joaquin Basin}} for the {{Westcarb Kimberlina Demonstration
  Project}}- {{A Status Report}}},'' {Lawrence Livermore National Lab. (LLNL),
  Livermore, CA (United States)}, Tech. Rep. LLNL-TR-412487, Apr. 2009.

\bibitem{Wainwright2013Modeling}
H.~Wainwright, S.~Finsterle, Q.~Zhou, and J.~Birkholzer, ``Modeling the
  performance of large-scale {CO}$_2$ storage systems: A comparison of
  different sensitivity analysis methods,'' \emph{Int. J. Greenhouse Gas
  Control}, vol.~17, pp. 189--205, 2013.

\bibitem{Birkholzer2011Sensitivity}
J.~Birkholzer, Q.~Zhou, A.~Cortis, and S.~Finsterle, ``A sensitivity study on
  regional pressure buildup from large-scale {CO}$_2$ storage projects,''
  \emph{Energy Procedia}, vol.~4, pp. 4371--4378, 2011.

\bibitem{Modeling-2018-Wang}
Z.~Wang, W.~Harbert, R.~Dilmore, and L.~Huang, ``Modeling of time-lapse seismic
  monitoring using {CO}$_2$ leakage simulations for a model {CO}$_2$ storage
  site with realistic geology: application in assessment of early
  leak-detection capabilities,'' \emph{Int. J. Greenhouse Gas Control},
  vol.~76, pp. 39--52, 2018.

\bibitem{NETL-2018-Kimberlina}
{NETL}, ``{{LLNL}} kimberlina 1.2 simulations,'' 2018.

\bibitem{paszke2019}
A.~Paszke, S.~Gross, F.~Massa \emph{et~al.}, ``{{PyTorch}}: {{An Imperative
  Style}}, {{High}}-{{Performance Deep Learning Library}},'' in \emph{Adv.
  Neural Inform. Process. Syst.}\hskip 1em plus 0.5em minus 0.4em\relax
  {Vancouver, Canada}: {Curran Associates, Inc.}, Dec. 2019, pp. 8026--8037.

\bibitem{loshchilov2019}
I.~Loshchilov and F.~Hutter, ``\BIBforeignlanguage{en}{Decoupled {{Weight Decay
  Regularization}}},'' in \emph{\BIBforeignlanguage{en}{Int. Conf. Learn.
  Represent.}}, {New Orleans, LA}, May 2019.

\bibitem{zhouwang2004}
{Zhou Wang}, A.~C. Bovik, H.~R. Sheikh, and E.~P. Simoncelli, ``Image quality
  assessment: From error visibility to structural similarity,'' \emph{IEEE
  Trans. Image Process.}, vol.~13, no.~4, pp. 600--612, Apr. 2004.

\bibitem{xu2006}
K.~Xu, S.~A. Greenhalgh, and M.~Wang, ``\BIBforeignlanguage{en}{Comparison of
  source-independent methods of elastic waveform inversion},''
  \emph{\BIBforeignlanguage{en}{Geophysics}}, vol.~71, no.~6, pp. R91--R100,
  Nov. 2006.

\bibitem{song2020}
C.~Song and T.~Alkhalifah, ``\BIBforeignlanguage{en}{Source-independent
  efficient wavefield inversion},'' \emph{\BIBforeignlanguage{en}{Geophys. J.
  Int.}}, vol. 222, no.~1, pp. 697--714, Jul. 2020.

\bibitem{carcione2002seismic}
J.~M. Carcione, G.~C. Herman, and A.~Ten~Kroode, ``Seismic modeling,''
  \emph{Geophysics}, vol.~67, no.~4, pp. 1304--1325, 2002.

\bibitem{mpi1993}
C.~The MPI~Forum, ``{{MPI}}: {{A}} message passing interface,'' in
  \emph{{{ACM}}/{{IEEE Conference}} on {{Supercomputing}}}.\hskip 1em plus
  0.5em minus 0.4em\relax {New York, NY, USA}: {Association for Computing
  Machinery}, 1993, pp. 878--883.

\bibitem{gabriel04}
E.~Gabriel, G.~E. Fagg, G.~Bosilca \emph{et~al.}, ``Open {{MPI}}: {{Goals}},
  concept, and design of a next generation {{MPI}} implementation,'' in
  \emph{11th {{European PVM}}/{{MPI Users}}' {{Group Meeting}}}, {Budapest,
  Hungary}, Sep. 2004, pp. 97--104.

\bibitem{imbert2011tips}
D.~Imbert, K.~Imadoueddine, P.~Thierry, H.~Chauris, and L.~Borges, ``Tips and
  tricks for finite difference and i/o‐less fwi,'' in \emph{SEG Tech. Prog.
  Exp. Abs.}, 2012, pp. 3174--3178.

\end{thebibliography}


\begin{thebibliography}{1}
\providecommand{\url}[1]{#1}
\csname url@samestyle\endcsname
\providecommand{\newblock}{\relax}
\providecommand{\bibinfo}[2]{#2}
\providecommand{\BIBentrySTDinterwordspacing}{\spaceskip=0pt\relax}
\providecommand{\BIBentryALTinterwordstretchfactor}{4}
\providecommand{\BIBentryALTinterwordspacing}{\spaceskip=\fontdimen2\font plus
\BIBentryALTinterwordstretchfactor\fontdimen3\font minus
  \fontdimen4\font\relax}
\providecommand{\BIBforeignlanguage}[2]{{%
\expandafter\ifx\csname l@#1\endcsname\relax
\typeout{** WARNING: IEEEtran.bst: No hyphenation pattern has been}%
\typeout{** loaded for the language `#1'. Using the pattern for}%
\typeout{** the default language instead.}%
\else
\language=\csname l@#1\endcsname
\fi
#2}}
\providecommand{\BIBdecl}{\relax}
\BIBdecl

\bibitem{howard2017}
A.~G. Howard, M.~Zhu, B.~Chen, D.~Kalenichenko, W.~Wang, T.~Weyand,
  M.~Andreetto, and H.~Adam, ``{{MobileNets}}: {{Efficient Convolutional Neural
  Networks}} for {{Mobile Vision Applications}},'' \emph{arXiv:1704.04861},
  Apr. 2017.

\bibitem{sandler2018}
M.~Sandler, A.~Howard, M.~Zhu, A.~Zhmoginov, and L.-C. Chen, ``{{MobileNetV2}}:
  {{Inverted Residuals}} and {{Linear Bottlenecks}},'' in \emph{IEEE Conf.
  Comput. Vis. Pattern Recog.}\hskip 1em plus 0.5em minus 0.4em\relax {Salt
  Lake City, UT}: {IEEE}, 2018, pp. 4510--4520.

\bibitem{zhang2018b}
X.~Zhang, X.~Zhou, M.~Lin, and J.~Sun, ``{{ShuffleNet}}: {{An Extremely
  Efficient Convolutional Neural Network}} for {{Mobile Devices}},'' in
  \emph{IEEE Conf. Comput. Vis. Pattern Recog.}\hskip 1em plus 0.5em minus
  0.4em\relax {IEEE}, 2018, pp. 6848--6856.

\bibitem{ma2018a}
N.~Ma, X.~Zhang, H.-T. Zheng, and J.~Sun, ``{{ShuffleNet V2}}: {{Practical
  Guidelines}} for {{Efficient CNN Architecture Design}},'' in \emph{Eur. Conf.
  Comput. Vis.}, 2018, pp. 116--131.

\bibitem{wang2020e}
Z.~Wang, J.~Chen, and S.~C. Hoi, ``Deep {{Learning}} for {{Image
  Super}}-resolution: {{A Survey}},'' \emph{IEEE Trans. Pattern Anal. Mach.
  Intell.}, pp. 1--1, 2020.

\bibitem{zhu2021b}
M.~Zhu, D.~He, X.~Li, C.~Li, F.~Li, X.~Liu, E.~Ding, and Z.~Zhang, ``Image
  {{Inpainting}} by {{End}}-to-{{End Cascaded Refinement With Mask
  Awareness}},'' \emph{IEEE Trans. Image Process.}, vol.~30, pp. 4855--4866,
  2021.

\bibitem{cheng2017}
Z.~Cheng, Q.~Yang, and B.~Sheng, ``Colorization {{Using Neural Network
  Ensemble}},'' \emph{IEEE Trans. Image Process.}, vol.~26, no.~11, pp.
  5491--5505, Nov. 2017.

\end{thebibliography}

%








\end{document}


\setcounter{table}{0}
\setcounter{figure}{0}
%
\title{InversionNet3D: Efficient and Scalable Learning for 3D Full Waveform Inversion \\ \LARGE{Supplementary Materials}}
%
%
%

\author{Qili~Zeng, Shihang~Feng,  Brendt~Wohlberg,~\IEEEmembership{Senior~Member,~IEEE}, and Youzuo Lin,~\IEEEmembership{Member,~IEEE}
}

%
%

\markboth{IEEE Transactions on Geoscience and Remote Sensing}%
{Zeng \MakeLowercase{\textit{et al.}}: InversionNet3D: Efficient and Scalable Learning for 3D Full Waveform Inversion}
%



\maketitle
\IEEEpeerreviewmaketitle

Here, we provide additional technical details to help with understanding our proposed methods. It includes: (1) more elaboration on the group convolution~(Section~\ref{sec:group}); (2) focused visualization of the inversion results~(Section~\ref{sec:zoom-vis}); (3) breakdown of the training and validation loss curves of different testing models~(Section~\ref{sec:loss}); (4) visualization of various source wavelets used in the test~(Section~\ref{sec:source}); and (5) illumination analysis for the selected source  strategy~(Section~\ref{sec:illumination}).

\begin{figure}[ht]
\renewcommand{\thefigure}{2A}
\centering
\includegraphics[width=0.98\linewidth]{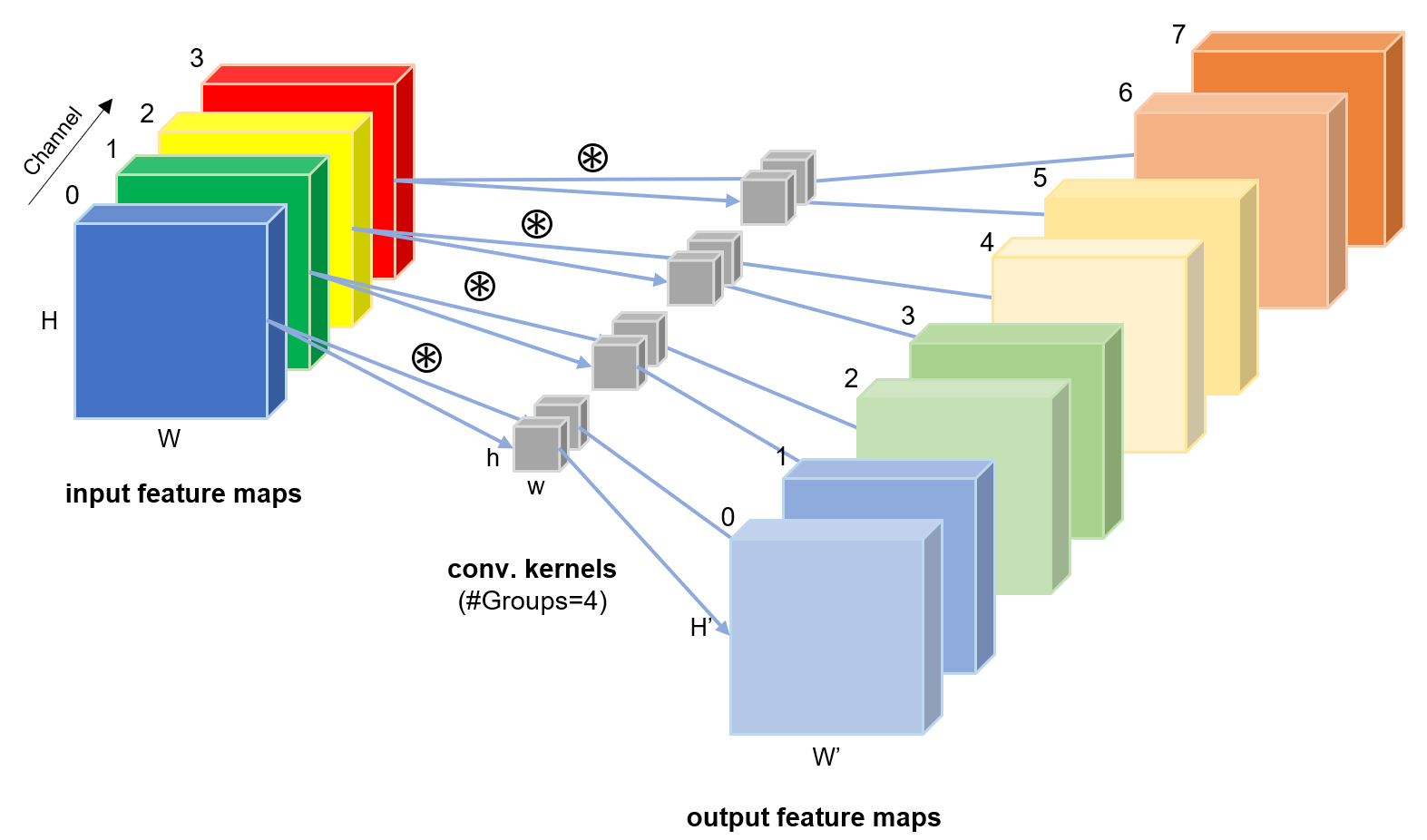} 
\caption{\textbf{Group Convolution.} This example corresponds to the first group convolution in Figure 2(b) in the main paper, where the input channel size, the output channel, and the group size are 4, 8, and 4, respectively.}
\label{fig:group-conv-detail}
\end{figure}

\section{Detailed diagram of group convolution}
\label{sec:group}

All the downsampling layers in the proposed Channel-Separated Encoder are built upon group convolution (GConv) \cite{howard2017, sandler2018, zhang2018b, ma2018a}. In the main paper, we provided a simple example of GConv-CS-GConv block in Figure 2, where subfigure (b) visualizes the information flow on the channel dimension. Here we provide a detailed diagram (Figure \ref{fig:group-conv-detail}) of the first GConv in Figure 2 (b), which may be helpful for broader readership to understand the details of group convolution and the design principles of our model.

\section{Zoom-in Visualizations for main experiments}
\label{sec:zoom-vis}

In the main paper, we provided the whole image of 2D velocity models in Figures~7, 9, and 10 and highlighted the areas of interest with white dashed-line rectangles for better comparisons. By doing so, we believe a balance between the preservation of intact information and an emphasis on the difference was achieved. However, it is also known to us that a zoom-in display of only the highlighted areas of difference would better serve the interest of certain readers. Therefore, we provide Figures~\ref{fig:vis-baseline-zoom}, \ref{fig:vis-nblocks-zoom} and \ref{fig:vis-as-temporal-zoom} corresponding to Figure 7, 9, and 10 in the main paper, respectively.

\section{Stand-alone Plots of Loss Curves}
\label{sec:loss}

Figures~5 and 8 in the main paper visualized the loss curves of the models in comparison, which, we believe, would enhance the reproducibility of our work and reveal certain information about the training and evaluation of the models. Since there is no necessity to plot the loss curves separately and there is also no enough place in the main paper to support such plots, the loss curves in Figures~5 and 8 were overlapping with each other. It will be useful for some readers to observe the loss curves of each model in a stand-alone manner and thus we provide such plots in Figure~\ref{fig:losses}.

\section{Source Wavelets}
\label{sec:source}

\begin{figure}[t]
\renewcommand{\thefigure}{19}
\centering
    \includegraphics[width=1.0\linewidth]{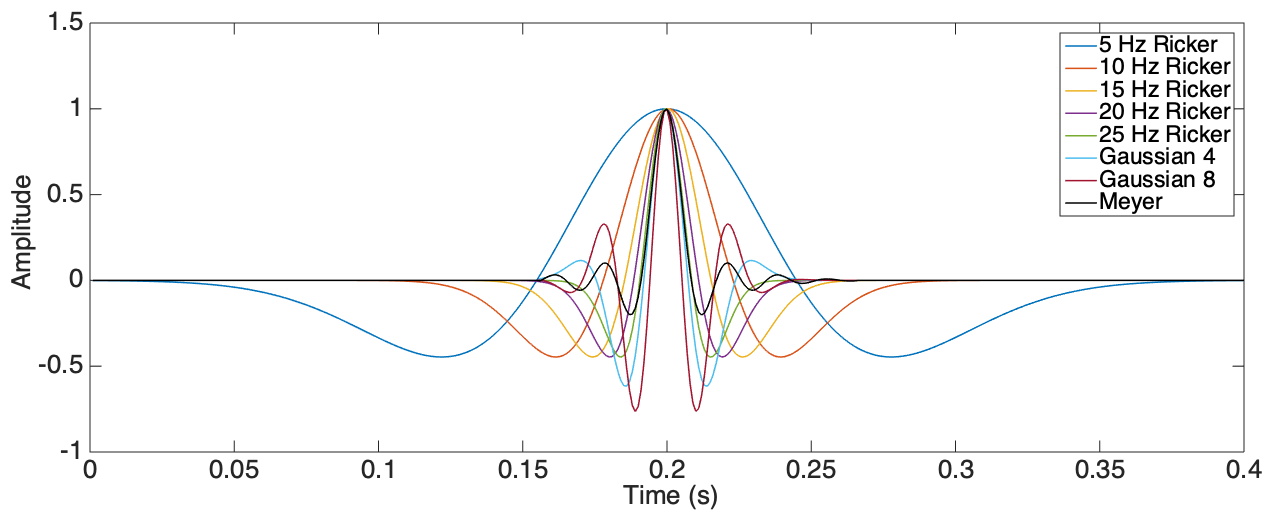} 
    \caption{\textbf{Source wavelets.} InvNet3D models in the main experiments are trained with data generated with 15Hz Ricker wavelets.}
\label{fig:source}
\end{figure}

Section V-E in the main body analyzed the adaption of a pre-trained InvNet3Dx1 into seismic data of different source signatures. Here we provide an intuitive visualization of the source wavelets in Figure~\ref{fig:source}, including 15Hz Ricker wavelets used to generate the seismic data for the main experiments and to pre-train the network.

\section{Source Illumination}
\label{sec:illumination}

\begin{figure*}[]
\renewcommand{\thefigure}{12A}
\centering
    \begin{minipage}{0.90\linewidth}
        \subfloat[$z=0$]{
            \centering
            \includegraphics[width=0.24\linewidth]{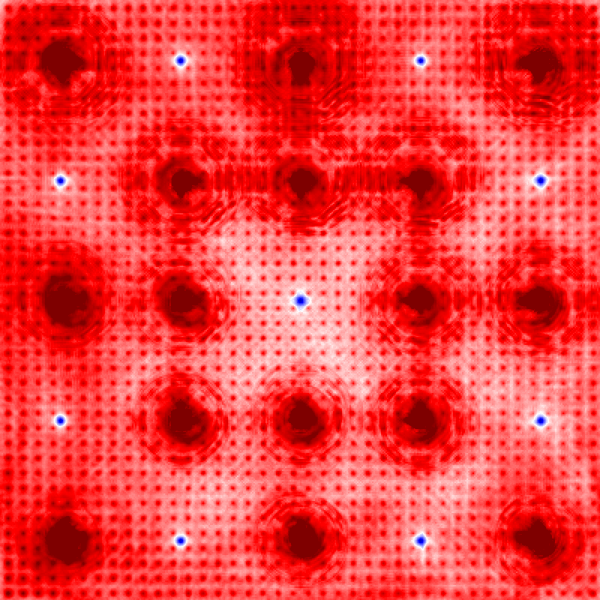}
        }
        \subfloat[$z=49$]{
            \centering
            \includegraphics[width=0.24\linewidth]{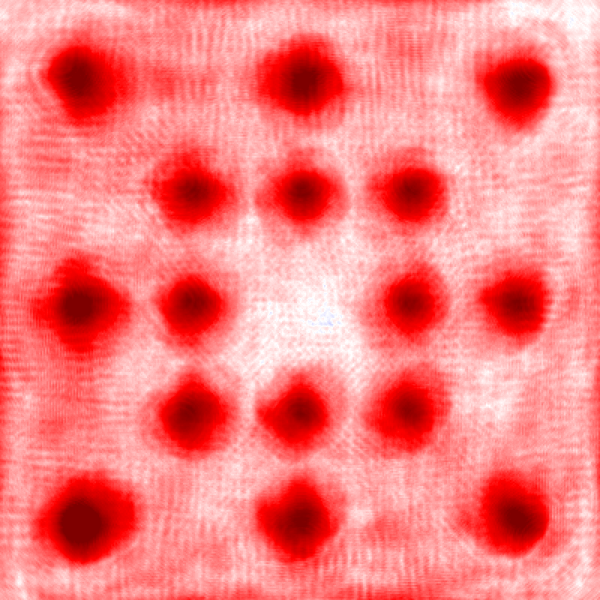} 
        }
        \subfloat[$z=99$]{ 
            \includegraphics[width=0.24\linewidth]{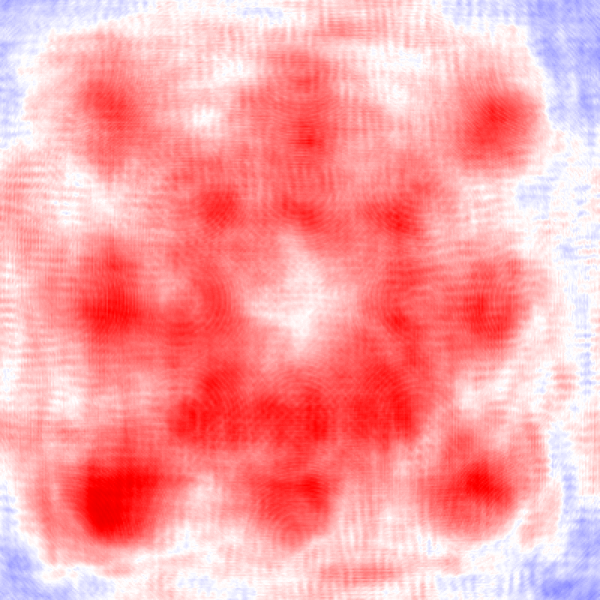} 
        }
        \subfloat[$z=149$]{ 
            \centering
            \includegraphics[width=0.24\linewidth]{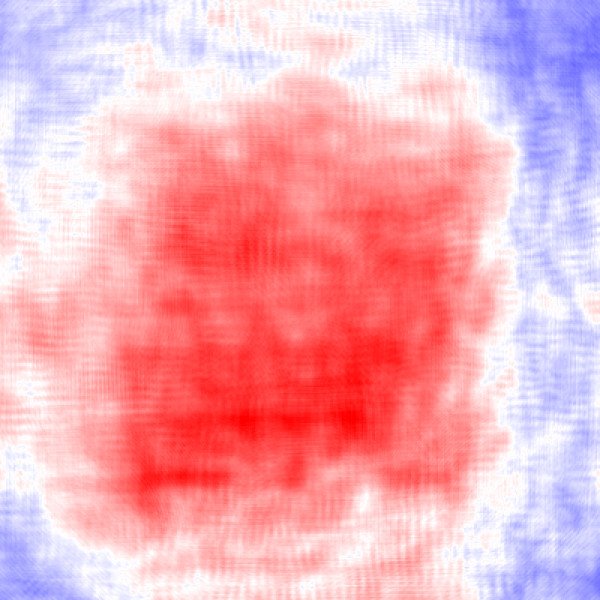}
        }\\[-2ex]%
        \subfloat[$z=199$]{ 
            \centering
            \includegraphics[width=0.24\linewidth]{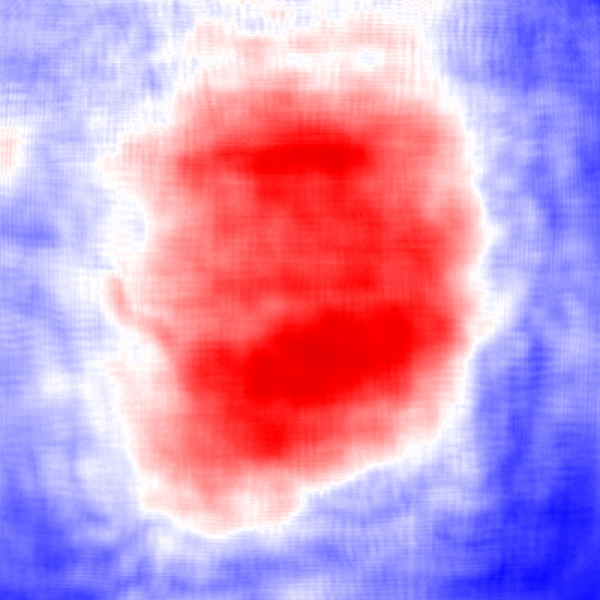} 
        }
         \subfloat[$z=249$]{ 
            \centering
            \includegraphics[width=0.24\linewidth]{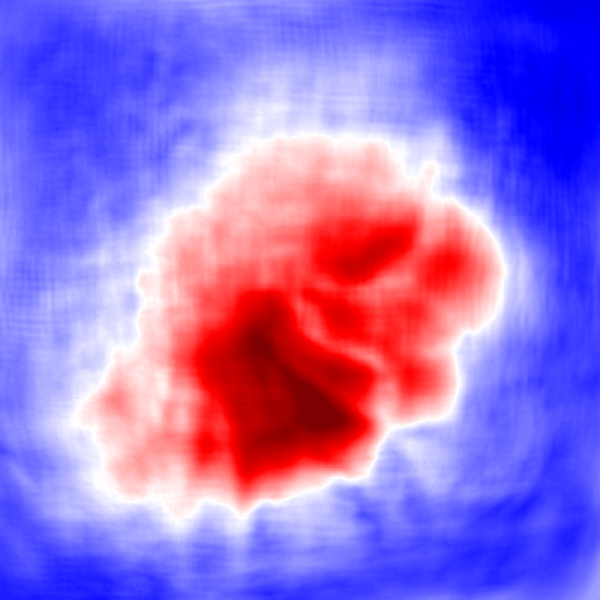} 
        }
        \subfloat[$z=299$]{
            \centering
            \includegraphics[width=0.24\linewidth]{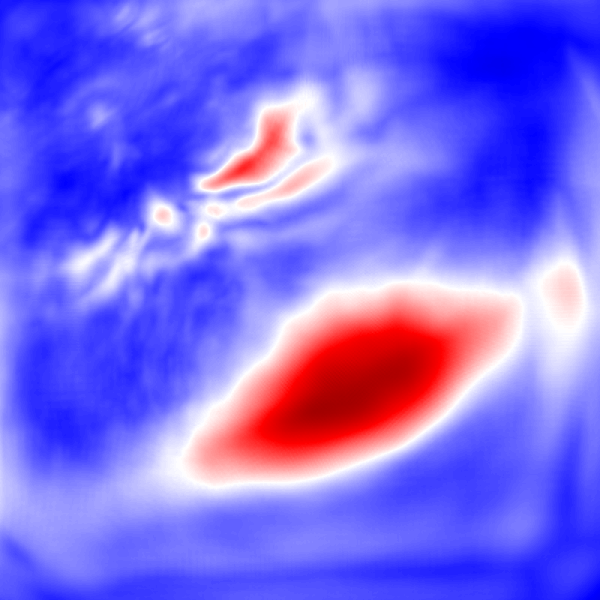}
        }
        \subfloat[$z=349$]{
            \centering
            \includegraphics[width=0.24\linewidth]{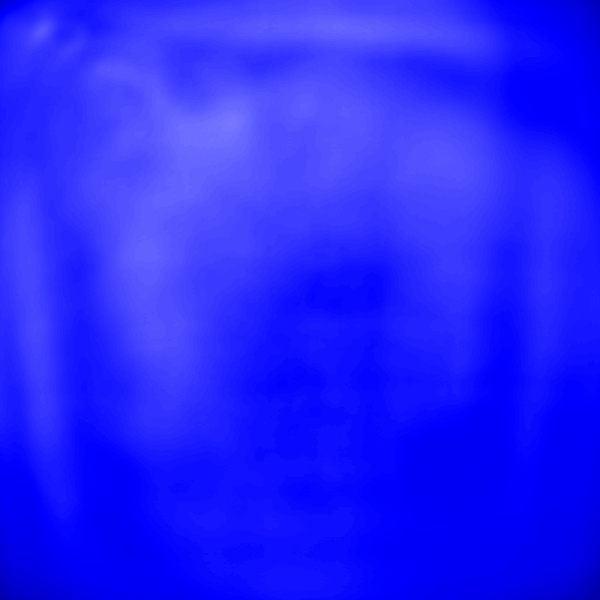} 
        }
    \end{minipage}
    \hfill
    \begin{minipage}{0.08\linewidth}
    \centering
    \includegraphics[width=0.95\linewidth]{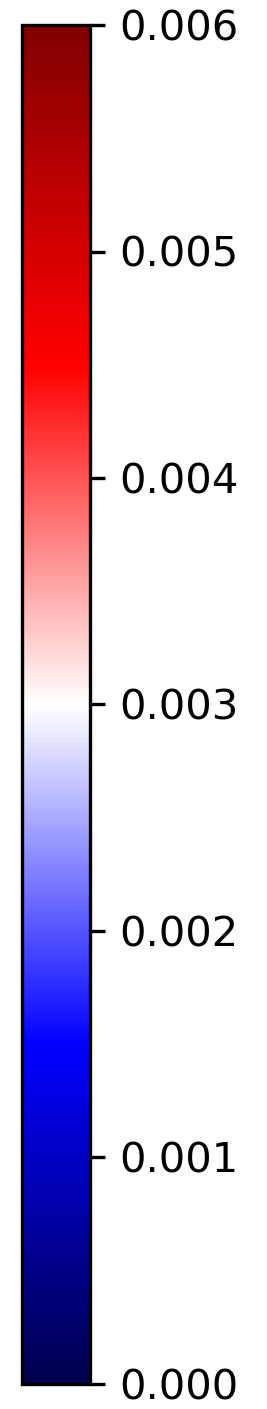}
    \end{minipage}
    
    \begin{minipage}{0.96\linewidth}
    \centering
    \caption{\textbf{Source illumination of a 16-source placement scheme.} This scheme sample corresponds to the strategy described in the top-left sub-figure of Figure 12. We uniformly extract 8 spatial ($x$-$y$) slices on the depth ($z$) dimension. }
    \label{fig:vis-c16-illum}
    \end{minipage}
\vspace{-0.75em}
\end{figure*}

\begin{figure*}[]
\renewcommand{\thefigure}{12B}
\centering
    \begin{minipage}{0.90\linewidth}
        \subfloat[$z=0$]{
            \centering
            \includegraphics[width=0.24\linewidth]{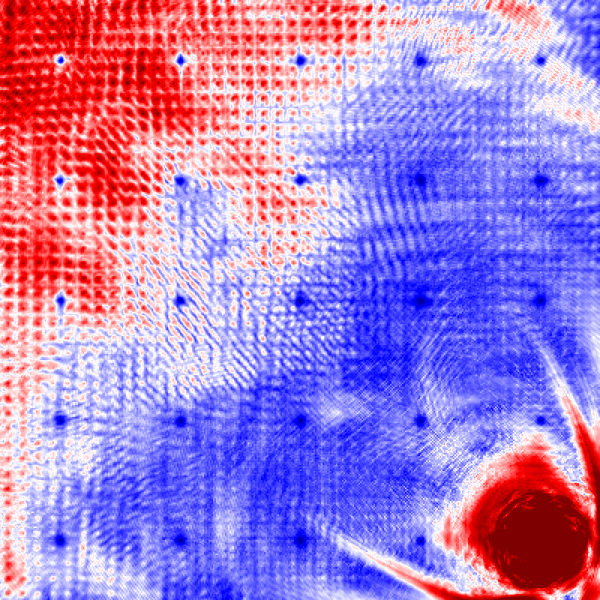}
        }
        \subfloat[$z=49$]{
            \centering
            \includegraphics[width=0.24\linewidth]{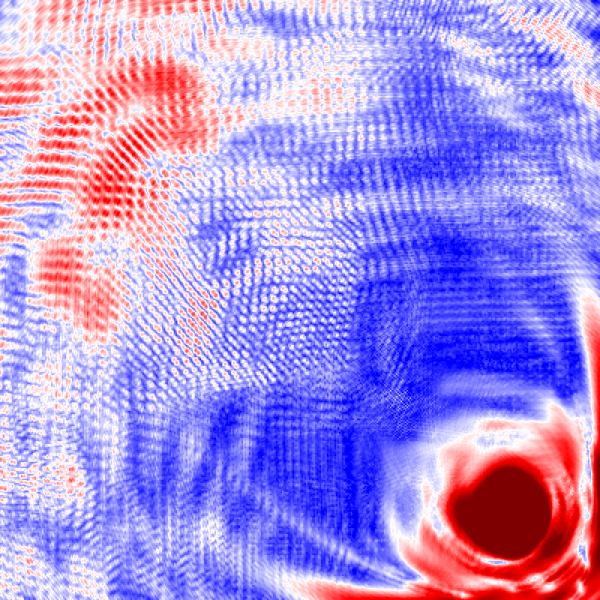} 
        }
        \subfloat[$z=99$]{ 
            \includegraphics[width=0.24\linewidth]{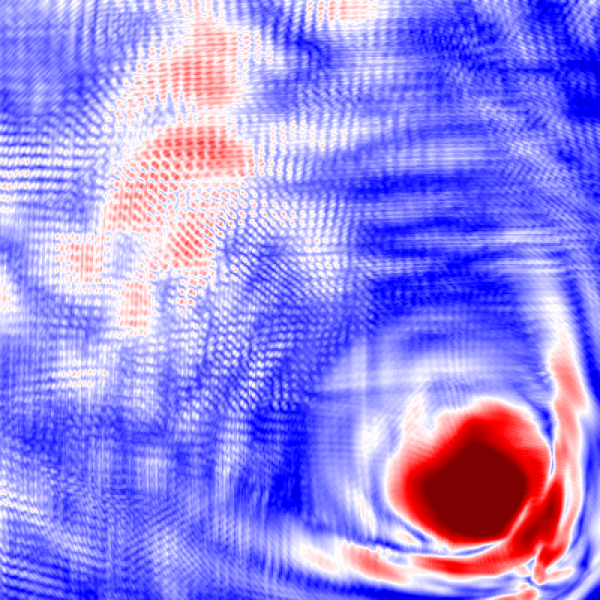} 
        }
        \subfloat[$z=149$]{ 
            \centering
            \includegraphics[width=0.24\linewidth]{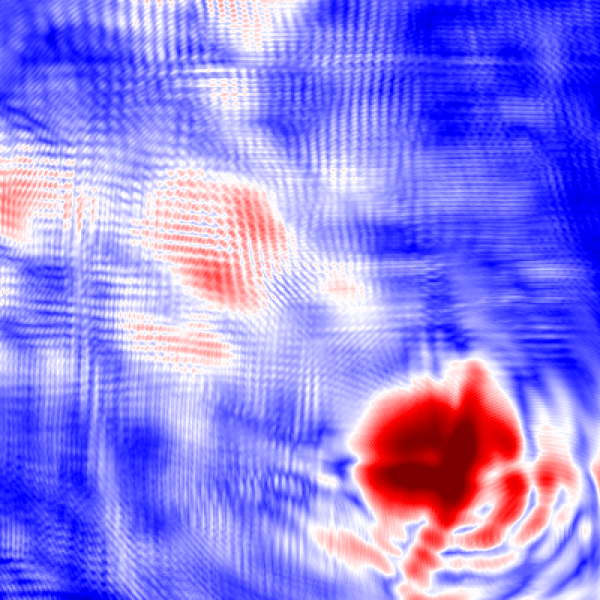}
        }\\[-2ex]%
        \subfloat[$z=199$]{ 
            \centering
            \includegraphics[width=0.24\linewidth]{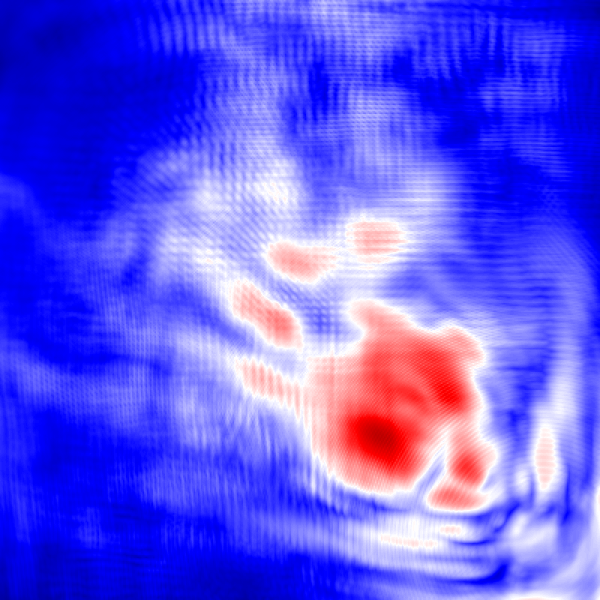} 
        }
         \subfloat[$z=249$]{ 
            \centering
            \includegraphics[width=0.24\linewidth]{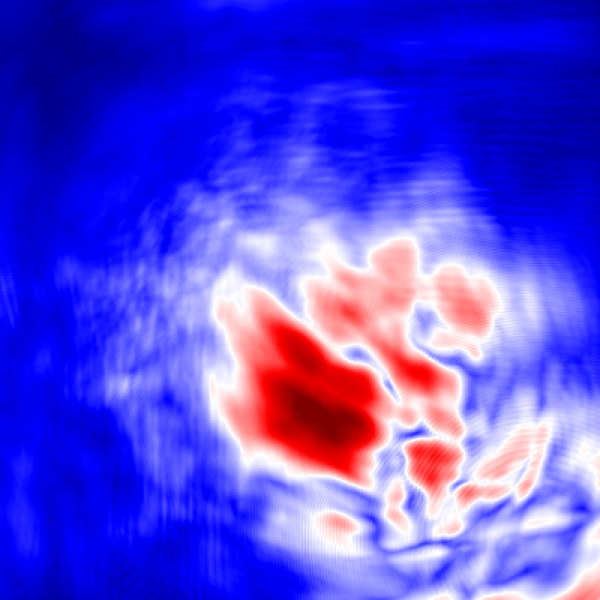} 
        }
        \subfloat[$z=299$]{
            \centering
            \includegraphics[width=0.24\linewidth]{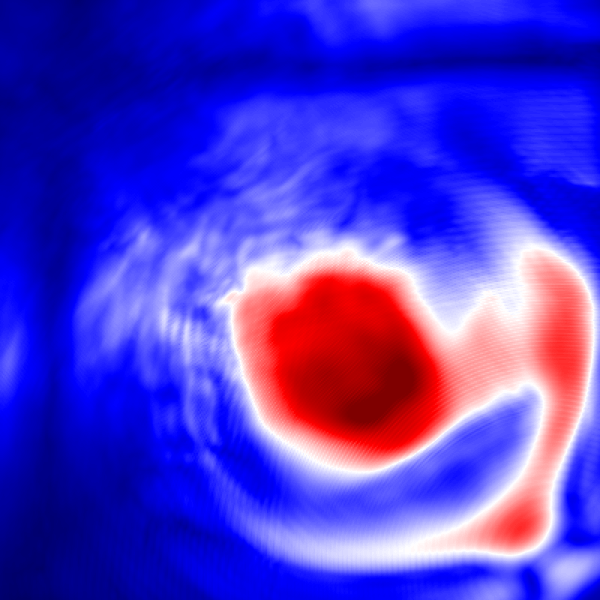}
        }
        \subfloat[$z=349$]{
            \centering
            \includegraphics[width=0.24\linewidth]{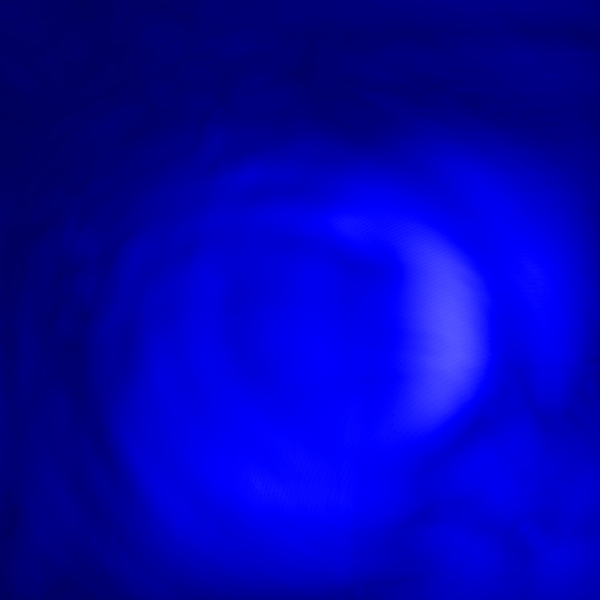} 
        }
    \end{minipage}
    \hfill
    \begin{minipage}{0.08\linewidth}
    \centering
    \includegraphics[width=0.95\linewidth]{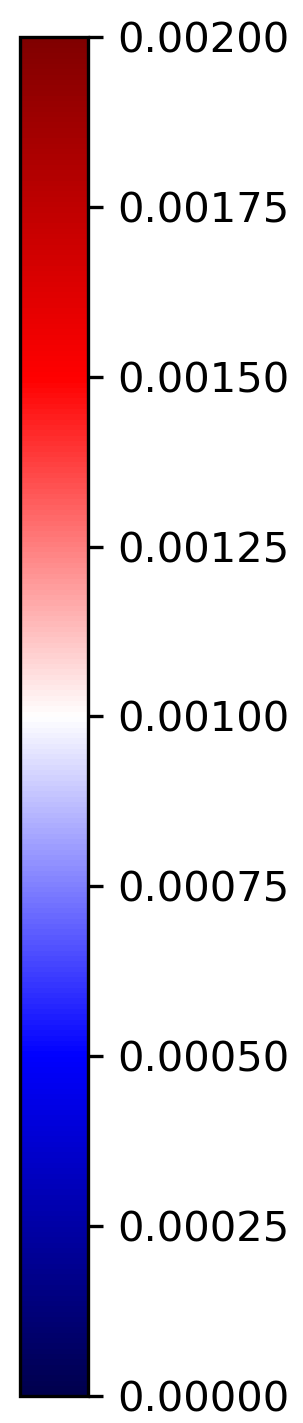}
    \end{minipage}
    
    \begin{minipage}{0.96\linewidth}
    \centering
    \caption{\textbf{Source illumination of a single-source placement scheme.} This scheme sample corresponds to the strategy described in the bottom-right sub-figure of Figure 12. We uniformly extract 8 spatial ($x$-$y$) slices on the depth ($z$) dimension. }
    \label{fig:vis-c1-illum}
    \end{minipage}
\vspace{-0.75em}
\end{figure*}

\begin{figure*}[]
\renewcommand{\thefigure}{7A}
\centering
    \subfloat[]{
        \begin{minipage}{0.18\linewidth}
        \centering
        \includegraphics[width=0.98\linewidth]{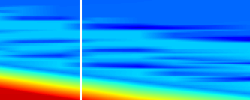} \\ \vspace{5pt}
        \includegraphics[width=0.98\linewidth]{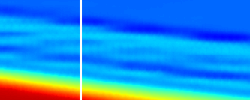} \\ \vspace{5pt}
        \includegraphics[width=0.98\linewidth]{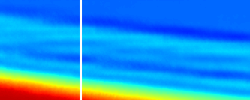} \\ \vspace{5pt}
        \includegraphics[width=0.98\linewidth]{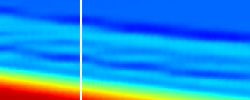} \\ \vspace{5pt}
        \includegraphics[width=0.98\linewidth]{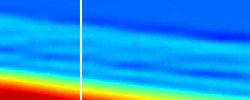}  \\\vspace{5pt}
        \includegraphics[width=0.98\linewidth]{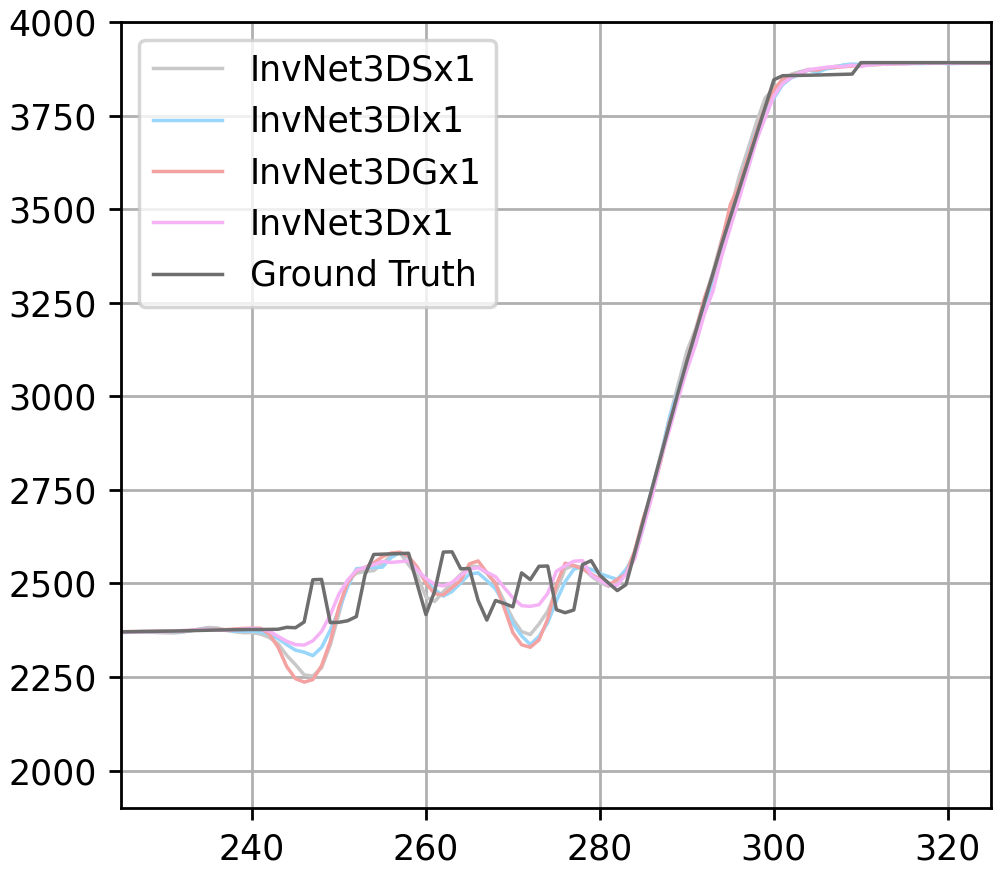} 
        \label{fig:vis-baseline-1}
        \end{minipage}
    }
    \subfloat[]{
        \begin{minipage}{0.18\linewidth}
        \centering
        \includegraphics[width=0.98\linewidth]{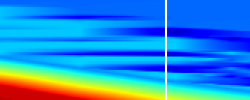}  \\\vspace{5pt}
        \includegraphics[width=0.98\linewidth]{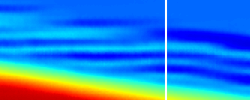}  \\\vspace{5pt}
        \includegraphics[width=0.98\linewidth]{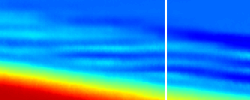}  \\\vspace{5pt}
        \includegraphics[width=0.98\linewidth]{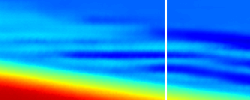}  \\\vspace{5pt}
        \includegraphics[width=0.98\linewidth]{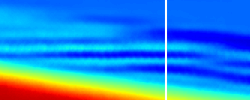} \\\vspace{5pt}
        \includegraphics[width=0.98\linewidth]{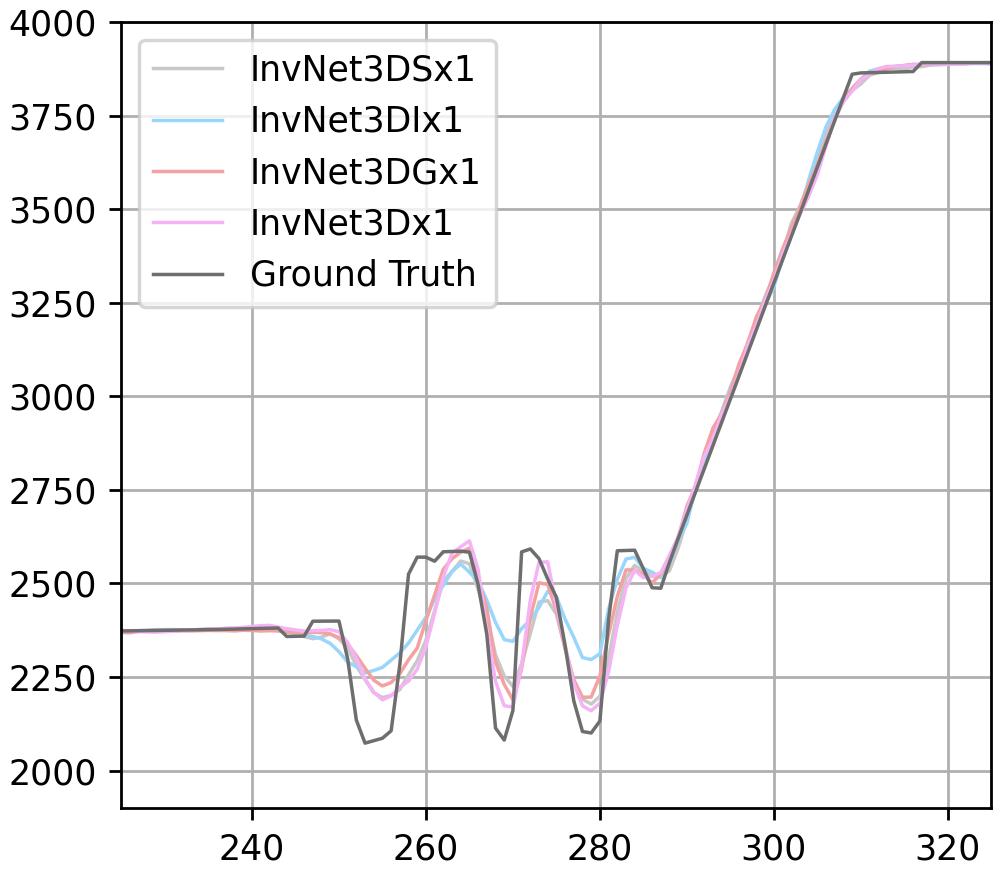} 
        \label{fig:vis-baseline-2}
        \end{minipage}
    }
    \subfloat[]{
        \begin{minipage}{0.18\linewidth}
        \centering
        \includegraphics[width=0.98\linewidth]{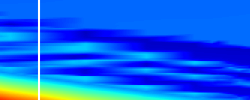}   \\\vspace{5pt}
        \includegraphics[width=0.98\linewidth]{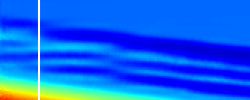}  \\\vspace{5pt}
        \includegraphics[width=0.98\linewidth]{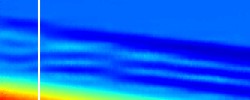}  \\\vspace{5pt}
        \includegraphics[width=0.98\linewidth]{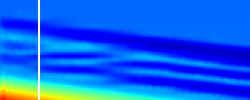}  \\\vspace{5pt}
        \includegraphics[width=0.98\linewidth]{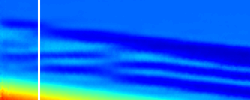} \\\vspace{5pt}
        \includegraphics[width=0.98\linewidth]{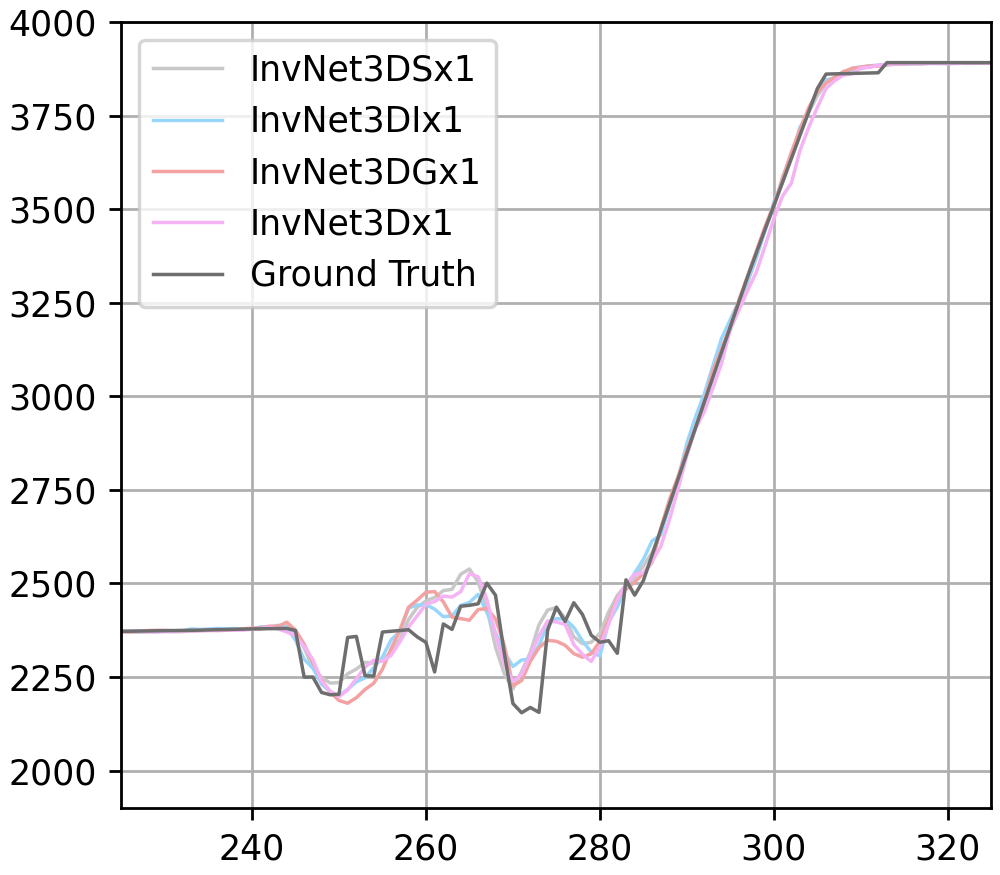}
        \label{fig:vis-baseline-3}
        \end{minipage}
    }
    \subfloat[]{
        \begin{minipage}{0.18\linewidth}
        \centering
        \includegraphics[width=0.98\linewidth]{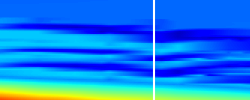}  \\\vspace{5pt}
        \includegraphics[width=0.98\linewidth]{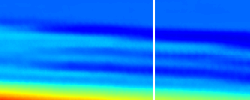}  \\\vspace{5pt}
        \includegraphics[width=0.98\linewidth]{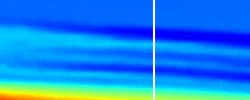}  \\\vspace{5pt}
        \includegraphics[width=0.98\linewidth]{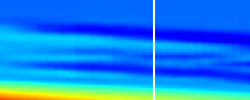}  \\\vspace{5pt}
        \includegraphics[width=0.98\linewidth]{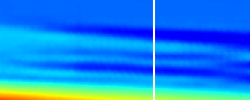} \\\vspace{5pt}
        \includegraphics[width=0.98\linewidth]{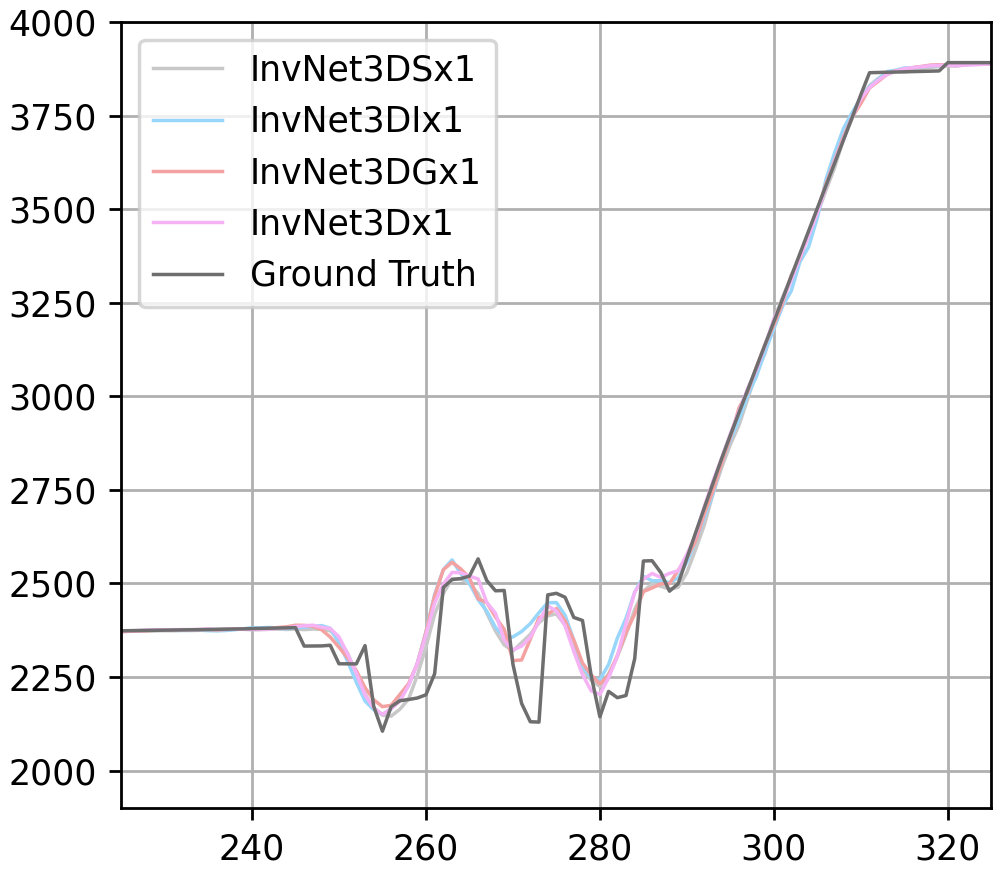}
        \label{fig:vis-baseline-4}
        \end{minipage}
    }
    \subfloat[]{
        \begin{minipage}{0.18\linewidth}
        \centering
        \includegraphics[width=0.98\linewidth]{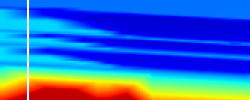} \\ \vspace{5pt}
        \includegraphics[width=0.98\linewidth]{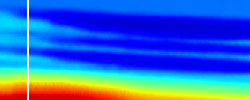}  \\\vspace{5pt}
        \includegraphics[width=0.98\linewidth]{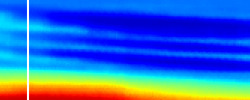}  \\\vspace{5pt}
        \includegraphics[width=0.98\linewidth]{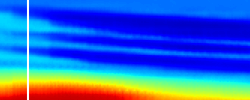}  \\\vspace{5pt}
        \includegraphics[width=0.98\linewidth]{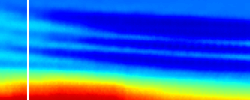} \\\vspace{5pt}
        \includegraphics[width=0.98\linewidth]{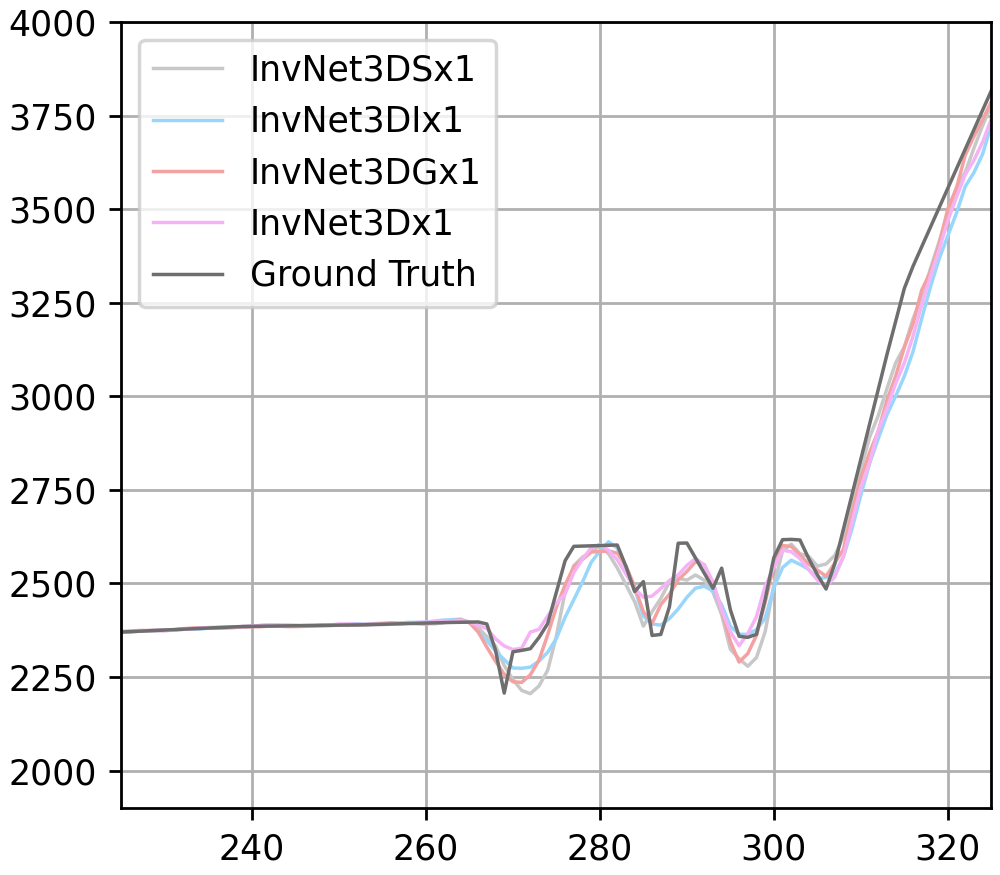}
        \label{fig:vis-baseline-5}
        \end{minipage}
    }
    \begin{minipage}{0.06\linewidth}
    \centering
    \includegraphics[width=0.95\linewidth]{Figure/color_bar.png}
    \end{minipage}
    
    \begin{minipage}{0.95\linewidth}
    \centering
    \caption{\textbf{Zoom-in visualization of ground truth velocity map (2D slice) (first row), corresponding predicted velocity map generated by InvNet3DS, InvNet3DI, InvNet3DG, and InvNet3D (second row to fifth row).} The sixth row displays vertical velocity profiles (velocity - depth) at the location of the white cursor on the 2D slices. These profiles are uniformly truncated to a depth range from 225 to 325, which covers the depth range of the zoom-in plots. All the models are built with 26 layers. Each column presents one sample.}
    \label{fig:vis-baseline-zoom}
    \end{minipage}
\end{figure*}

\begin{figure*}[]
\renewcommand{\thefigure}{9A}
\centering
    \subfloat[]{
        \begin{minipage}{0.18\linewidth}
        \centering
        \includegraphics[width=0.98\linewidth]{Figure/vis_zoom/year30_cut37_x108/4-InvNet3DSRCx1.png} \\\vspace{5pt}
        \includegraphics[width=0.98\linewidth]{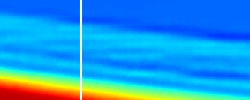} \\\vspace{5pt}
        \includegraphics[width=0.98\linewidth]{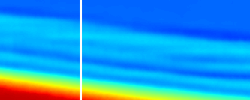} \\\vspace{5pt}
        \includegraphics[width=0.98\linewidth]{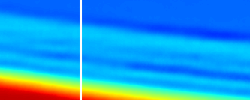} \\\vspace{5pt}
        \includegraphics[width=0.98\linewidth]{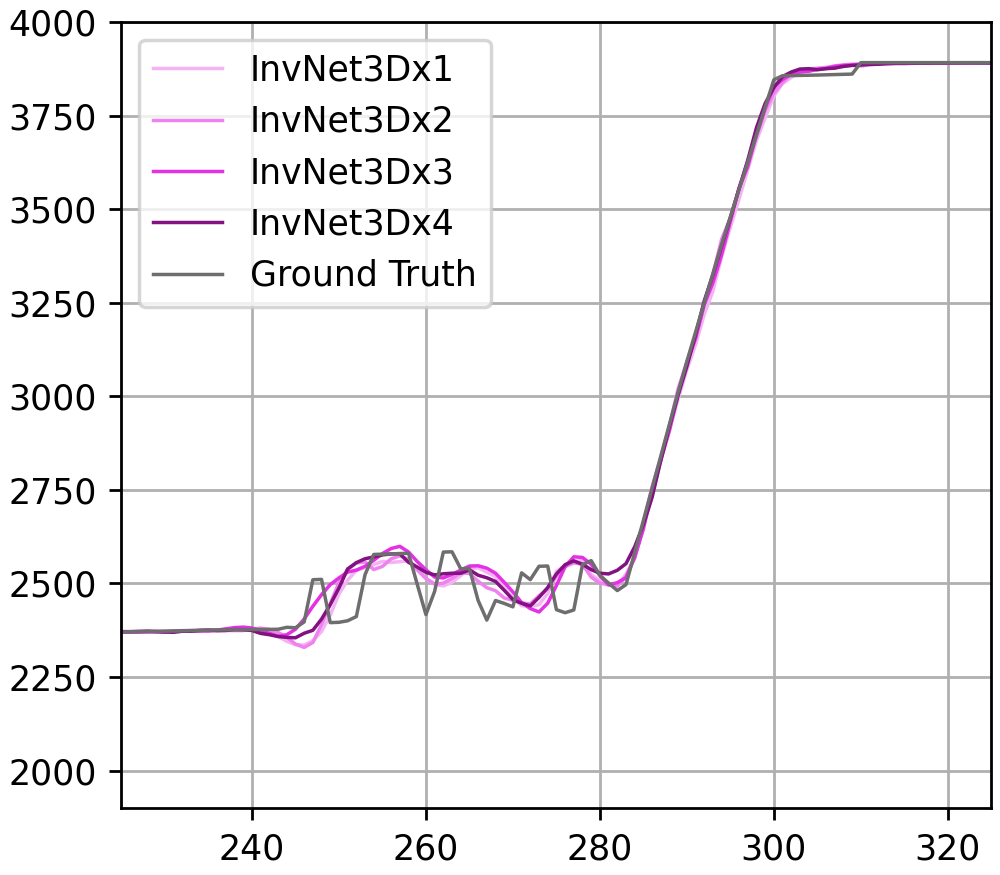}
        \end{minipage}
    }
    \subfloat[]{
        \begin{minipage}{0.18\linewidth}
        \centering
        \includegraphics[width=0.98\linewidth]{Figure/vis_zoom/year25_cut7_x135/4-InvNet3DSRCx1.png} \\\vspace{5pt}
        \includegraphics[width=0.98\linewidth]{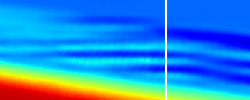} \\\vspace{5pt}
        \includegraphics[width=0.98\linewidth]{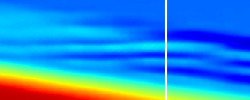} \\\vspace{5pt}
        \includegraphics[width=0.98\linewidth]{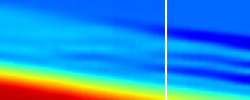} \\ \vspace{5pt}
        \includegraphics[width=0.98\linewidth]{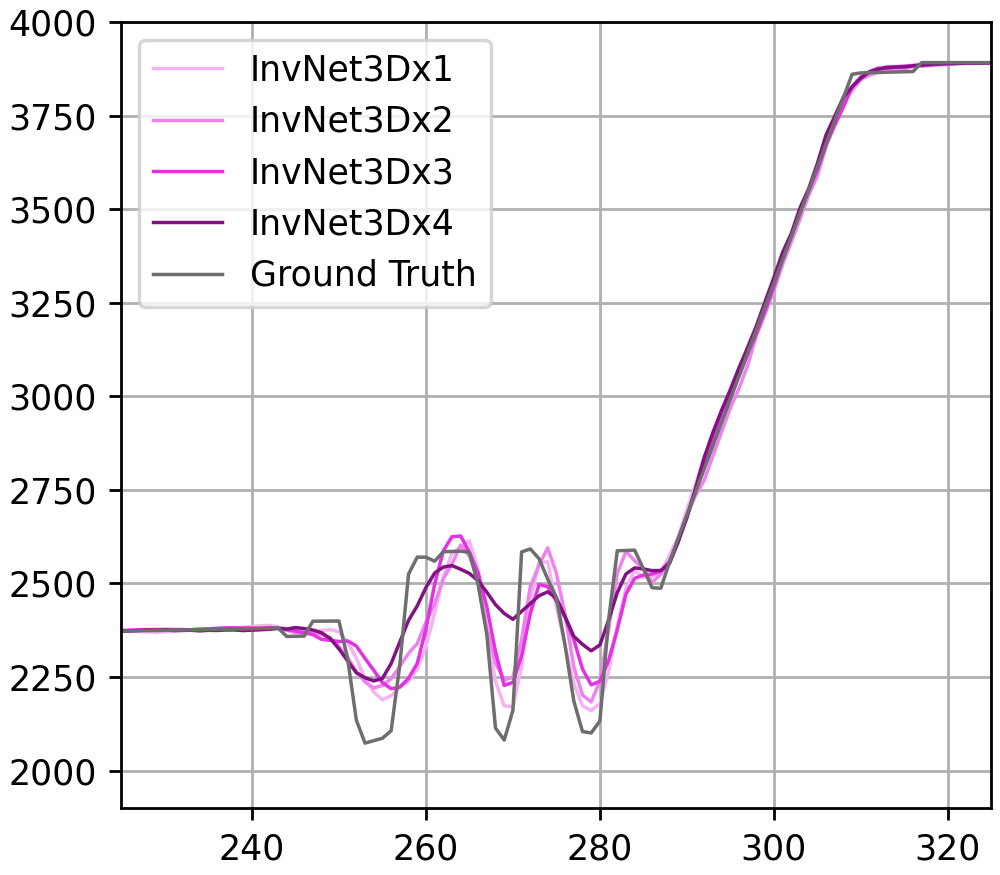}
        \end{minipage}
    }
    \subfloat[]{
        \begin{minipage}{0.18\linewidth}
        \centering
        \includegraphics[width=0.98\linewidth]{Figure/vis_zoom/year30_cut37_x275/4-InvNet3DSRCx1.png} \\\vspace{5pt}
        \includegraphics[width=0.98\linewidth]{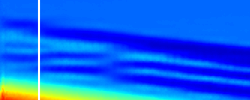} \\\vspace{5pt}
        \includegraphics[width=0.98\linewidth]{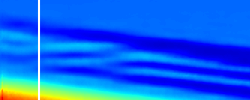} \\\vspace{5pt}
        \includegraphics[width=0.98\linewidth]{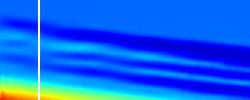} \\ \vspace{5pt}
        \includegraphics[width=0.98\linewidth]{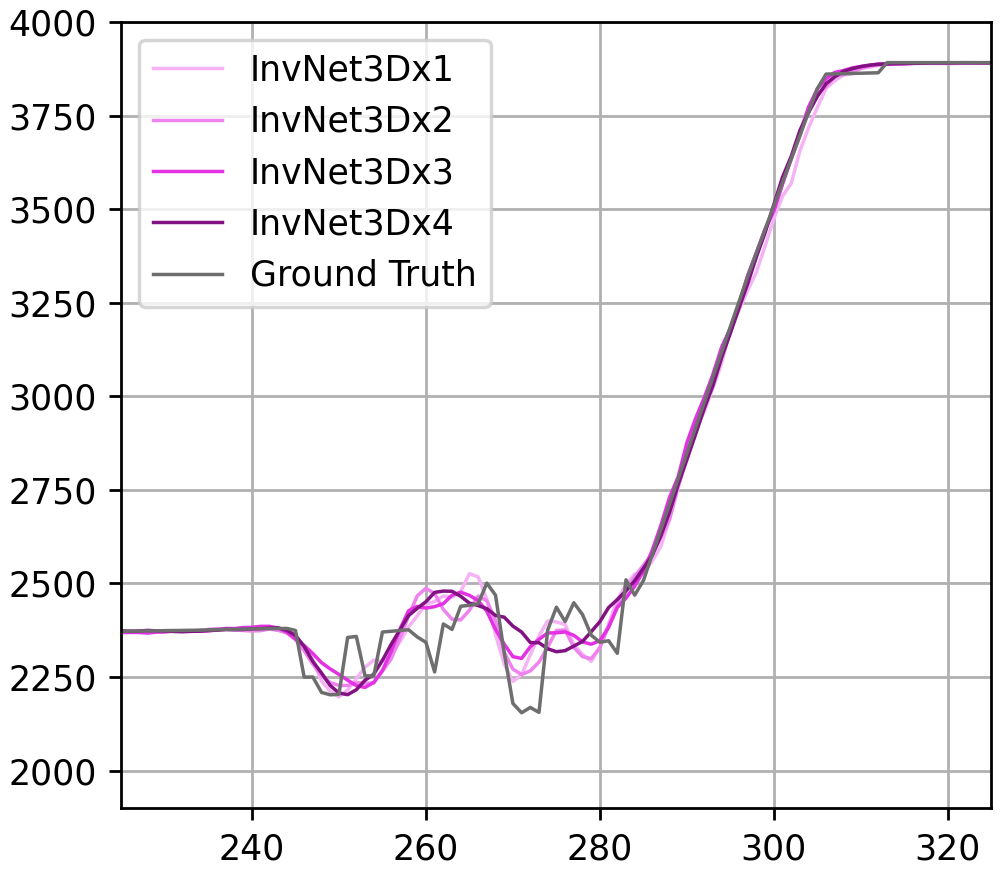}
        \end{minipage}
    }
    \subfloat[]{
        \begin{minipage}{0.18\linewidth}
        \centering
        \includegraphics[width=0.98\linewidth]{Figure/vis_zoom/year30_cut42_y186/4-InvNet3DSRCx1.png} \\\vspace{5pt}
        \includegraphics[width=0.98\linewidth]{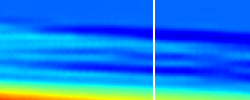} \\\vspace{5pt}
        \includegraphics[width=0.98\linewidth]{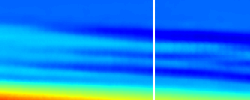} \\\vspace{5pt}
        \includegraphics[width=0.98\linewidth]{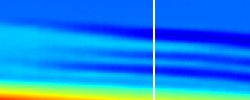} \\\vspace{5pt}
        \includegraphics[width=0.98\linewidth]{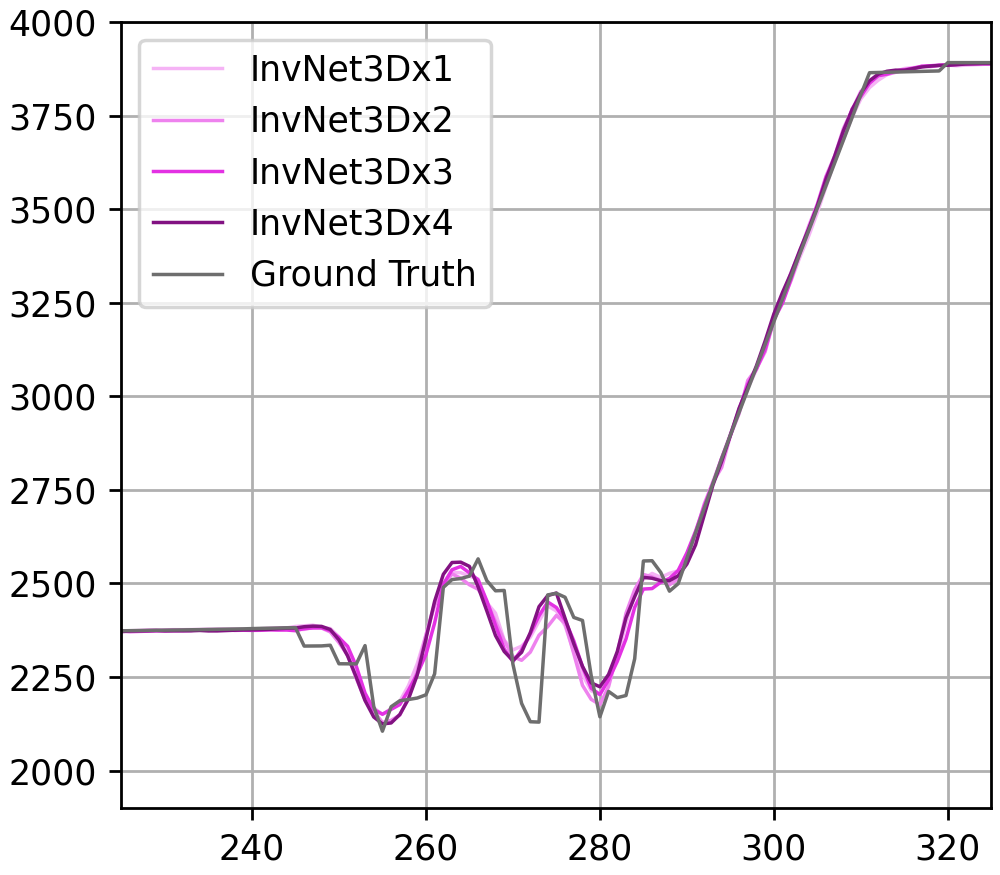}
        \end{minipage}
    }
    \subfloat[]{
        \begin{minipage}{0.18\linewidth}
        \centering
        \includegraphics[width=0.98\linewidth]{Figure/vis_zoom/year10_cut48_y324/4-InvNet3DSRCx1.png} \\\vspace{5pt}
        \includegraphics[width=0.98\linewidth]{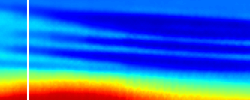} \\\vspace{5pt}
        \includegraphics[width=0.98\linewidth]{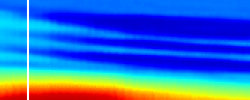} \\\vspace{5pt}
        \includegraphics[width=0.98\linewidth]{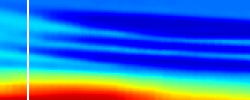} \\\vspace{5pt}
        \includegraphics[width=0.98\linewidth]{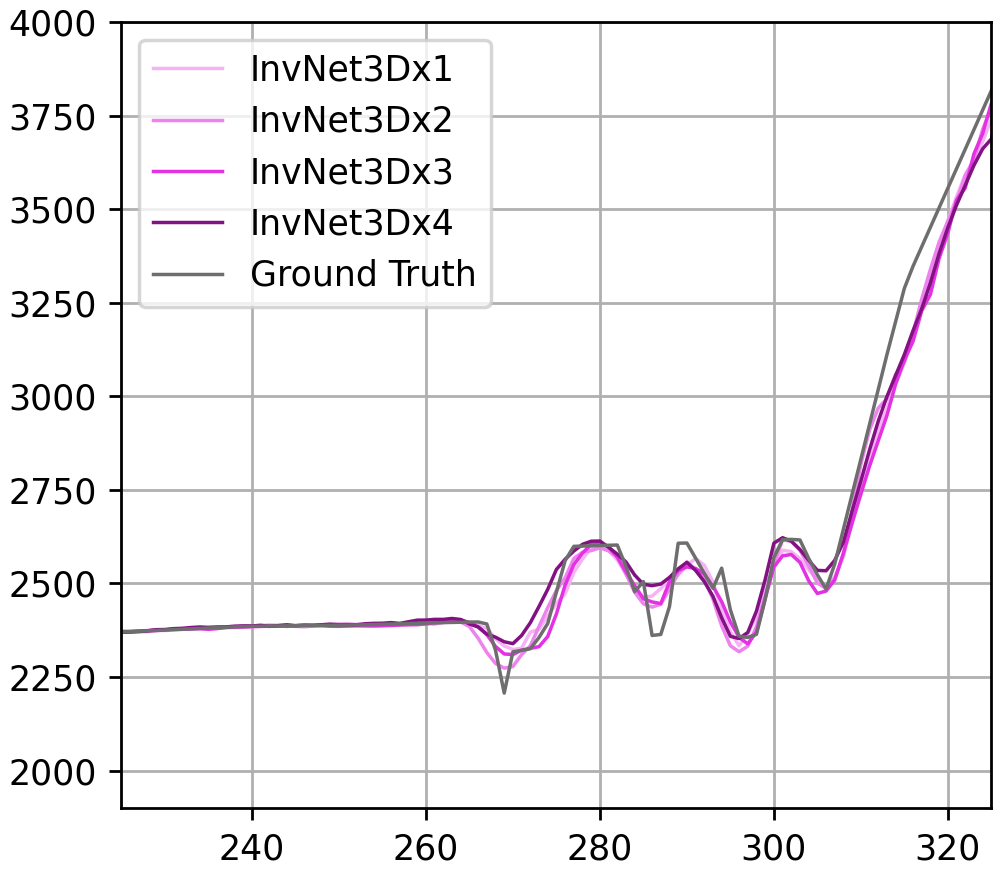} \\
        \end{minipage}
    }
    \begin{minipage}{0.06\linewidth}
    \centering
    \includegraphics[width=0.95\linewidth]{Figure/color_bar.png}
    \end{minipage}
    
    \begin{minipage}{0.95\linewidth}
    \centering
    \caption{\textbf{Zoom-in visualization of predicted velocity map samples (2D slices) generated by InvNet3Dx1, InvNet3Dx2, InvNet3Dx3 and InvNet3Dx4 (first row to fourth row).} The fifth row displays vertical velocity profiles (velocity - depth) at the location of the white cursor on 2D slices. These profiles are uniformly truncated to a depth range from 225 to 325, which covers the depth range of the zoom-in plots. 2D Ground truth velocity maps are provided in the first row of Figure \ref{fig:vis-baseline-zoom}. }
    \label{fig:vis-nblocks-zoom}
    \end{minipage}
\end{figure*}

\begin{figure*}[]
\renewcommand{\thefigure}{10A}
\centering
    \subfloat[]{ 
        \begin{minipage}{0.18\linewidth}
        \centering
        \includegraphics[width=0.98\linewidth]{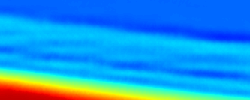} \\ \vspace{5pt}
        \includegraphics[width=0.98\linewidth]{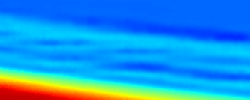} \\ \vspace{5pt}
        \includegraphics[width=0.98\linewidth]{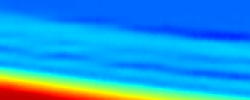} 
        \end{minipage}
    }
    \subfloat[]{
        \begin{minipage}{0.18\linewidth}
        \centering
        \includegraphics[width=0.98\linewidth]{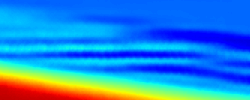} \\ \vspace{5pt}
        \includegraphics[width=0.98\linewidth]{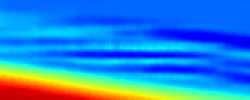} \\ \vspace{5pt}
        \includegraphics[width=0.98\linewidth]{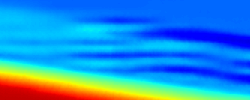} 
        \end{minipage}
    }
    \subfloat[]{ 
        \begin{minipage}{0.18\linewidth}
        \centering
        \includegraphics[width=0.98\linewidth]{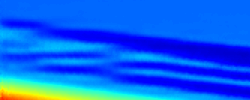} \\ \vspace{5pt}
        \includegraphics[width=0.98\linewidth]{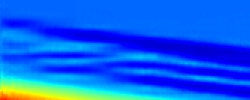} \\ \vspace{5pt}
        \includegraphics[width=0.98\linewidth]{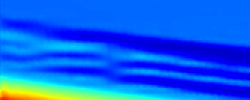} 
        \end{minipage}
    }
    \subfloat[]{ 
        \begin{minipage}{0.18\linewidth}
        \centering
        \includegraphics[width=0.98\linewidth]{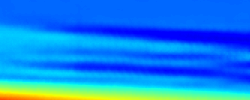} \\ \vspace{5pt}
        \includegraphics[width=0.98\linewidth]{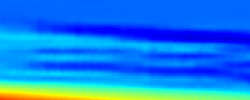} \\ \vspace{5pt}
        \includegraphics[width=0.98\linewidth]{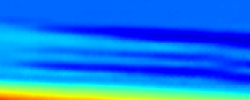} 
        \end{minipage}
    }
    \subfloat[]{
        \begin{minipage}{0.18\linewidth}
        \centering
        \includegraphics[width=0.98\linewidth]{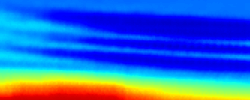} \\ \vspace{5pt}
        \includegraphics[width=0.98\linewidth]{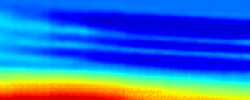} \\ \vspace{5pt}
        \includegraphics[width=0.98\linewidth]{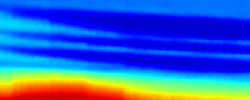} 
        \end{minipage}
    }
    \begin{minipage}{0.06\linewidth}
    \centering
    \includegraphics[width=0.95\linewidth]{Figure/color_bar.png}
    \end{minipage}
    
    \begin{minipage}{0.95\linewidth}
    \centering
    \caption{\textbf{Zoom-in visualization of predicted velocity map samples (2D slices) generated by InvNet3Dx1 with the input temporal length of 896, 448, and 224 (first row to third row).} Ground truth velocity maps are provided in the first row of Figure \ref{fig:vis-baseline-zoom}. }
    \label{fig:vis-as-temporal-zoom}
    \end{minipage}
\vspace{-0.75em}
\end{figure*}

\begin{figure*}[]
\renewcommand{\thefigure}{18}
\centering

    \begin{minipage}{1.0\linewidth}
    \centering
        \subfloat[InvNet3DSx1 \textcolor{InvNet3DSx1}{$\blacksquare$}]{
        \includegraphics[width=0.24\linewidth]{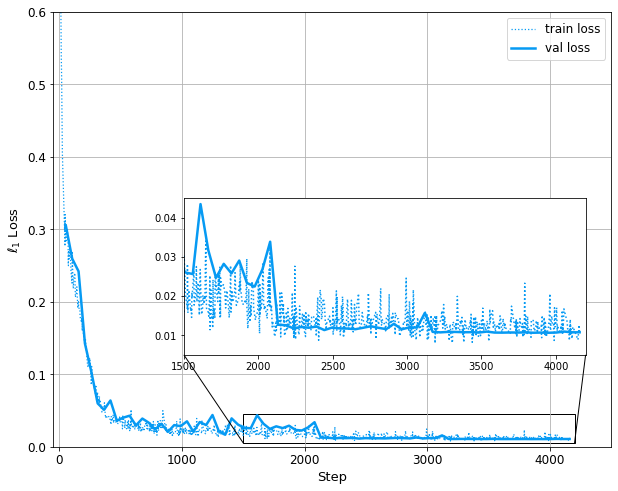}
        }
        \subfloat[InvNet3DIx1 \textcolor{InvNet3DIx1}{$\blacksquare$}]{
        \includegraphics[width=0.24\linewidth]{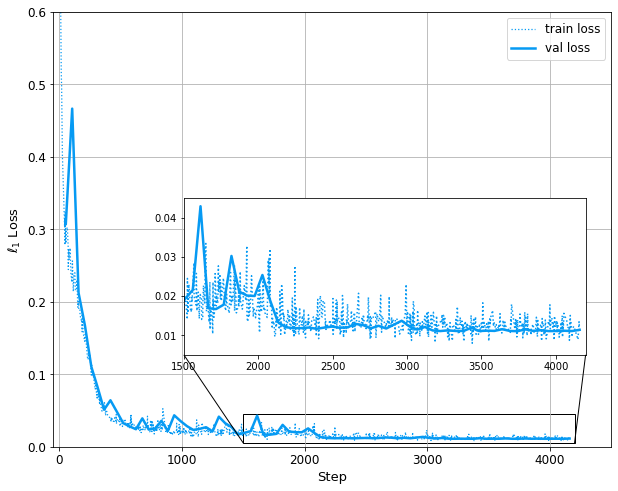}
        }
        \subfloat[InvNet3DGx1 \textcolor{InvNet3DGx1}{$\blacksquare$}]{
        \includegraphics[width=0.24\linewidth]{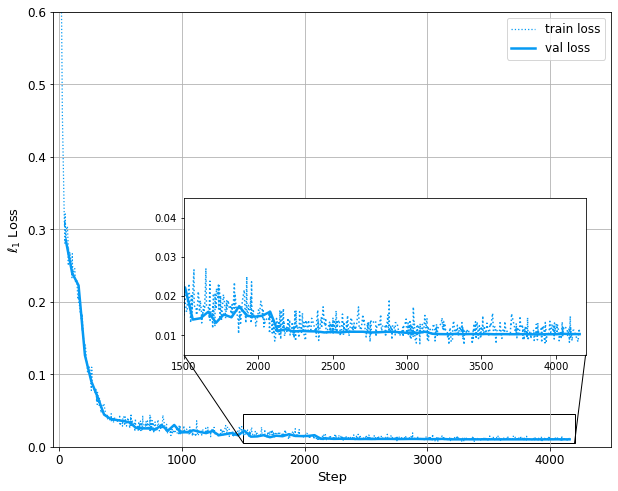}
        }
    \end{minipage} \\%
    \begin{minipage}{1.0\linewidth}
    \centering
        \subfloat[InvNet3Dx1 \textcolor{InvNet3Dx1}{$\blacksquare$}]{
        \includegraphics[width=0.24\linewidth]{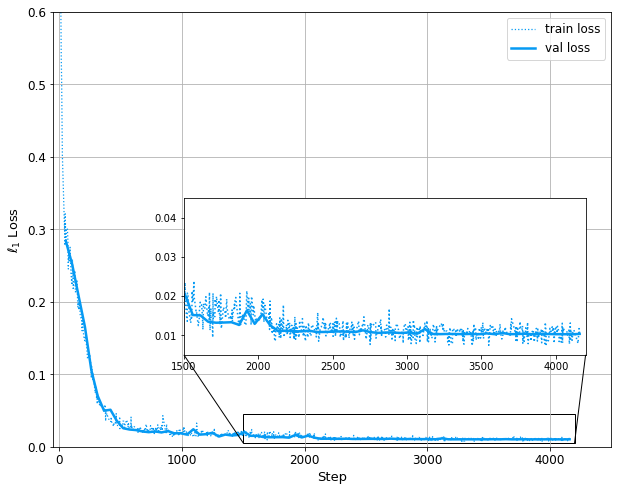}
        }
        \subfloat[InvNet3Dx2 \textcolor{InvNet3Dx2}{$\blacksquare$}]{
        \includegraphics[width=0.24\linewidth]{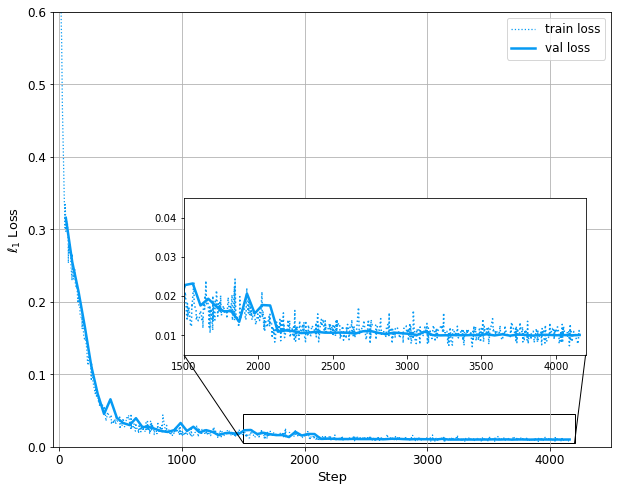}
        }
        \subfloat[InvNet3Dx3 \textcolor{InvNet3Dx3}{$\blacksquare$}]{
        \includegraphics[width=0.24\linewidth]{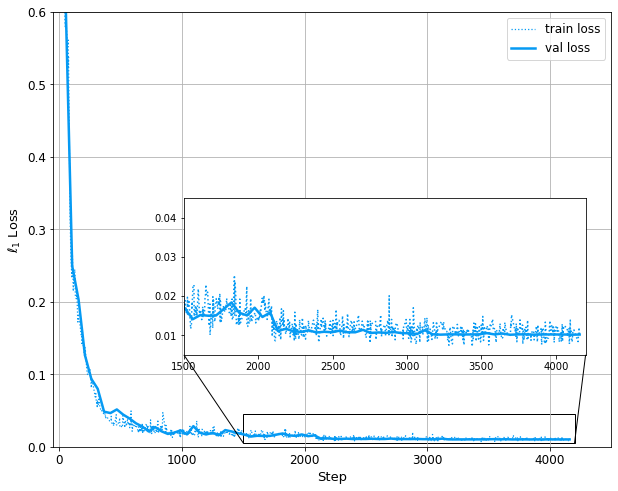}
        }
        \subfloat[InvNet3Dx4 \textcolor{InvNet3Dx4}{$\blacksquare$}]{
        \includegraphics[width=0.24\linewidth]{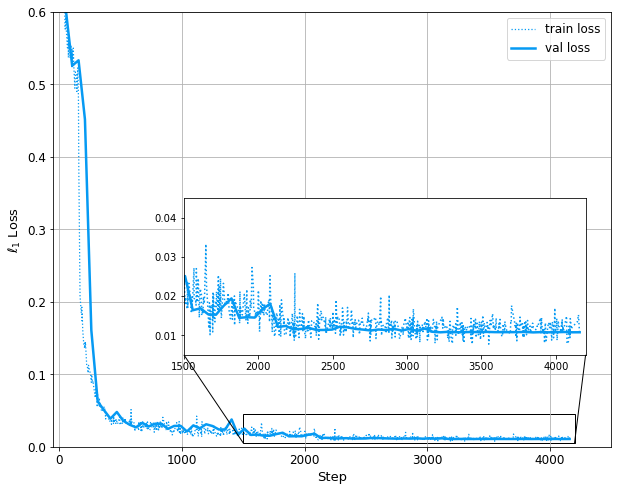}
        }
    \end{minipage} \\%
    
    \begin{minipage}{1.0\linewidth}
    \centering
    \caption{\textbf{Training and validation $\ell_1$ loss of models in main experiments.} Rectangles in captions represent the color of models in Figure 5 and/or Figure 8.}
    \label{fig:losses}
    \end{minipage}

\vspace{-0.75em}
\end{figure*}

In Section V-B, we discussed the influence of different source selection strategy on the reconstruction performance. From a traditional point of view, spatial areas that are weakly illuminated may not be correctly inverted, so an adequate number of sources and sufficient illumination must be guaranteed. However, for a deep neural network trained on large-scale dataset, learning from highly deteriorated data is possible, which has been demonstrated in image super-resolution \cite{wang2020e}, image inpainting \cite{zhu2021b}, image colorization \cite{cheng2017}, and other computer vision tasks. That being said, since FWI is a problem with strong physical background, it is important that predictions made by our network have a solid basis in the input data. It is undesirable, even though achievable, to reconstruct the velocity model of local areas using the reference from other data or certain latent principles learned by the network when corresponding seismic signals are not present in the input. Therefore, in order to understand the reliability of our results, especially when a limited number of sources are used, we computed and visualized the illumination of a sampled model with sixteen sources and one source in Figure~\ref{fig:vis-c16-illum} and Figure~\ref{fig:vis-c1-illum}, respectively. It can be clearly observed that, although the illumination with only one source is indeed much weaker than that with sixteen sources, the coverage of seismic waves still reaches almost every spatial location, even the areas of interest with reservoirs ($200\leq z\leq 300$). This indicates that our network does not predict velocity models out of thin air even with a single source and thus such predictions are not unreliable.

\bibliographystyle{IEEEtran}
\begin{bibliography}{IEEEabrv.bib,LANL-InvNet3D.bib}
\end{bibliography}